\documentclass{article}

\usepackage[preprint]{neurips_2025}

\setcitestyle{numbers,square}
\usepackage{pifont}  
\usepackage{xcolor}  
\usepackage[utf8]{inputenc}
\usepackage{amsmath}
\usepackage{amsfonts}
\usepackage{graphicx}
\usepackage{enumitem}
\usepackage{booktabs}
\usepackage{hyperref}
\usepackage{xcolor}
\usepackage{mdframed}
\usepackage{fontawesome5}
\usepackage[inkscapelatex=false]{svg}
\usepackage[breakable]{tcolorbox}
\usepackage{tabularx}
\tcbuselibrary{listings,skins}
\usepackage{calc}
\usepackage{float}
\usepackage[font=small,font=it,labelfont=bf]{caption}
\usepackage{xcolor}
\usepackage{setspace}
\usepackage{colortbl}

\usepackage{xcolor}
\definecolor{codegray}{gray}{0.95}
\newcommand{\inlinecode}[1]{\colorbox{codegray}{\texttt{#1}}}

\definecolor{humanblue}{RGB}{70, 130, 180}      
\definecolor{aigray}{RGB}{105, 105, 105}        
\definecolor{expgreen}{RGB}{46, 125, 50}        
\definecolor{lossred}{RGB}{178, 34, 34}         

\newcommand{\DirectLossIcon}{\textcolor{lossred}{\faHourglassHalf}}
\newcommand{\ExperimentalArtifactIcon}{\textcolor{expgreen}{\faFlask}}
\newcommand{\RaisesHumanIcon}{\textcolor{humanblue}{\faUser}}
\newcommand{\LimitsAIIcon}{\textcolor{aigray}{\faRobot}}

\newtcolorbox{instructionbox}{
  colback=gray!5,
  colframe=black,
  boxrule=0.8pt,
  arc=0pt,
  left=10pt,
  right=10pt,
  top=10pt,
  bottom=10pt,
  breakable=true,
  before upper={\setlength{\parskip}{7pt}},  
  floatplacement=b,  
  nofloat            
}

\usepackage{xcolor}
\definecolor{quotecolor}{RGB}{64,64,64} 
\newcommand{\devquote}[1]{\textcolor{quotecolor}{\textit{``#1''}}}
\usepackage{pifont} 
\definecolor{darkgreen}{RGB}{0,100,0}
\newcommand{\xmark}{\textcolor{red}{\ding{55}}}
\newcommand{\cmark}{\textcolor{darkgreen}{\ding{51}}}
\usepackage{tablefootnote}
\usepackage{threeparttable}

\usepackage{wrapfig}  

\newtcblisting{taskprompt}[3][Task]{%
  skin=enhanced,
  colback=blue!5!white,
  colframe=blue!75!black,
  title=#1,
  fonttitle=\bfseries,
  boxsep=1pt,
  top=3pt, 
  bottom=-5pt,
  width=#2,
  left skip=#3,
  listing only,
  listing options={
    basicstyle=\footnotesize\ttfamily,
    showspaces=false,
    showtabs=false,
    breaklines=true,
    showstringspaces=false,
    breakindent=0pt,
  }
}

\newtcblisting{taskmessage}[3][Output]{%
  skin=enhanced,
  colback=green!5!white,
  colframe=green!75!black,
  title=#1,
  fonttitle=\bfseries,
  boxsep=1pt,
  top=3pt, 
  bottom=-5pt,
  width=#2,
  left skip=#3,
  listing only,
  listing options={
    basicstyle=\footnotesize\ttfamily,
    showspaces=false,
    showtabs=false,
    breaklines=true,
    showstringspaces=false,
    breakindent=0pt,
  }
}

\hypersetup{
    colorlinks,
    linkcolor={red!50!black},
    citecolor={blue!50!black},
    urlcolor={blue!80!black}
}

\newcommand{\PctAcceptedAddedLinesOfCode}{44}

\newcommand{\PctLoomWithModelClaudeSonnetThreeFive}{23}
\newcommand{\PctLoomWithModelClaudeSonnetThreeSeven}{34}
\newcommand{\PctLoomWithModelClaudeSonnetThreeSevenThinking}{25}
\newcommand{\PctLoomWithModelOOne}{2}
\newcommand{\PctLoomWithModelGPTFourO}{11}

\newcommand{\PctLoomWithUICursorAgentOrComposer}{55}
\newcommand{\PctLoomWithUICursorChat}{48}

\newcommand{\PctDaysWithModelGeminiTwoFivePro}{3}

\newcommand{\NumIssuesAllInitialImplementationFinishedValid}{246}
\newcommand{\NumIssuesAIAllowedAllInitialImplementationFinishedValid}{136}
\newcommand{\NumIssuesNoAIAllInitialImplementationFinishedValid}{110}

\newcommand{\NumIssuesNonCoinFlipAllInitialImplementationFinishedValid}{25}
\newcommand{\MeanImplementationTimeAllInitialImplementationFinishedValidHours}{2.0}

\newcommand{\PctDevelopersPreviouslyUsedCursor}{44}

\newcommand{\PctDevelopersExitSurveySubmitHighQualityPRs}{100}
\newcommand{\PctDevelopersExitSurveyReadEveryLineAICode}{75}
\newcommand{\PctDevelopersExitSurveyEditMajorChangesAICode}{56}

\newcommand{\PctSpeedupForecastAbsValue}{24}

\newcommand{\CorrForecastAndImplementationTimeNoAI}{0.59}
\newcommand{\CorrForecastAndImplementationTimeAIAllowed}{0.64}

\newcommand{\PctSpeedupBackcastAbsValue}{20}

\newcommand{\NumForecastsMLProfessionals}{54}
\newcommand{\NumForecastsEconomists}{34}

\newcommand{\PctSpeedupForecastMLProfessionalsAbsValue}{38}

\newcommand{\PctSpeedupForecastEconomistsAbsValue}{39}

\newcommand{\PctMoreLinesOfCodeInAIAllowedIssuesPerAIDisallowedForecastedHour}{47}
\newcommand{\PValuePctMoreLinesOfCodeInAIAllowedIssuesPerAIDisallowedForecastedHour}{0.16}

\newcommand{\NumIssuesInitialImplementationStartedButNotFinished}{27}

\newcommand{\NumIssuesInvalidOtherReasons}{24}

\newcommand{\NumIssuesInvalidDitchedTooHard}{11}
\newcommand{\NumIssuesInvalidDitchedNotRelevant}{5}
\newcommand{\NumIssuesInvalidTakenByOtherMaintainer}{3}

\newcommand{\PctReviewingAIOutputAIAllowed}{9}

\newcommand{\PctWaitingOnAIAIAllowed}{4}

\newcommand{\NumLoomVideosLabelled}{128}

\newcommand{\NumLoomVideosFiltered}{74}

\newcommand{\NumLoomVideosFilteredAIAllowed}{44}
\newcommand{\NumLoomVideosInputNoAI}{54}

\newcommand{\NumLabeledHoursVideoRecordingIncludingFiltered}{143}
\newcommand{\NumLabeledHoursVideoRecordingFinalData}{84}
\newcommand{\PctTotalIssueTimeHasLabeledVideo}{29}

\newcommand{\PctDevsPreviouslyUsedWebLLM}{93}

\newcommand{\NumDevelopers}{16}
\newcommand{\MeanMaintainerCommits}{1,500}
\newcommand{\MeanDevExperienceOnReposYears}{5}
\newcommand{\MeanPctOfRepoSpentAsContrib}{59}

\newcommand{\NumDevelopersWithCursorData}{13}
\newcommand{\NumDevelopersWithNoCursorData}{3}

\newcommand{\PctDevelopersUsedCursorAfterExperiment}{69}

\newcommand{\NumAIAllowedRandomizedIncompleteIssues}{6}
\newcommand{\NumNoAIIssuesRandomizedIncompleteIssues}{7}
\newcommand{\NumDevsWithRandomizedIncompleteNoAIIssues}{3}
\newcommand{\NumDevsWithRandomizedIncompleteAIIssues}{4}

\newcommand{\MeanRepoCommits}{20,000}
\newcommand{\MeanRepoCommitters}{710}
\newcommand{\MeanRepoForks}{4,900}
\newcommand{\MeanRepoStars}{23,000}
\newcommand{\MeanRepoAgeYears}{10}
\newcommand{\MeanRepoLoC}{1,100,000}

\newcommand{\MeanReviewTimeMinutesAIAllowed}{15}

\newcommand{\MeanReviewTimeMinutesNoAI}{9}

\newcommand{\PctSpeedupBaselineAbsValue}{19}

\newcommand{\PctSpeedupBaselineImputationAddOneHourAIDisallowedIssuesAbsValue}{14}

\newcommand{\PctSpeedupBaselineImputationAddOneHourAIAllowedIssuesAbsValue}{23}

\newcommand{\PctSpeedupInitialImplementationFinishedIssuesCoinFlipRandomizationInitialImplementationTime}{20}

\newcommand{\PctSpeedupLoomDuration}{25}

\newcommand{\PctSpeedupSelfReportedTimeForThoseWithLoomDuration}{24}

\newcommand{\PctSpeedupRatioEstimator}{34}

\newcommand{\PctSpeedupDevelopersDroppedNoAIIssues}{42}

\newcommand{\PctSpeedupDevelopersDidNotDropAIIssues}{21}

\newcommand{\PctSpeedupDevelopersDidNotAnyIssues}{21}

\newcommand{\PctSpeedupUsedComparableIDEs}{24}

\newcommand{\NumLoomVideosCheating}{3}

\newcommand{\PctTreatmentNotTakenRate}{16.4}

\newcommand{\NumFactorsTotal}{21}
\newcommand{\NumFactorsThatLikelyContribute}{5}
\newcommand{\NumFactorsThatAreUnclear}{10}
\newcommand{\NumFactorsThatDoNotContribute}{6}

\usepackage{titlesec}

\renewcommand{\subsubsubsection}[1]{\paragraph{#1}~\\}

\begin{document}

\title{Measuring the Impact of Early-2025 AI on Experienced Open-Source Developer Productivity}

\author{%
  \textbf{Joel Becker}$^*$, \textbf{Nate Rush}$^*$, \textbf{Beth Barnes}, \textbf{David Rein}\\[2ex]
  \normalsize\texttt{Model Evaluation \& Threat Research (METR)}
}

\maketitle
\begin{abstract}
    Despite widespread adoption, the impact of AI tools on software development in the wild remains understudied. We conduct a randomized controlled trial (RCT) to understand how AI tools at the February--June 2025 frontier affect the productivity of experienced open-source developers. 
    \NumDevelopers\ developers with moderate AI experience complete \NumIssuesAllInitialImplementationFinishedValid\ tasks in mature projects on which they have an average of \MeanDevExperienceOnReposYears\ years of prior experience. 
    Each task is randomly assigned to allow or disallow usage of early-2025 AI tools. When AI tools are allowed, developers primarily use Cursor Pro, a popular code editor, and Claude 3.5/3.7 Sonnet. 
    Before starting tasks, developers forecast that allowing AI will reduce completion time by \PctSpeedupForecastAbsValue \%.
    After completing the study, developers estimate that allowing AI reduced completion time by \PctSpeedupBackcastAbsValue \%.
    Surprisingly, we find that allowing AI actually \emph{increases} completion time by \PctSpeedupBaselineAbsValue \%---AI tooling slowed developers down.
    This slowdown also contradicts predictions from experts in economics (\PctSpeedupForecastEconomistsAbsValue\% shorter) and ML (\PctSpeedupForecastMLProfessionalsAbsValue\% shorter).
    To understand this result, we collect and evaluate evidence for \NumFactorsTotal\ properties of our setting that \textit{a priori} could contribute to the observed slowdown effect---for example, the size and quality standards of projects, or prior developer experience with AI tooling.
    Although the influence of experimental artifacts cannot be entirely ruled out, the robustness of the slowdown effect across our analyses suggests it is unlikely to primarily be a function of our experimental design.
\end{abstract}

\renewcommand{\thefootnote}{\fnsymbol{footnote}}

\footnotetext[1]{Equal contribution. Correspondence to \texttt{\{nate, joel\}@metr.org}}

\renewcommand{\thefootnote}{\arabic{footnote}}


\section{Introduction}
\label{sec:introduction}

Software development is an important part of the modern economy, and a key domain for understanding and forecasting AI capabilities \cite{maslej2025aiindex, kwa2025measuringai}. Frontier AI systems demonstrate impressive capabilities on a wide range of software benchmarks \cite{wijk2025rebenchevaluatingfrontierai, chan2025mlebenchevaluatingmachinelearning, starace2025paperbenchevaluatingaisability, quan2025codeelobenchmarkingcompetitionlevelcode, miserendino2025swe, rein2023, rein2025hcasthumancalibratedautonomysoftware} and in experiments measuring AI's impact on developer productivity when completing synthetic tasks \cite{peng2023impactaideveloperproductivity, paradis2024doesaiimpactdevelopment}. However, tasks used in these \textit{lab experiments} sacrifice realism for scale and efficiency: the tasks are typically self-contained, do not require much prior context/familiarity to understand and complete, and use algorithmic evaluation metrics which do not capture many important capabilities \cite{raji2021everythinginthewholewideworldbenchmark, biderman2024lessonstrenchesreproducibleevaluation, davis2023benchmarksautomatedcommonsensereasoning}. As a result, it can be difficult to draw inferences from results on these evaluations about AI's impact in practice.

To reduce the inferential gap between measurements of AI capabilities and real-world impact, one can measure the impact of AI systems in real-world settings (i.e. \textit{field experiments}). Existing field experiments aimed at measuring AI's impact on software development measure outcomes like number of added lines of code or number of tasks completed \cite{gambacorta2024generative,yeverechyahu2025impactllmopensource,cui2025effectsgenerativeai}. However, AI systems can affect these outcomes without productivity actually increasing---for example, code can be more verbose but functionally equivalent, and tasks can be broken up into multiple smaller tasks without the total amount of work changing---making it challenging to interpret these results.

To directly measure the impact of AI tools on developer productivity, we conduct a randomized controlled trial by having \NumDevelopers\ developers complete \NumIssuesAllInitialImplementationFinishedValid\ tasks (\MeanImplementationTimeAllInitialImplementationFinishedValidHours\ hours on average) on well-known open-source repositories (\MeanRepoStars\ stars on average) they regularly contribute to. Each task is randomly assigned to allow or disallow AI usage, and we measure how long it takes developers to complete tasks in each condition\footnote{Crucially, the tasks are defined \textit{before} they are randomized, limiting the impact of effects from AI assistance unrelated to productivity (e.g., more verbose but functionally equivalent code)}. Developers, who typically have tens to hundreds of hours of prior experience using LLMs\footnote{While \PctDevsPreviouslyUsedWebLLM\% of developers have previously used LLMs, only \PctDevelopersPreviouslyUsedCursor\% have prior experience using the Cursor IDE.}, use AI tools considered state-of-the-art during February--June 2025 (primarily \href{https://www.cursor.com/}{Cursor Pro} with Claude 3.5/3.7 Sonnet). We collect screen recordings as they work, providing a rich data source for analysis. 

\begin{figure}[t]
    \centering
    \includegraphics[width=1\linewidth]{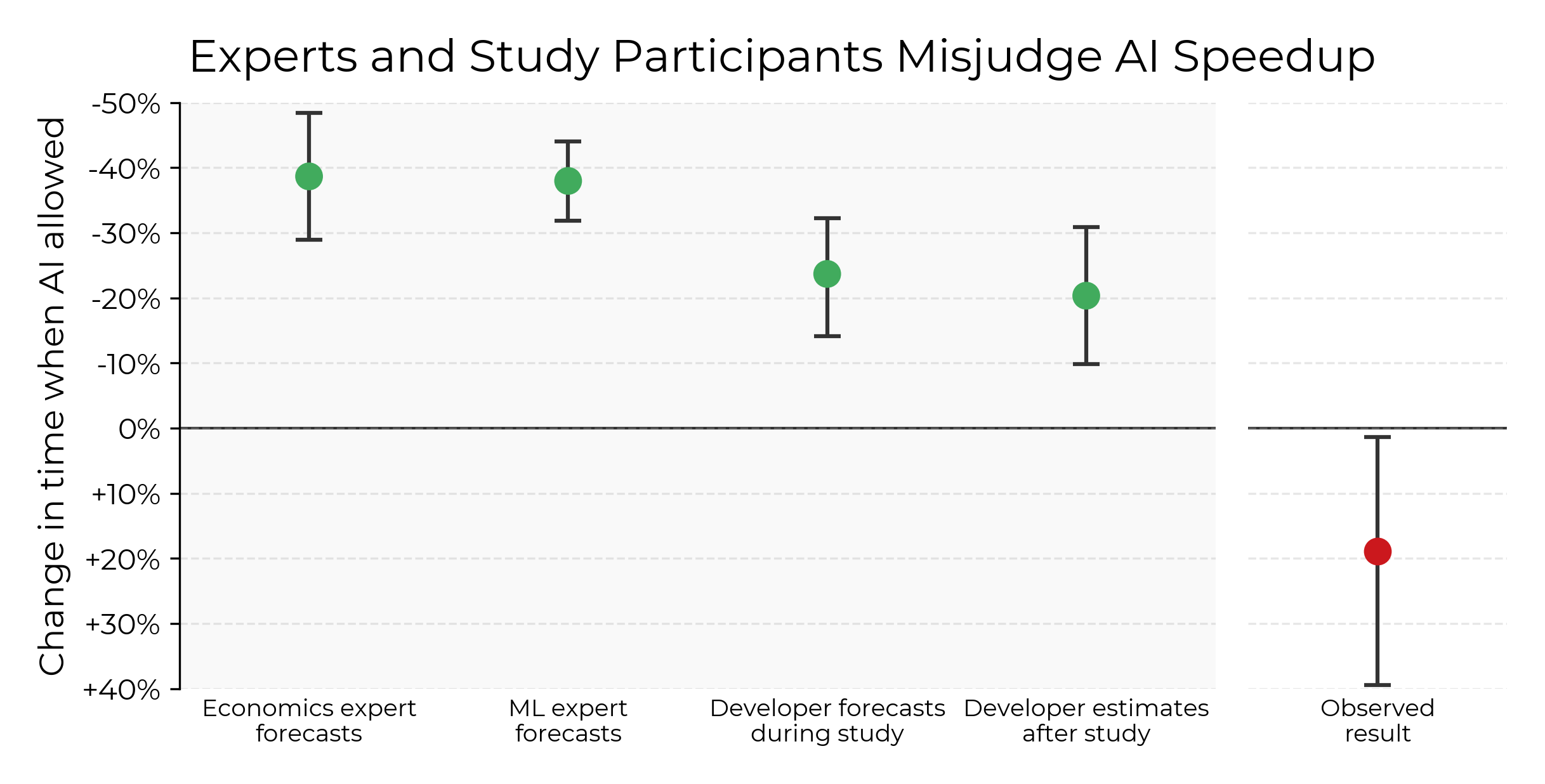}
    \caption{Experts and study participants (experienced open-source contributors) substantially overestimate how much AI assistance will speed up developers---tasks take \PctSpeedupBaselineAbsValue\% more time when study participants can use AI tools like Cursor Pro. See \autoref{sec:empirical-strategy} for detail on speedup percentage and confidence interval methodology.}
    \label{fig:horizontal_iceberg}
\end{figure}

Before tasks are randomized, developers forecast that allowing AI will reduce completion time by \PctSpeedupForecastAbsValue \%. After study participation, developers estimate that allowing AI reduced completion time by \PctSpeedupBackcastAbsValue \%.
Surprisingly, we find that allowing AI actually \emph{increases} completion time by \PctSpeedupBaselineAbsValue \%--- developers are slower when using AI tooling. 
\autoref{fig:horizontal_iceberg} displays this observed slowdown in contrast with forecasts and post-hoc developer estimates of speedup from AI. We also collect forecasts of speedup from machine learning and economics experts in academia and industry, and find that they also substantially overestimate our observed speedup.

To understand this surprising result, we manually label \NumLabeledHoursVideoRecordingIncludingFiltered\ hours of recordings of developers' computer screens while they work (representing \PctTotalIssueTimeHasLabeledVideo\% of the total hours spent by developers), which allows us to decompose how they spend their time when working with and without AI assistance at a resolution of $\sim\!10$ seconds. We additionally collect rich statistics from source-code management systems, interview and survey participating developers, and conduct subset analyses to better understand the nature of the slowdown result.


Using these various sources of data, we identify \NumFactorsTotal~properties of our setting and experimental design that we hypothesize \textit{a priori} may contribute to the slowdown effect. 
We group these factors into four categories: a) direct productivity loss, b) experimental artifact, c) factors raising human performance, and d) factors limiting AI performance. 
We find evidence that \NumFactorsThatLikelyContribute~factors contribute to the slowdown effect, we find mixed/unclear/no evidence for \NumFactorsThatAreUnclear~factors, and we find evidence against \NumFactorsThatDoNotContribute~factors contributing to the slowdown effect. \autoref{sec:factor_analysis} presents these factors at a high level, and \autoref{sec:allfactors} discusses each factor in detail.
While we can't completely rule out the impact of experimental artifacts, the slowdown effect appears broadly robust across a wide range of experimental design decisions.

That said, many of the factors we find evidence for contributing to slowdown are specific to the setting we study---these results do \textit{not} imply that current AI systems are not useful in many realistic, economically relevant settings. 
Furthermore, these results do not imply that future models will not speed up developers in this exact setting---this is a salient possibility given the rapid pace of progress in AI capabilities recently \cite{kwa2025measuringai}. Finally, it remains possible that further improvements to current AI systems (e.g. better prompting/agent scaffolding, or domain-specific finetuning) could yield positive speedup in this setting.

Nonetheless, our results reveal a large disconnect between perceived and actual AI impact on developer productivity. Despite widespread adoption of AI tools and confident predictions of positive speedup from both experts and developers, we observe that AI actually slows down experienced developers in this setting.

\subsection{Background}
\label{sec:Background}

\paragraph*{Speedup, but on synthetic tasks}
Literature on productivity improvements on software tasks due to AI usage broadly finds that AI tools increase productivity. 
\citet{peng2023impactaideveloperproductivity} and \citet{paradis2024doesaiimpactdevelopment} find 56\% and 21\% speedups on coding tasks when using AI assistance, and \citet{10.1145/3661145} finds a 65\% increase in the rate of task requirements satisfied with AI tools. However, these studies use artificial/synthetic tasks that make it difficult to directly draw inferences about the real-world impact of AI tools. For example, \citet{peng2023impactaideveloperproductivity} asks developers to implement a very basic HTTP server in JavaScript to satisfy several automatic test cases that are shown to the developers---this task is a) unrepresentative of most software development work, and b) likely to be similar to a large amount of LLM training data, which may unfairly advantage AI systems relative to humans.

\paragraph*{Speedup, but with non-fixed outcome measures}
Other literature uses tasks found ``in the wild,'' either via natural experiments \cite{yeverechyahu2025impactllmopensource} or randomized controlled trials \cite{gambacorta2024generative, cui2025effectsgenerativeai}, finding 14-51\% increases in output productivity metrics. However, these studies use outcome measures that are not fixed in advance---i.e. lines of code written, number of code commits, and pull requests\footnote{See \autoref{sec:Primer on Open Source Development} for a primer on open-source development terminology.} (PRs) as their key outcome measures respectively. It's possible for AI assistance to affect the outcomes without actually increasing productivity, e.g. by causing developers to write more verbose but functionally equivalent code, or causing them to break up pull requests into smaller chunks of work.


\paragraph*{Impressive AI benchmark results}
This general consensus around AI tooling's effect on software developer productivity is perhaps unsurprising, given the impressive apparent capabilities of frontier AIs on challenging question-answering and agentic tasks used in popular AI benchmarks \cite{openai2025o3o4mini, anthropic2025claude4}.

\paragraph*{Heterogeneous effects by experience}
One important question that emerges given these impressive results is whether productivity gains are captured by individuals of all experience levels. 
The canonical framework of \citet{AmbiguousLaborMarket} treats AI as a fall in the cost of prediction, with distributional consequences depending on which complementary sub-problems the tool does not solve. 
Existing empirical work on the micro-level effects of generative AI tools tends to find that access to these tools benefits less experienced workers more, compressing performance distributions \cite{brynjolfsson2023generativeaiatwork, noy2023experimental, peng2023impactaideveloperproductivity, choi2024ai}. 

These heterogeneous effects motivate our focus on highly skilled open‑source developers, as there has been relatively less research in this setting. 

\paragraph*{Mixed speedup results in other domains}
Some literature measures the impact of frontier AI systems in settings other than software development, for example, for \href{https://en.wikipedia.org/wiki/CBRN_defense}{CBRN} uplift risk assessment, finding mixed results with recent AI systems \cite{anthropic2024claude3, mouton2024operational, grattafiori2024llama3herdmodels, patwardhan2024earlywarning}. 
Other research finds substantial productivity increases in non-software domains \cite{brynjolfsson2023generativeaiatwork, noy2023experimental}.

\paragraph*{Understanding AI's economic impact}
Finally, some literature tries to predict how AI capability advances might a) affect the rate of AI progress (e.g. if AI systems can substitute for human AI R\&D labor), or b) broadly impact the economy. \citet{leibowich2024advancedai} interview AI researchers about how full automation of AI R\&D might alter the pace of advancement, several papers explore the possibility of explosive economic growth via large-scale AI labor substitution \cite{erdil2024explosivegrowthaiautomation, erdil2025gateintegratedassessmentmodel, davidson2023compute}, and the economics literature includes both optimistic and skeptical perspectives on AI's productivity impact \cite{acemoglu2024simplemacroeconomics, agrawal2018economicpolicyforai, furman2018aiandeconomy}.

Our study primarily complements existing literature measuring the impact of AI on software development by: 
\begin{enumerate}
    \item Testing AI models at the February--June 2025 frontier,
    \item Using unfiltered, ``live'' open-source repository tasks rather than synthetic or cherry‑picked tasks,
    \item Using a fixed outcome measure (speedup on tasks defined before randomized treatment assignment),
    \item Recruiting experienced engineers with years of expertise in the target repositories, and
    \item Collecting rich data on time usage, AI code suggestions, and developers' qualitative experiences.
\end{enumerate}


\section{Methodology}
\label{sec:Methodology}

\begin{figure}[t]
    \centering
    \includegraphics[width=1\linewidth]{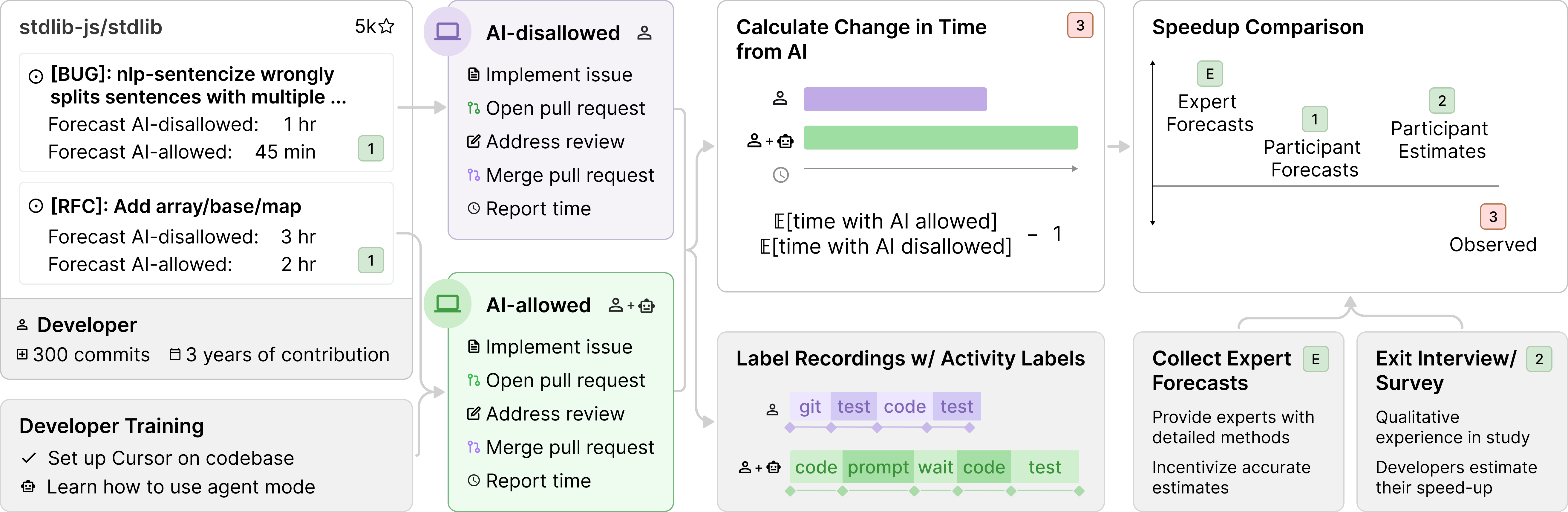}
    \caption{Our experimental design. Tasks (referred to as \textit{issues}) are defined before treatment assignment, screen recordings let us verify compliance (and provide a rich source of data for analysis), and forecasts from experts and developers help us measure the gap between expectations and observed results.}
    \label{fig:figure_zero}
\end{figure}

\subsection{Developers and Repositories}

We recruit experienced developers from large open source repositories to work on real tasks defined on these repositories.
Developers come from a mix of our professional networks and from outreach to active contributors to large, popular Github repositories. 
The developers are experienced software engineers (typically over a decade of experience), and are regular contributors to the repositories we use---on average, they have \MeanDevExperienceOnReposYears\ years of experience working on their repository, representing \MeanPctOfRepoSpentAsContrib\% of that repository's lifetime, over which time they have made \MeanMaintainerCommits\ commits to the repo. 
As an incentive to participate, we pay developers \$150/hour. \autoref{sec:RecruitmentAndOnboarding} provides more detail about our recruitment and incentivization process.

The repositories themselves are large and mature. On average, they have \MeanRepoStars\ stars, \MeanRepoLoC\ lines of code, \MeanRepoForks\ forks, \MeanRepoCommits\ commits, and \MeanRepoCommitters\ committers, and they broadly have very high quality bars for code contributions. For example, one set of repository contribution guidelines concludes: \devquote{Phew. While the above may be a lot to remember [..] the motivation for enforcing process is to ensure that all code contributions meet a certain quality threshold.} 
\autoref{sec:devandrepostatistics} details further statistics about individual developers and repositories.

\subsection{Experimental Design}

\begin{figure}[t]
    \centering
    \includegraphics[width=1\linewidth]{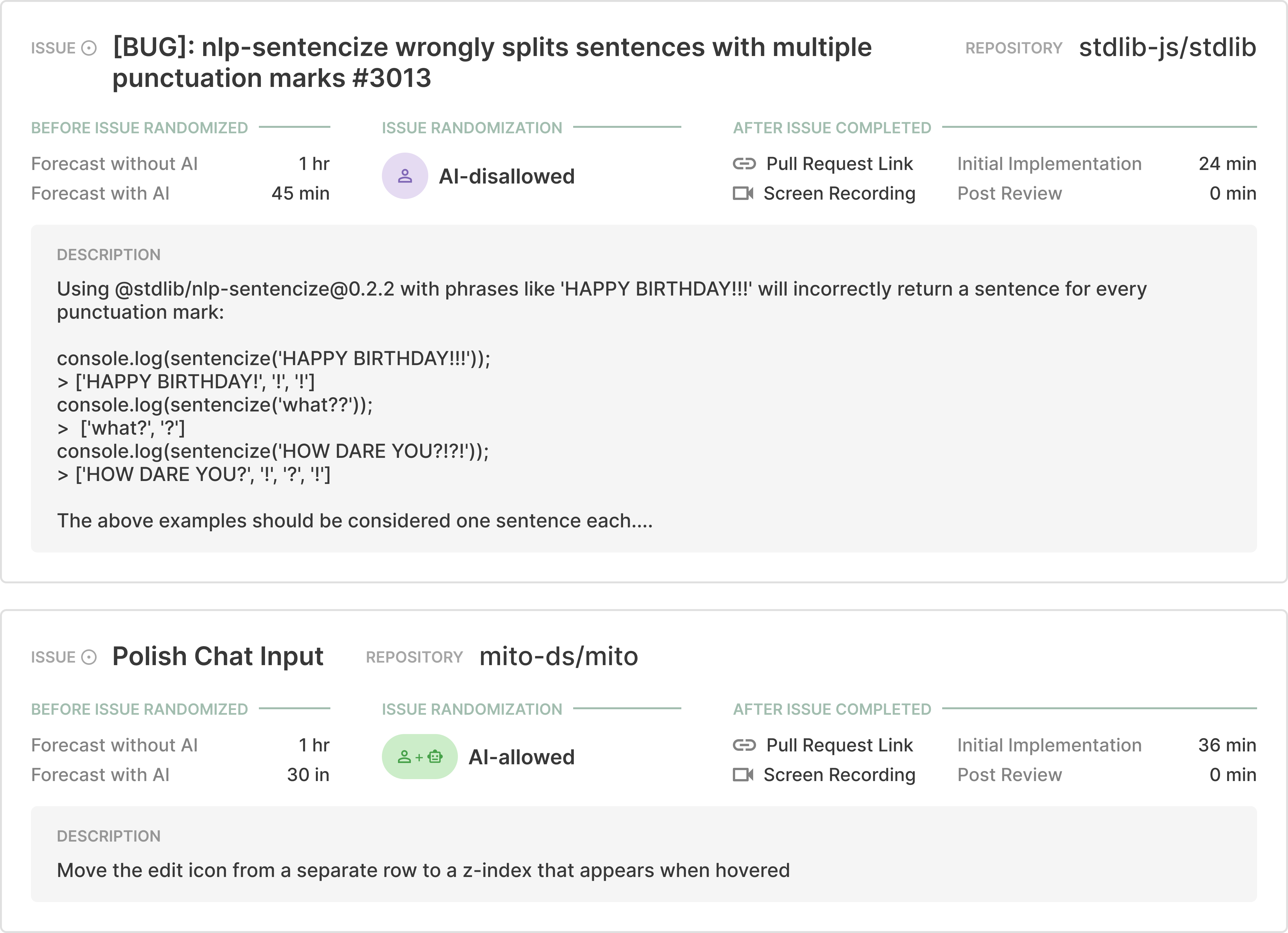}
    \caption{Real issues completed during the study from the \texttt{stdlibjs} and \texttt{mito} repositories}
    \label{fig:figure_issues}
\end{figure}

Each developer provides a list of real issues in their repository to work on as part of this study. 
Issues are typically bug reports, feature requests, or work items used to coordinate development. They range from brief problem descriptions to detailed analyses and represent work ranging from minutes to hours. Two example issues are shown in \autoref{fig:figure_issues}.
Many issues are defined before the study period begins, but some are created during the study period.\footnote{About half of issues included in the study were not formally defined on the repository, and instead were tracked separately for our experiment. Importantly, all issues represent real work that developers wanted to contribute to their repositories, and all work completed by developers is submitted and reviewed through each repository's standard source-code management system (e.g. GitHub/GitLab). Developers are asked to contribute issues taking a maximum of two hours, or to break up issues taking longer into multiple issues.}

After collecting this issue list, developers forecast how long each issue would take if they were to complete it both with and without AI assistance. We use these forecasts as a proxy for issue difficulty, and to measure per-issue speedup anticipated by the developer. These issues are then randomized to one or the other condition via a simulated fair coin flip.\footnote{\NumIssuesNonCoinFlipAllInitialImplementationFinishedValid\ issues early in the study were randomized differently. We show that results are not sensitive to inclusion/exclusion of these issues in \autoref{sec:factor-specific-outcome-measure}, and we include these issues for statistical power.} If AI is allowed, developers can use any AI tools or models they choose, \textit{including no AI tooling if they expect it to not be helpful}. If AI is not allowed, no generative AI tooling can be used.\footnote{AI-based tab autocomplete is disallowed in the AI-disallowed condition if it uses LLMs (e.g. GitHub Copilot) but allowed otherwise. Search engines, which sometimes use AI under the hood, remain allowed in the AI-disallowed condition.} 

Developers then work on their assigned issues in their preferred order---they are allowed to flexibly complete their work as they normally would, and sometimes work on multiple issues at a time. After completing an issue to their satisfaction, they submit a pull request (PR) to their repository, which is typically reviewed by another developer. They make any changes suggested by the PR reviewer, and merge their completed PR into the repository\footnote{Not all PRs end up getting merged by the end of the study period---see \autoref{sec:factor-specific-outcome-measure} for discussion of how this may affect results.}. As the repositories included in the study have very high quality and review standards, merged PRs rarely contain mistakes or flaws. Finally, they self-report how long they spend working on each issue before and after PR review.

See \autoref{sec:dev_instructions_and_survey_data} for the full written instructions given to developers before they start working.

\subsubsection{AI Tools and Training}

Two popular means of using modern large language model (LLM) based AI tools are via web-based user interfaces (e.g. \href{https://chatgpt.com}{chatgpt.com}) and the integrated development environment (IDE) \href{https://www.cursor.com/}{Cursor Pro} (which we provide a subscription for). 
Cursor is a fork of the widely used VSCode IDE with near-identical features, that additionally includes extra AI features like a language model chat interface, and an AI agent tool that can search and edit files, run arbitrary bash commands, prompt/ask the user for more details when relevant, and iterate/debug programs without constant input from users. Developers have a range of experience using AI tools: \PctDevsPreviouslyUsedWebLLM\% have prior experience with tools like ChatGPT, but only \PctDevelopersPreviouslyUsedCursor\% have experience using Cursor.

We provide developers with Cursor Pro subscriptions and conduct live basic training, validating that developers are able to prompt Cursor effectively to edit files in their own codebase, accept changes, and revert to previous checkpoints. However, we don't require that they use Cursor specifically. Developers working on issues for which AI is allowed can use any AI tools of their choosing, or no AI tools if they prefer.
See \autoref{sec:Primer on AI Tooling} for further information on these two methods of accessing AI assistance, and \autoref{sec:RecruitmentAndOnboarding} for more detail about our training and onboarding process.

\subsubsection{Data Collection}

Contributors completed issues largely as they would outside of our experiment, with a few exceptions: they typically record their screen as they work (providing us with a source of rich data on their AI usage and working patterns), when using AI they often use the Cursor IDE, which sometimes differs from their normal development environment (e.g. neovim, PyCharm), and they record details of their working patterns after each issue is completed. Broadly, developers reported that they were not significantly inconvenienced or affected by these differences compared to their normal workflows (see section \autoref{sec:factor_analysis} for discussion of how experimental design may affect results). Developers self-report the length of time they spend working on each issue before and after PR review.

\subsection{Effect Estimation}
\label{sec:effect_estimation}

We aim to measure the percentage change in issue time to completion due to AI, defined as:

$$S = \frac{\mathbb{E}[\text{completion time with AI allowed}]}{\mathbb{E}[\text{completion time with AI disallowed}]} - 1$$

A value of -50\% would indicate issues assigned to AI-allowed taking half the time of issues assigned to AI-disallowed; 0\% would indicate that issues take the same time to complete with and without AI-allowed; 100\% would indicate that issues assigned to AI-allowed take twice as long as issues assigned to AI-disallowed. For convenience and clarity, we typically report the absolute value of $S$, and colloquially refer to negative values of as \textit{speedup}, and positive values as \textit{slowdown}\footnote{This is a mild abuse of language, because speedup and slowdown refer to \textit{rates}, while the quantity we actually estimate and report is the percentage change in time from AI.}

We use the total implementation time (including both pre- and post-pull request review time) developers self-report spending on each issue to estimate the percentage change in issue completion time due to AI. A small number of issues do not end up getting reviewed/merged for various reasons---we impute the post-review implementation time of these issues with the mean post-review time per treatment condition. \autoref{sec:imputation_robustness} shows that our results are robust to the specific imputation method used.

We estimate $S$ via a log-linear regression for total implementation time, incorporating estimates of issue difficulty to reduce noise in our estimate. We also use adapted versions of this regression when examining how performance varies in different subsets of the data (to efficiently account for any post-randomization issue difficulty imbalances that may be more pronounced within subsets). See \autoref{sec:empirical-strategy} for details of the regression approach used. 

\subsection{Supplementary Data Collection}

\paragraph*{Developer Forecasts and Post hoc Estimates}
Before randomizing each issue to either AI-allowed or AI-disallowed, we ask developers to forecast how long they expect each issue to take if they are allowed to use AI assistance, and if they are disallowed from using AI. 
This provides a measure of issue difficulty independent of random treatment assignment, and enables us to assess developer calibration and expectations regarding the usefulness of AI assistance for their work. 

After the study's conclusion, we ask developers to estimate post hoc how much AI tooling affected the time it took them to complete issues. This helps us understand the relationship between individuals' beliefs about AI helpfulness and our measured effects.

The detailed instructions given to developers for per-issue forecasts and exit surveys are available in \autoref{sec:dev_instructions_and_survey_data}.

\paragraph*{Expert Forecasts}
\label{sec:expert_forecasts}

We solicit forecasts from experts in machine learning and economics in order to compare our results against expert prior expectations regarding AI capabilities \cite{PilotinganElicitation}. Machine learning experts are a mix of industry professionals from frontier AI labs, and academics from universities with leading graduate computer science programs. Economics experts are primarily academics and PhD students from universities with leading economics graduate programs. See \autoref{sec:other_analysis_expert_forecasts} for summary statistics of expert forecasts by expert group. 

Experts are given detailed information about our study design, repositories participating in our experiment, and the proportion of developers who have prior experience with AI tooling (\autoref{sec:instructions-given-to-expert-forecasters}). They then report their prediction of our point estimate for 
$\frac{\mathbb{E}[\text{time with AI disallowed}]}{\mathbb{E}[\text{time with AI allowed}]}$
\footnote{Note that the estimate we use for \autoref{fig:horizontal_iceberg} is transformed to represent $\frac{\mathbb{E}[T\,|\,\text{AI}=1]}{\mathbb{E}[T\,|\,\text{AI}=0]} - 1$. It is not necessarily true that forecasters' belief about the point estimate of $\frac{\mathbb{E}[T\,|\,\text{AI}=1]}{\mathbb{E}[T\,|\,\text{AI}=0]}$ is equal to the reciprocal of their belief about the point estimate of $\frac{\mathbb{E}[T\,|\,\text{AI}=0]}{\mathbb{E}[T\,|\,\text{AI}=1]}$.}.

To incentivize experts to make accurate forecasts, we offer to pay $\max(0, \$50 \times (1 - (\text{true answer} - \text{guessed answer})^2))$ for their point estimates.\footnote{Approximately one-third of forecasters are offered a maximum of \$100 rather than \$50.}\footnote{Taking a maximum with 0 makes our scoring rule improper \cite{Gneiting01032007}---their reward is not necessarily maximized at their true belief. We use this scoring rule for simplicity and clarity.}

\paragraph*{Cursor Analytics Data}
We collect per-developer analytics data from Cursor's analytics dashboard, for the \NumDevelopersWithCursorData\ developers we provide Cursor Pro subscriptions to. For each developer, we collect their most-used model and the number of suggested and accepted lines of AI-generated code.

\paragraph*{Exit Interviews and Surveys} 
After the study's conclusion, we conduct exit interviews and surveys with all developers to assess where they found AI helpful, what strategies they used to effectively elicit work from AI, whether they felt they improved at using AI tooling over the course of the study, and to estimate how much they were sped up by AI during the study. Full details of the exit interviews are available in ~\autoref{sec:exit_interviews}.

\paragraph*{Qualitative Evidence} 
Throughout the study we collect qualitative evidence from developers, to form a more comprehensive understanding of their experiences using AI tools. Developers are instructed to take detailed notes regarding their experience and usage of AI tools, and we use inductive coding---where we iteratively read through the data to identify recurring patterns, create and refine categories as they emerge, and reorganize excerpts until stable themes develop---to cluster excerpts from these notes. Qualitative results in \autoref{sec:factor_analysis} (particularly quotes from developers) were collected primarily with this methodology. However, to investigate initial hypotheses, we often ask developers probing/targeted questions, so we cannot rule out bias from this type of qualitative evidence.

\paragraph*{Fine-Grained Screen Recording Activity Labels}
\label{sec:screen_recording_labels}

To compare how developers spend their time with and without AI assistance, we manually label a subset of \NumLoomVideosLabelled\ screen recordings with fine-grained activity labels, totaling \NumLabeledHoursVideoRecordingIncludingFiltered\ hours of video. In results based on these labeled screen recordings, we filter to remove issues where we find cheating, issues where the screen recording is broken for \textgreater 10\% of recording time, and issues with a \textgreater 20\% discrepancy between self-reported time and the recording time. This results in \NumLoomVideosFiltered\ recordings representing \NumLabeledHoursVideoRecordingFinalData\ hours of video for further analysis.

We label whether developers are: actively writing code, testing and debugging their code, reading or searching for information, using Git or managing their environment, prompting an AI system, waiting on an AI system to generate output, reviewing AI outputs, or idling/doing other miscellaneous work. Each high-level label is further broken down into one of 27 fine-grained categories. Labels have a resolution of $\sim \!10$ seconds. \autoref{sec:screen_recording_instructions} describes the instructions and process used for labeling screen recordings.


\section{Results}
\label{sec:results}

\begin{figure}[t]
    \centering
    \includegraphics[width=1\linewidth]{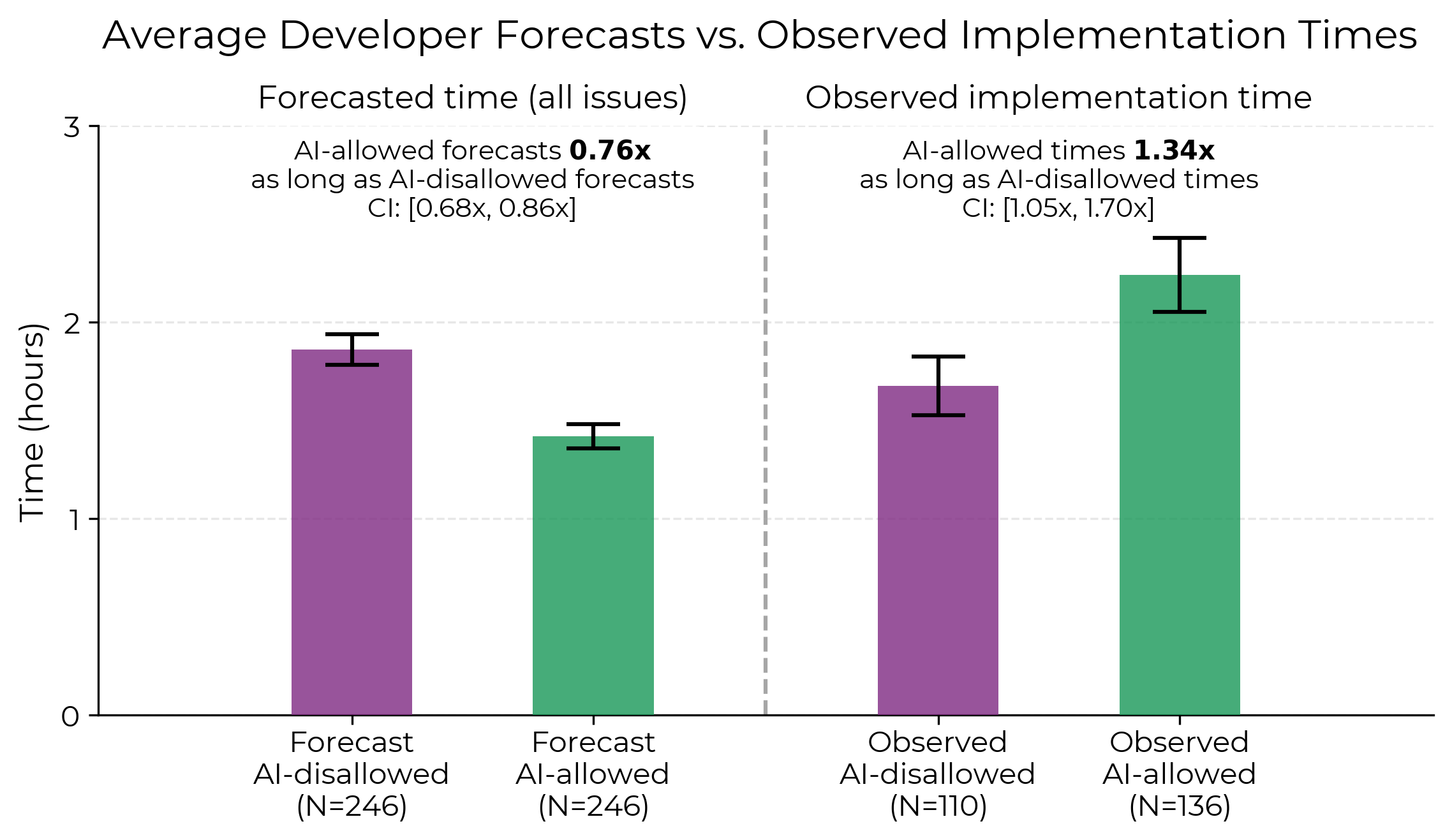}
    \caption{Left: Raw average forecasted implementation times. Right: Raw average observed implementation times. The ratio of observed implementation times gives a more extreme slowdown estimate than regression-based estimates (\autoref{sec:regression}) because AI-allowed issues are forecasted (importantly, before treatment assignment) by developers to take slightly longer, which the regression corrects for. Both: \autoref{sec:ratio-estimator} describes confidence intervals around ratios of average times.}
    \label{fig:forecast_vs_observed_means}
\end{figure}

\begin{figure}[t]
    \centering
    \includegraphics[width=1\linewidth]{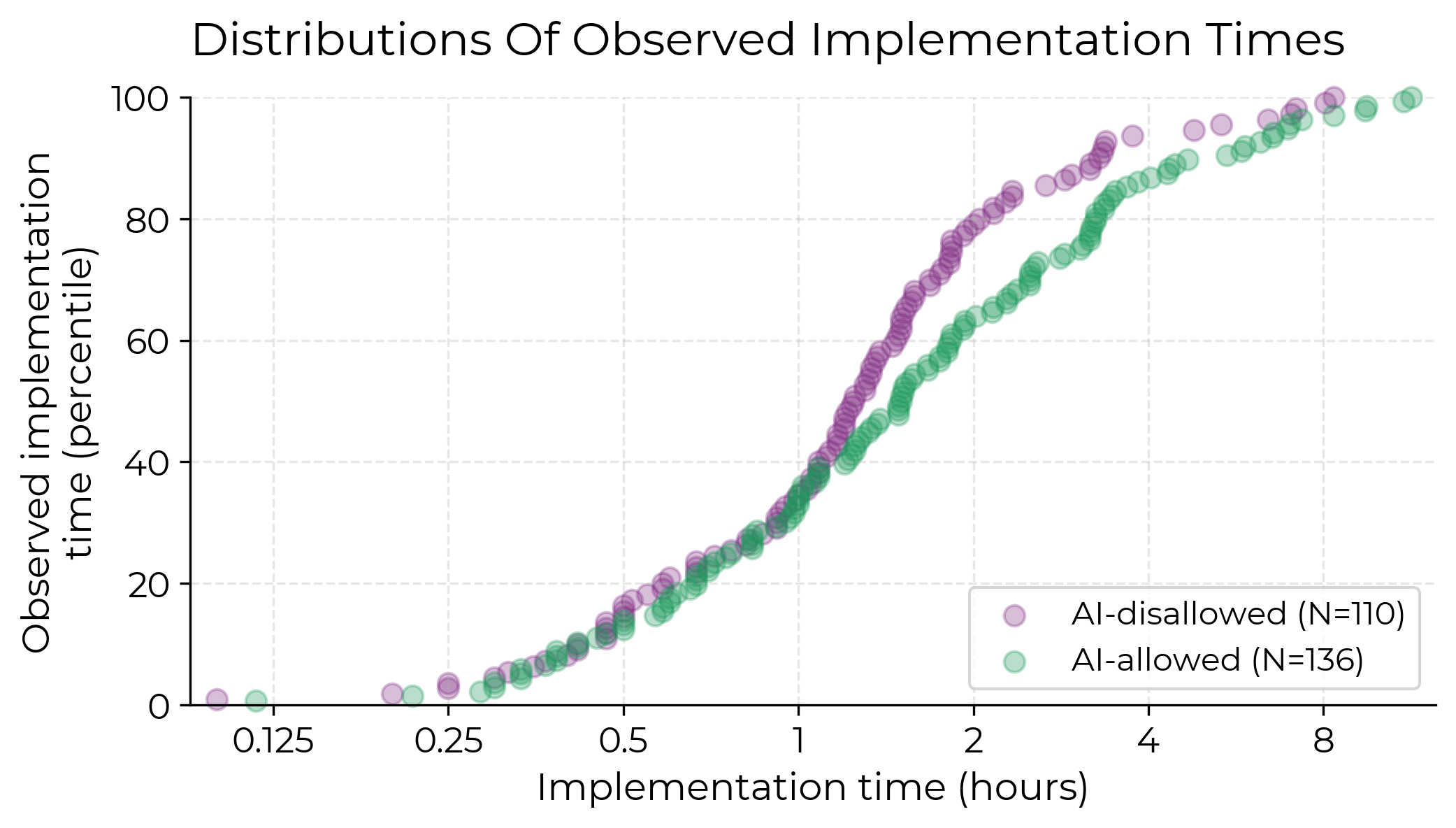}
    \caption{Empirical cumulative distribution functions of observed implementation times. Percentile ordering is calculated separately for each treatment group.}
    \label{fig:observed_ecdf_scatter}
\end{figure}

Developers complete \NumIssuesAIAllowedAllInitialImplementationFinishedValid\ issues with AI-allowed and \NumIssuesNoAIAllInitialImplementationFinishedValid\ issues with AI-disallowed. \autoref{sec:devandrepostatistics} shows the number of issues completed across repositories and developers, respectively. We find that when developers use AI tools, they implement issues in \PctSpeedupBaselineAbsValue \% \textit{more} time on average (\autoref{fig:horizontal_iceberg}), and nearly all quantiles of observed implementation time see AI-allowed issues taking longer (\autoref{fig:observed_ecdf_scatter}). That is, developers are slower when using AI is allowed. Colloquially, we refer to this result that issues with AI-allowed take longer than issues with AI-disallowed as \textit{slowdown}.

\subsection{Forecasts}

\paragraph*{Developer Forecasts and Post hoc Estimates}
Before developers complete each issue, they forecast how long they expect them to take with and without AI assistance. On average, they forecast speedup of \PctSpeedupForecastAbsValue \%. Interestingly, after the experiment they post-hoc estimate that they were sped-up by \PctSpeedupBackcastAbsValue \% when using AI is allowed---after they used AI assistance, they estimate similar speedup as before, despite the fact that they are in fact slowed down by \PctSpeedupBaselineAbsValue\% (\autoref{fig:horizontal_iceberg}).
\autoref{fig:forecast_vs_observed_means} displays the raw average forecasted and observed implementation times\footnote{The raw percentage difference in implementation times between AI-allowed and AI-disallowed issues is \PctSpeedupRatioEstimator\%, which is larger in absolute value than the \PctSpeedupBaselineAbsValue\% slowdown estimated using the regression specified in \autoref{sec:regression}. This is true because AI-allowed issues ended up being slightly more difficult than AI-disallowed issues after randomization, as measured by the forecasted AI-disallowed times (not statistically significant; see \autoref{tab:balance_and_proportions}). Our regression accounts for this, while this simple ratio estimator does not. See \autoref{fig:subset-alternative-estimators} for results implied by alternative estimators.}.

Despite developers forecasting speedup from AI (while they are slowed down), developer forecasts \emph{are} informative about completion time---the Pearson correlation between the time developers forecast AI-allowed issues taking and the actual time they take is \CorrForecastAndImplementationTimeAIAllowed, and the correlation between the time developers forecast AI-disallowed issues taking and the actual time they take is \CorrForecastAndImplementationTimeNoAI. This suggests that developers are broadly well-calibrated on the relative amount of time that issues will take, but their expectations regarding the usefulness of AI assistance are reversed.

\paragraph*{Expert Forecasts}

Speedup forecasts from \NumForecastsEconomists\ economics experts and \NumForecastsMLProfessionals\ machine learning experts overestimate speedup even more drastically than developers, predicting AI will lead to decreases in implementation time of \PctSpeedupForecastEconomistsAbsValue \% and \PctSpeedupForecastMLProfessionalsAbsValue \%, respectively (\autoref{fig:horizontal_iceberg}). We show distributions of expert forecasts in \autoref{sec:other_analysis_expert_forecasts}.

\subsection{Activity Labels}

\begin{figure}[t]
    \centering
    \includegraphics[width=1\linewidth]{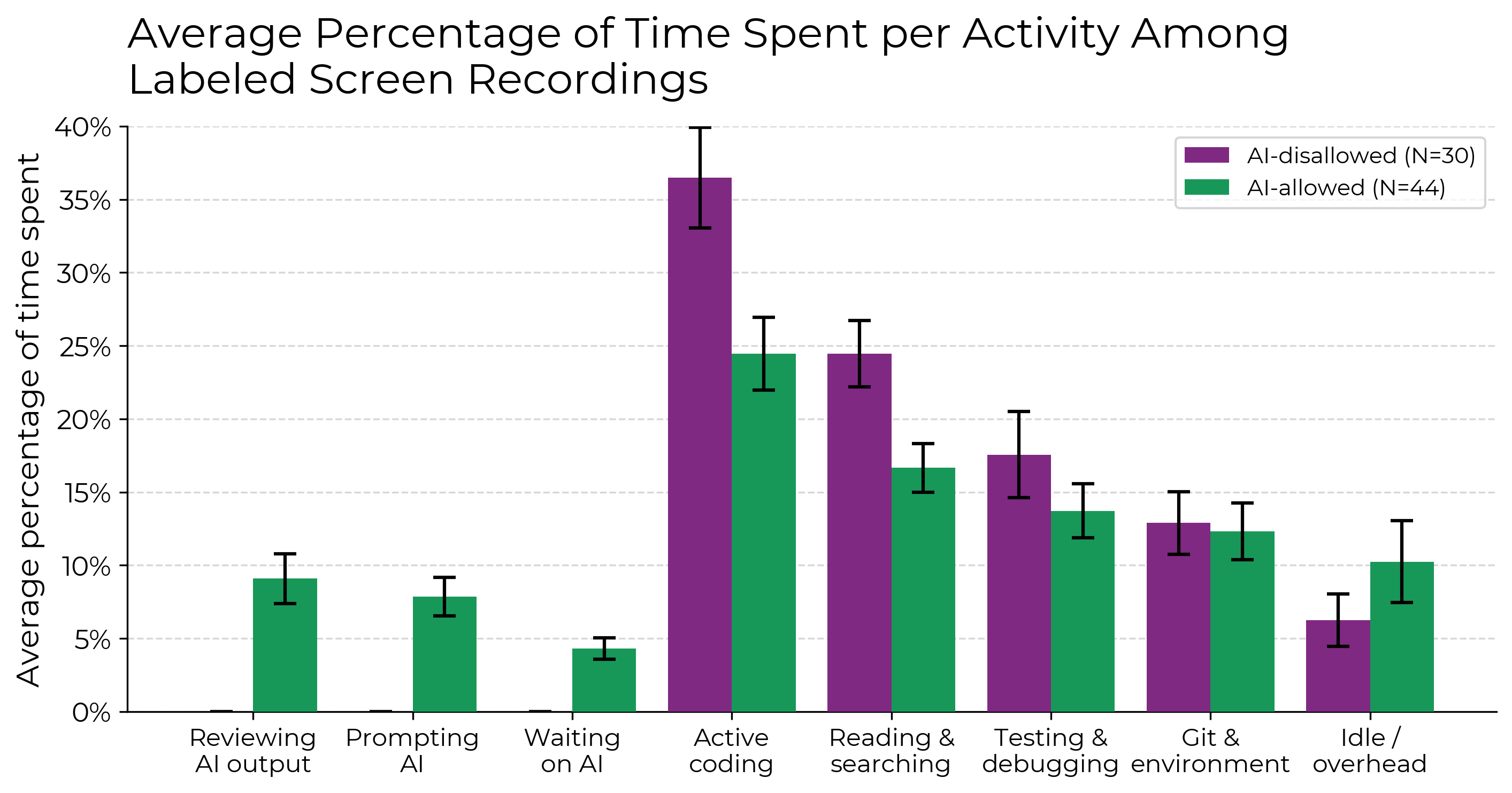}
    \caption{On the subset of labeled screen recordings, when AI is allowed, developers spend less time actively coding and searching for/reading information, and instead spend time prompting AI, waiting on and reviewing AI outputs, and idle. \autoref{fig:loom-high-category-minutes} shows the absolute (average) minutes spent in each category, and \autoref{fig:loom-low-category-percentage} presents these results broken down into 27 fine-grained categories.}
    \label{fig:loom-high-category-percentage}
\end{figure}

On a subset of \NumLoomVideosFiltered\ issues for which we have valid screen recordings, we manually label the activities developers engage in while they work. \autoref{fig:loom-high-category-percentage} shows the percentage of time developers spend for each type of issue (AI-allowed or AI-disallowed). When allowed to use AI, developers spend a smaller proportion of their time actively coding and reading/searching for information. Instead, they spend time reviewing AI outputs, prompting AI systems, and waiting for AI generations. Interestingly, they also spend a somewhat higher proportion of their time idle, where their screen recording doesn't show any activity. \autoref{sec:other_loom_plots} shows the number of minutes spent on average in each category (instead of percentage time spent), as well as the distributions of labels broken down into more fine-grained activities.

\subsection{Factor Analysis}
\label{sec:factor_analysis}

\begin{table}[thbp]
    \centering \small \scriptsize

    \noindent{\footnotesize\textbf{Factors likely to contribute to slowdown}}
    \begin{tcolorbox}[colback=white, colframe=green!70!black, boxrule=0.7pt, boxsep=1pt, left=3pt, right=3pt, top=3pt, bottom=3pt]
    \begin{tabularx}{\textwidth}{ p{4cm} p{0.8cm} X }
        \textbf{Factor} & \textbf{Type} & \textbf{Relevant Observations} \\
        \midrule

        Over-optimism about AI usefulness (\ref{sec:factor-over-optimistic-beliefs-about-ai-usefulness}) & \DirectLossIcon & 
        \begin{tabular}[t]{@{}p{\linewidth}@{}}
        $\bullet$ Developers forecast AI will decrease implementation time by \PctSpeedupForecastAbsValue \% \\
        $\bullet$ Developers post hoc estimate AI decreased implementation time by \PctSpeedupBackcastAbsValue \% \\
        \end{tabular} \\
        \midrule    

        High developer familiarity with repositories (\ref{sec:factor-high-dev-familiarity-with-repositories}) & 
        \RaisesHumanIcon & 
        \begin{tabular}[t]{@{}p{\linewidth}@{}}
            $\bullet$ Developers slowed down more on issues they are more familiar with \\
            $\bullet$ Developers report that their experience makes it difficult for AI to help them \\
            $\bullet$ Developers average \MeanDevExperienceOnReposYears\ years experience and \MeanMaintainerCommits\ commits on repositories \\
        \end{tabular} \\
        \midrule
        
        Large and complex repositories (\ref{sec:factor-large-and-complex-repositories}) & 
        \LimitsAIIcon & 
        \begin{tabular}[t]{@{}p{\linewidth}@{}}
            $\bullet$ Developers report AI performs worse in large and complex environments\\
            $\bullet$ Repositories average \MeanRepoAgeYears\ years old with $>$\MeanRepoLoC\ lines of code\\
        \end{tabular} \\
        \midrule
        
        Low AI reliability (\ref{sec:factor-low-ai-reliability}) & 
        \LimitsAIIcon & 
        \begin{tabular}[t]{@{}p{\linewidth}@{}}
        $\bullet$ Developers accept $<$\PctAcceptedAddedLinesOfCode\% of AI generations \\
        $\bullet$ Majority report making major changes to clean up AI code\\
        $\bullet$ \PctReviewingAIOutputAIAllowed\% of time spent reviewing/cleaning AI outputs \\
        \end{tabular} \\
        \midrule
        
        Implicit repository context (\ref{sec:factor-implicit-repository-context}) & 
        \LimitsAIIcon\ \RaisesHumanIcon & 
        \begin{tabular}[t]{@{}p{\linewidth}@{}}
        $\bullet$ Developers report AI doesn't utilize important tacit knowledge or context \\
        \end{tabular} \\
    \end{tabularx}
    \end{tcolorbox}
    \vspace{1em}

    \noindent{\footnotesize\textbf{Factors with unclear effect on slowdown}}
    \begin{tcolorbox}[colback=white, colframe=yellow!80!orange, boxrule=0.7pt, boxsep=1pt, left=3pt, right=3pt, top=3pt, bottom=3pt] 
    \begin{tabularx}{\textwidth}{ p{4cm} p{1cm} X }
        \textbf{Factor} & \textbf{Type} & \textbf{Relevant Observations} \\
        \midrule

        Experimentally driven overuse of AI (\ref{sec:factor-overuse-of-ai}) & 
        \ExperimentalArtifactIcon & 
        \begin{tabular}[t]{@{}p{\linewidth}@{}}
        $\bullet$ Developers sometimes report overuse due to experiment \\
        $\bullet$ Similar slowdown from developers reporting overuse vs. normal use \\
        \end{tabular} \\
        \midrule

        Unrepresentative task distribution (\ref{sec:factor-unrepresentative-task-distribution}) & 
        \ExperimentalArtifactIcon & 
        \begin{tabular}[t]{@{}p{\linewidth}@{}}
        $\bullet$ Developers report issues are standard but on the shorter side \\
        $\bullet$ Excludes non-programming tasks developers complete in normal work
        \end{tabular} \\
        \midrule
        
        AI increasing issue scope (\ref{sec:factor-scope-creep}) & 
        \ExperimentalArtifactIcon & 
        \begin{tabular}[t]{@{}p{\linewidth}@{}}
        $\bullet$ Developers who report scope creep see \textit{less} slowdown \\
        $\bullet$ Mixed developer reports on AI's impact on scope \\
        $\bullet$ \PctMoreLinesOfCodeInAIAllowedIssuesPerAIDisallowedForecastedHour \% more lines of code per forecasted hour in AI-allowed issues 
        \end{tabular} \\
        \midrule

        Bias from issue completion order (\ref{sec:factor-bias-from-issue-completion-order}) & 
        \ExperimentalArtifactIcon & 
        \begin{tabular}[t]{@{}p{6cm}@{}}
        $\bullet$ Developers decide order post randomization
        \end{tabular} \\
        \midrule

        Sampling bias in developer recruitment (\ref{sec:factor-developer-sampling-bias}) &
        \ExperimentalArtifactIcon &
        \begin{tabular}[t]{@{}p{\linewidth}@{}}
        $\bullet$ Developers who rely heavily on AI may be less likely to participate
        \end{tabular} \\
        \midrule
        
        Trading speed for ease (\ref{sec:factor-productivity-for-leisure}) & 
        \DirectLossIcon & 
        \begin{tabular}[t]{@{}p{6cm}@{}}
        $\bullet$ Some developers report using AI is less effortful \\
        $\bullet$ High developer retention on Cursor \\
        \end{tabular} \\
        \midrule

        Low quality initial pull requests (\ref{sec:factor-low-quality-inital-pull-requests}) & 
        \DirectLossIcon & 
        \begin{tabular}[t]{@{}p{\linewidth}@{}}
        $\bullet$ Minor difference in mean post-review times between conditions \\
        $\bullet$ Qualitatively similar PR quality between conditions \\
        \end{tabular} \\
        \midrule

        Below-average use of AI tools (\ref{sec:factor-below-average-use-of-ai-tools}) & 
        \LimitsAIIcon & 
        \begin{tabular}[t]{@{}p{\linewidth}@{}}
        $\bullet$ Similar slowdown for developers with prior Cursor experience \\
        $\bullet$ No clear learning effect across first 30-50 hours of Cursor usage \\
        $\bullet$ Developers appear qualitatively in distribution for Cursor Pro users
        \end{tabular} \\
        \midrule

        AI generation latency (\ref{sec:factor-ai-generation-latency}) & \LimitsAIIcon & 
        \begin{tabular}[t]{@{}p{\linewidth}@{}}
        $\bullet$ Mixed developer reports that waiting on AI generations was important \\
        $\bullet$ Developers spend \PctWaitingOnAIAIAllowed \% of time waiting on AI generations 
        \end{tabular} \\
        \midrule
        
        Suboptimal elicitation (\ref{sec:factor-suboptimal-elicitation}) & 
        \LimitsAIIcon & 
        \begin{tabular}[t]{@{}p{\linewidth}@{}}
        $\bullet$ Developers use Cursor agents/chat in majority of AI-allowed issues \\
        $\bullet$ Developers sample few tokens from models \\
        $\bullet$ But existing literature finding positive speedup also uses few tokens \\
        $\bullet$ Unused elicitation strategies could improve AI reliability
        \end{tabular} \\

    \end{tabularx}
    \end{tcolorbox}
    \vspace{1em}

    \noindent{\footnotesize\textbf{Factors unlikely to contribute to slowdown}}
    \begin{tcolorbox}[colback=white, colframe=red!90!black, boxrule=0.7pt, boxsep=1pt, left=3pt, right=3pt, top=3pt, bottom=3pt] 
    \begin{tabularx}{\textwidth}{ p{4cm} p{1cm} X }
        \textbf{Factor} & \textbf{Type} & \textbf{Relevant Observations} \\
        \midrule

        Unfamiliar development environment (\ref{sec:factor-unfamiliar-dev-environment}) & 
        \ExperimentalArtifactIcon & 
        \begin{tabular}[t]{@{}p{\linewidth}@{}}
        $\bullet$ Most developers use comparable IDEs between treatment conditions \\
        $\bullet$ These developers still see slowdown of \PctSpeedupUsedComparableIDEs\% \\
        $\bullet$ No clear learning effects across first 30-50 hours of Cursor usage
        \end{tabular} \\
        \midrule

        Cheating or under-use of AI (\ref{sec:factor-cheating-or-underuse-of-ai}) & 
        \ExperimentalArtifactIcon & 
        \begin{tabular}[t]{@{}p{\linewidth}@{}}
        $\bullet$ AI used in all but \PctTreatmentNotTakenRate\% of allowed cases with labeled screen recordings\\
        $\bullet$ Only \NumLoomVideosCheating\ cheating instances in \NumLoomVideosInputNoAI\ screen recordings
        \end{tabular} \\
        \midrule
        
        Issue dropout (\ref{sec:factor-issue-dropout}) & 
        \ExperimentalArtifactIcon & 
        \begin{tabular}[t]{@{}p{\linewidth}@{}}
        $\bullet$ Developers with no accidental dropout see similar slowdown \\
        $\bullet$ Issues dropped intentionally are qualitatively unbiased\\
        \end{tabular} \\
        \midrule
                        
        Non-robust outcome measure (\ref{sec:factor-specific-outcome-measure}) & 
        \ExperimentalArtifactIcon & 
        \begin{tabular}[t]{@{}p{\linewidth}@{}}
        $\bullet$ Alternative outcome measures yield similar slowdown \\
        \end{tabular} \\
        \midrule
        
        Non-robust estimator (\ref{sec:factor-specific-estimator}) & 
        \ExperimentalArtifactIcon & 
        \begin{tabular}[t]{@{}p{\linewidth}@{}}
        $\bullet$ Alternative estimators yield similar slowdown \\
        \end{tabular} \\
        \midrule

        Non-frontier model usage (\ref{sec:factor-non-frontier-model-usage}) & 
        \LimitsAIIcon & 
        \begin{tabular}[t]{@{}p{\linewidth}@{}}
        $\bullet$ Developers mostly use (at the time) frontier models \\
        \end{tabular} \\

    \end{tabularx}
    \end{tcolorbox}
    \vspace{1em}

    \caption{Summary of factors that may \textnormal{a priori} explain or contribute to slowdown, grouped by the state of evidence for or against their impact on the slowdown effect. \RaisesHumanIcon\ are factors that raise human performance, \LimitsAIIcon\ are factors that limit AI performance, \ExperimentalArtifactIcon\ are experimental artifacts that may bias/confound results, and \DirectLossIcon\ are factors that directly contribute to productivity losses.}
    \label{tab:all_factors_table}

\end{table}

Given the surprising nature of this result, we investigate \NumFactorsTotal\ potential contributing factors that may contribute to developers spending more time on tasks when AI usage is allowed. We group these factors into four categories:
\begin{itemize}
    \item \textbf{Direct productivity loss} (\DirectLossIcon): mechanisms by which the use of AI tools actively slows down development.
    \item \textbf{Experimental artifact} (\ExperimentalArtifactIcon): confounders from our experimental setup or procedures that may introduce biases, or limit the external validity.
    \item \textbf{Raises developer performance} (\RaisesHumanIcon): attributes of the issues, repositories, or setting that improve developer ability relative to AI.
    \item \textbf{Limits AI performance} (\LimitsAIIcon): attributes of the issues, repositories, or AI/environment tooling that diminish AI's effectiveness relative to developers.
\end{itemize}

Using entry and exit surveys, screen recordings, developer interviews, and subset analyses we find qualitative and quantitative evidence that \NumFactorsThatLikelyContribute\
of the \NumFactorsTotal\ factors contribute to slowdown, we find mixed/unclear/no evidence that \NumFactorsThatAreUnclear\ of the factors contribute to slowdown, and we find evidence against \NumFactorsThatDoNotContribute\ of the factors contributing. However, we strongly caution against over-indexing on the basis of any individual pieces of evidence, as we are not powered for statistically significant multiple comparisons when subsetting our data. This analysis is intended to provide speculative, suggestive evidence about the mechanisms behind slowdown. \autoref{sec:allfactors} discusses the evidence for/against each factor in \autoref{tab:all_factors_table}.

\section{Discussion}
\label{sec:Discussion}

We provide evidence that recent AI systems slow down experienced open-source developers with moderate AI experience completing real issues on large, popular repositories they are highly familiar with. 
This observed slowdown serves as some evidence that AI capabilities in the wild may be lower than results on commonly used benchmarks may suggest.

Furthermore, we show that both experts and developers drastically overestimate the usefulness of AI on developer productivity, \textit{even after they have spent many hours using the tools}. 
This underscores the importance of conducting field experiments with robust outcome measures, compared to relying solely on expert forecasts or developer surveys.

\subsection{Key Caveats}

\paragraph*{Setting-specific factors}
We caution readers against overgeneralizing on the basis of our results. The slowdown we observe does \textit{not} imply that current AI tools do not often improve developer's productivity---we find evidence that the high developer familiarity with repositories and the size and maturity of the repositories both contribute to the observed slowdown, and these factors do not apply in many software development settings. For example, our results are consistent with small greenfield projects or development in unfamiliar codebases seeing substantial speedup from AI assistance.

\paragraph*{AI-specific factors}
We expect that AI systems that have higher fundamental reliability, lower latency, and/or are better elicited (e.g. via more inference compute/tokens, more skilled prompting/scaffolding, or explicit fine-tuning on repositories) could speed up developers in our setting (i.e. experienced open-source developers on large repositories).

\paragraph*{Agents can make meaningful progress on issues}
We have preliminary evidence (forthcoming) that fully autonomous AI agents using Claude 3.7 Sonnet can often correctly implement the core functionality of issues on several repositories that are included in our study, although they fail to fully satisfy all requirements (typically leaving out important documentation, failing linting/styling rules, and leaving out key unit or integration tests). This represents immense progress relative to the state of AI just 1-2 years ago, and if progress continues apace (which is \textit{a priori} at least plausible, although not guaranteed), we may soon see significant speedup in this setting. 

\section{Acknowledgments}

We thank the open-source developers who participated in this study. Your hard work, diligent record keeping, and excellent software made it a pleasure to work with you. Thanks to Aaron Diamond-Reivich, Alan Akbik, Domenic Denicola, Dens Sumesh, Jaden Fiotto-Kaufman, João Gante, Liam DeVoe, Matthew Pickering, Muhammad Haris, Philipp Burckhardt, Quentin Anthony, Ruben Bloom, Sam Derbyshire, and other participating developers.

We thank the following reviewers for feedback on the experimental design and paper drafts: Adrien Ecoffet, Alexander Barry, Ali Merali, Ajeya Cotra, Andres Campero, Andrey Fradkin, Basil Halperin, Cozmin Ududec, Eli Lifland, Ernest Davis, Gregory Sun, Hjalmar Wijk, James Requeima, Jide Alaga, Josh Jacobson, Lawrence Chan, Megan Kinniment, Michael Sklar, Neev Parikh, Rif A. Saurous, Rob Miles, Ryan Greenblatt, Seraphina Nix, Sydney Von Arx, Thomas Kwa, and Tom Cunningham.

We thank the following individuals for help with data collection: Adam Hanson, Amy Ngo, Chris Canal, Jebastin Nadar, Luis Slyfield, and Martin Milbradt.

We thank Sami Jawar and Thomas Broadley for technical support throughout the project.

We thank the following for their operational support through the project: Bhaskar Chaturvedi, Emma Abele, Kit Harris, Kris Chari, Kyle Scott, Rebecca Baron, and Rae She.

The authors thank Stephanie He for graphic design contributions.

The authors especially thank Aron Lajko, Chris Painter, Jasmine Dhaliwal, and Steve Newman for close review, feedback, and support throughout the project.

\newpage 

\bibliographystyle{unsrtnat}
\bibliography{main}

\newpage
\appendix

\section{Author contributions}

\textbf{Joel Becker} and \textbf{Nate Rush} designed, implemented, and led the project.

\textbf{Beth Barnes} gave feedback and guidance on the project.

\textbf{David Rein} contributed substantially to the writing and framing of the results.

\section{Extended Discussion}
\label{sec:extended_discussion}

\begin{table}[h]
    \centering
    \renewcommand{\arraystretch}{1.3}
    \begin{tabular}{p{6cm} p{7cm}}
        \toprule
        \textbf{We do \textit{not} provide evidence that:} & \textbf{Clarification} \\
        \midrule
        
        AI systems do not currently speed up many or most software developers & 
        We do not claim that our developers or repositories represent a majority or plurality of software development work \\
        \midrule
        
        AI systems do not speed up individuals or groups in domains other than software development & 
        We only study software development \\
        \midrule
        
        AI systems in the near future will not speed up developers in our exact setting & 
        Progress is difficult to predict, and there has been substantial AI progress over the past five years \cite{kwa2025measuringai} \\
        \midrule
        
        There are not ways of using existing AI systems more effectively to achieve positive speedup in our exact setting & 
        Cursor does not sample many tokens from LLMs, it may not use optimal prompting/scaffolding, and domain/repository-specific training/finetuning/few-shot learning could yield positive speedup \\
        \midrule

        Developers with much more experience using AI systems wouldn't see speedup & While our developers appear to use AI tools competently, it's plausible that much more experience could yield speedup. See \autoref{sec:factor-below-average-use-of-ai-tools} for discussion/analysis of the impact of developers' AI experience on slowdown. \\
        \midrule

        These developers are not sped up on any tasks & We estimate that some developers experience speedup from AI---see \autoref{fig:per-developer-speedup} for per-developer speedup estimates. \\

        \bottomrule
    \end{tabular}
    \vspace{1em}
    \caption{Potential misconceptions about our work: what our evidence does \textit{not} demonstrate about AI and developer productivity.}
    \label{tab:potential_misconceptions}
\end{table}

\paragraph*{Potential Misreadings of Results}
Given both the importance of understanding AI capabilities/risks, and the diversity of perspectives on these topics, we feel it's important to forestall potential misunderstandings or over-generalizations of our results. We list claims that we do \textbf{\textit{not}} provide evidence for in \autoref{tab:potential_misconceptions}.

\paragraph*{Literature Comparison}
\label{sec:literature_comparison}

\begin{table}[htbp]
    \centering \footnotesize
    \renewcommand{\arraystretch}{1.3}
    \begin{tabular}{l l p{1.5cm} p{1.5cm} p{2cm} p{1.6cm}}
        \toprule
        \textbf{Paper} & \textbf{Result} & \textbf{AI $\geq$} \textbf{GPT-4?} & \textbf{Non-synthetic tasks} & \textbf{Experienced, high-familiarity} \textbf{devs} & \textbf{Fixed outcome measure}\\
        \midrule
        
        \citet{peng2023impactaideveloperproductivity} &
        $\uparrow$ 56\% faster & \xmark & \xmark &\xmark & \cmark\\
        
        \citet{10.1145/3661145} &
        $\uparrow$ 65\% faster & \xmark & \xmark & \xmark &\cmark\\
        
        \citet{cui2025effectsgenerativeai} &
        $\uparrow$ 26\% output & \xmark& \cmark &\cmark & \xmark\\
        
        \citet{paradis2024doesaiimpactdevelopment} &
        $\uparrow$ 21\% faster & \textbf{?} & \xmark & \xmark & \cmark\\
        
        \citet{gambacorta2024generative} &
        $\uparrow$ 55\% output &\xmark & \cmark & \cmark & \xmark \\
        
        \citet{yeverechyahu2025impactllmopensource} &
        $\uparrow$ 37\% output & \xmark & \cmark &\cmark & \xmark\\
        
        \textbf{Our study}&\textbf{$\downarrow$ \PctSpeedupBaselineAbsValue\% slower} & \textbf{\cmark} & \cmark & \cmark & \cmark \\
        
        \bottomrule
    \end{tabular}
    \vspace{1em}
    \caption{Overview of key studies measuring the impact of AI tools on software development productivity. \citet{paradis2024doesaiimpactdevelopment} does not report the model(s) used internally.}
    \label{tab:lit_comparison}
\end{table}

\autoref{tab:lit_comparison} compares relevant studies that measure the impact of AI tools on software developer productivity, along key dimensions that distinguish our results from prior work. Other relevant literature typically does not investigate the impact of AI systems more capable than GPT-4 (which as of mid-2025 isn't itself close to the frontier), and does not (in any individual study) analyze tasks in the wild, with experienced developers, using outcome measures that correspond directly with productivity and that are fixed before treatment assignment.

\section{Factor Analysis}
\label{sec:allfactors}

\subsection{Factors driving slowdown}
\label{sec:factorsdriving}

We observe \NumFactorsThatLikelyContribute\ setting-specific factors that contribute to our observed slowdown, summarized in \autoref{tab:all_factors_table}.

\subsubsection{Over-optimism about AI usefulness \textnormal{\textit{(Direct productivity loss)}}}
\label{sec:factor-over-optimistic-beliefs-about-ai-usefulness}

On AI-allowed issues, developers are not \textit{required} to use AI tools---they are instructed to use AI assistance as much or as little as they would find helpful. Given this, we might expect the percentage change in issue completion time due to AI to be lower-bounded at 0\%---if a developer is aiming to be as productive as possible, and they see that they are being slowed down by an AI tool, we'd expect them to not continue using the AI tool.

However, developers have an overoptimistic picture of how AI affects their productivity, both before and after they complete issues. Before starting issues, developers forecast that using AI will reduce completion time by \PctSpeedupForecastAbsValue \%. After completing issues, developers estimate that using AI reduced their issue completion time by \PctSpeedupBackcastAbsValue \% on average. This overoptimistic view may lead developers to overuse AI assistance, despite its negative effect on their productivity.

\subsubsection{High developer familiarity with repositories \textnormal{\textit{(Raises developer performance)}}}
\label{sec:factor-high-dev-familiarity-with-repositories}

We might expect that AI assistance is less helpful on tasks where developers are already highly skilled, for example, because they have completed similar tasks previously, or because they have all of the knowledge and skills required already.

Before developers complete each issue, they score how much ``Prior Task Exposure'', they have from 1 - 5, where 1 means ``Never done this type of task before'', and 5 means ``I am an expert at this type of task''. We consider ratings of less than or equal to 3 to mean they have low prior task exposure, and ratings higher than 3 to imply high prior task exposure (we selected this cutoff to achieve roughly balanced ratings). 

\begin{figure}[t]
    \centering
    \includegraphics[width=1\linewidth]{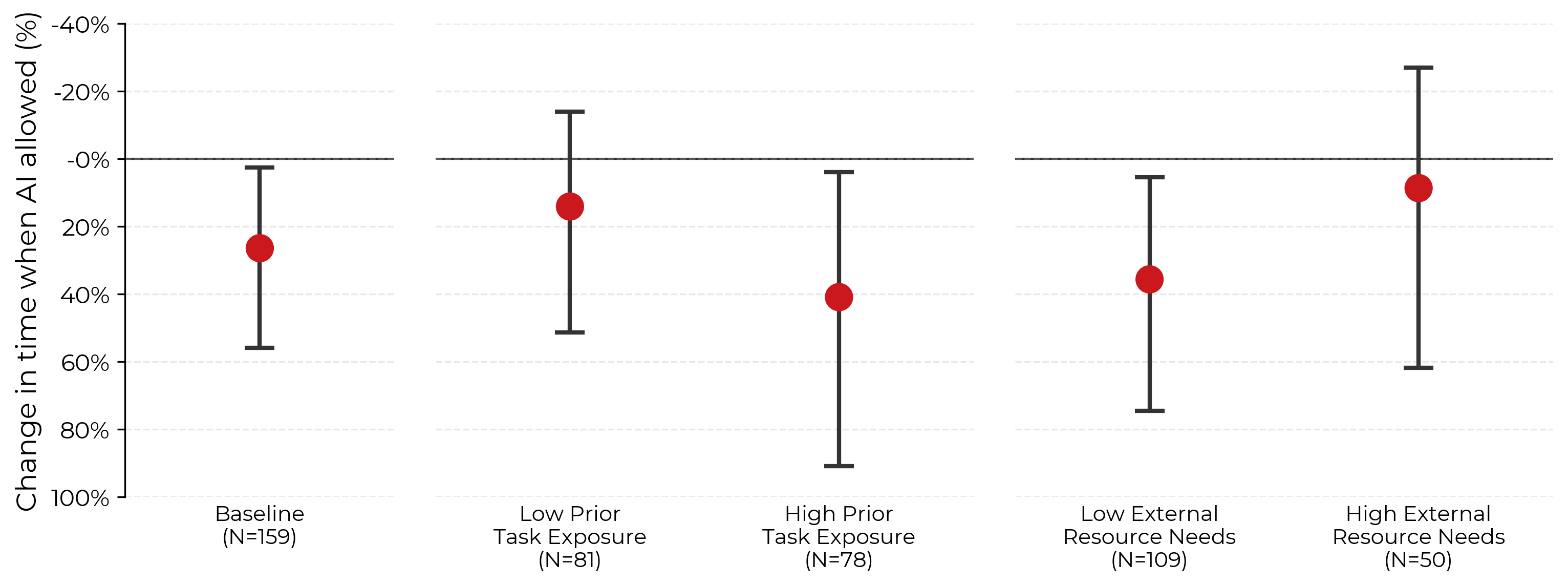}
    \caption{Developers are slowed down more on issues where they self-report having significant prior task exposure, and on issues where they self-report having low external resource needs (e.g. documentation, reference materials). We only collected this data for the latter half of issues completed in the study---this is why we have a smaller number of issues for our baseline slowdown estimate. See \autoref{sec:heterogeneous_treatment_effect_estimation} for details on how we estimate heterogeneous treatment effects.}
    \label{fig:subset-issue-factors}
\end{figure}

\autoref{fig:subset-issue-factors} gives moderate evidence that developers are slowed down more on on issues where they have high prior task exposure.

Developers also rate (before issues are randomized and completed) ``External Resource Needs'' from 1 - 3, where 1 means ``I need extensive documentation / reference material / googling to complete this task'' and 3 means ``I could complete this task entirely from memory and experience.'' We consider ratings of less than or equal to 2 to mean they have low external resource needs, and ratings higher than 2 to imply high external resource needs (we selected this cutoff to achieve roughly balanced ratings).
\autoref{fig:subset-issue-factors} presents moderate evidence that developers are slowed down more on issues where they need fewer external resources. 

Qualitatively, developers note that AI is particularly helpful when working on unfamiliar issues, and less helpful when working on familiar ones. 

One developer working with unfamiliar datasets found that AI was helpful in answering \devquote{general questions about e.g. EICAR.} Another developer noted that Cursor was \devquote{super helpful in figuring out how to write a [frontend test.] I didn't know how to do this before and on my third time asking cursor for help with it, it came up with this solution.} Another developer, working with Git hooks, noted that \devquote{Given that it was my first time with Git hooks, without AI the implementation would've taken me [3 additional hours].} Sometimes, portions of one’s own codebase can be as unknown as a new API. One developer noted that \devquote{cursor found a helper test function that I didn't even know existed when I asked it how we tested deprecations.} 

On the other hand, developers note that AI is much less helpful on issues where they are expert. One developer notes that \devquote{if I am the dedicated maintainer of a very specialized part of the codebase, there is no way agent mode can do better than me.}

Broadly, we present moderate evidence that on the issues in our study, developers are slowed down more when they have high prior task exposure and lower external resource needs. We hypothesize that analogously, AI helps our developers less compared to existing literature \cite{peng2023impactaideveloperproductivity, paradis2024doesaiimpactdevelopment} because our developers have substantially more experience on their respective repositories (\MeanDevExperienceOnReposYears~ years and \MeanMaintainerCommits~ commits on average). This would be consistent with the experience/familiarity effects observed in \citet{noy2023experimental, cui2025effectsgenerativeai}.

\subsubsection{Large and complex repositories \textnormal{\textit{(Limits AI performance)}}}
\label{sec:factor-large-and-complex-repositories}

Developers qualitatively note LLM tooling performs worse in more complex environments. One developer says \devquote{it also made some weird changes in other parts of the code that cost me time to find and remove [...] My feeling is the refactoring necessary for this PR was ``too big'' [and genAI] introduced as many errors as it fixed.} Another developer comments that one prompt \devquote{failed to properly apply the edits and started editing random other parts of the file,} and that these failures seemed to be heavily related to \devquote{the size of a single file it is attempting to perform edits on.}.

We hypothesize that analogously to these size and complexity effects within our study, AI broadly helps our developers less compared to existing randomized controlled trials (RCTs) measuring speedup from AI tools because of the overall size and complexity of the repositories included---participating repositories are on average about \MeanRepoAgeYears\ years old and contain $>$\MeanRepoLoC\ lines of code, compared to the more greenfield projects completed in \citet{peng2023impactaideveloperproductivity}, \citet{paradis2024doesaiimpactdevelopment}, and \citet{10.1145/3661145}. This would be consistent with existing literature studying the effects of environment complexity on AI performance \cite{wijk2025rebenchevaluatingfrontierai, jimenez2024swebench, liu2023lostmiddlelanguagemodels}.

\subsubsection{Low AI reliability \textnormal{\textit{(Limits AI performance)}}}
\label{sec:factor-low-ai-reliability}

When using Cursor, developers accept $<$\PctAcceptedAddedLinesOfCode \% of the generations.\footnote{We were not able to collect this data from \NumDevelopersWithNoCursorData\ developers who used their own pre-existing Cursor Pro subscriptions, so this statistic excludes them.} When developers do not accept generations, we observe a mix of reattempting with different prompts, and giving up (i.e. reverting the proposed changes).

This relatively low reliability qualitatively results in significant wasted time, as developers often spend time reviewing, testing, or modifying AI generated code before they decide to reject it. One developer notes that he \devquote{wasted at least an hour first trying to [solve a specific issue] with AI} before eventually reverting all code changes and just implementing it without AI assistance.

Developers further note that even when they accept AI generations, they spend a significant amount of time reviewing and editing AI generated code to ensure it meets their high standards. \PctDevelopersExitSurveyReadEveryLineAICode\% report that they read every line of AI generated code, and \PctDevelopersExitSurveyEditMajorChangesAICode\%  of developers report that they often need to make major changes to clean up AI code---when asked, 100\% developers report needing to modify AI generated code. One developer comments that AI generated code resulted in \devquote{more to fix up, in terms of code conventions, code style}. Another developer says \devquote{I accept [AI generated code] [...] then I feel like I do a lot of simplifying its code.} 

Quantitatively, on the subset of \NumLoomVideosFilteredAIAllowed\ issues with valid labeled screen recordings, we find that when AI is allowed, developers spend approximately \PctReviewingAIOutputAIAllowed\% of their time reviewing and cleaning AI generated outputs when working with AI.

\subsubsection{Implicit repository context \textnormal{\textit{(Limits AI performance, Raises developer performance)}}}
\label{sec:factor-implicit-repository-context}

In software development, developers often rely on their own undocumented knowledge of the codebase to assist design and implementation decisions. In our study, developers often note that AIs lack this tacit codebase knowledge, resulting in less useful AI outputs. One developer notes that AI often acts like a new contributor to the repository, and that \devquote{AI doesn’t pick the right location to make the edits.} 
Another developer notes that while \devquote{we [..] know the data that will interact with the code, but the model doesn't know the data. It doesn't know we need to take care of this weird case of backwards compatibility and [thus] keep this specific line. And this is very hard to give as [context to the model].}.

We hypothesize that the size and maturity of the included repositories increases the amount of tacit knowledge that experienced developers rely on when completing their work---because AI systems may have less access to this knowledge, it may be more difficult for them to assist experienced developers on these issues.


\subsection{Factors with an unclear effect on slowdown}
\label{sec:factorsuncleardriving}

We consider \NumFactorsThatAreUnclear\ factors that have mixed/unclear directional effect on speedup, including effects for which we have no evidence in either direction.

\subsubsection{Experimentally driven overuse of AI \textnormal{\textit{(Experimental artifact)}}}
\label{sec:factor-overuse-of-ai}

As this study is not blinded, developers' awareness of the study may have changed their behavior.

\begin{figure}[t]
    \centering
    \includegraphics[width=1\linewidth]{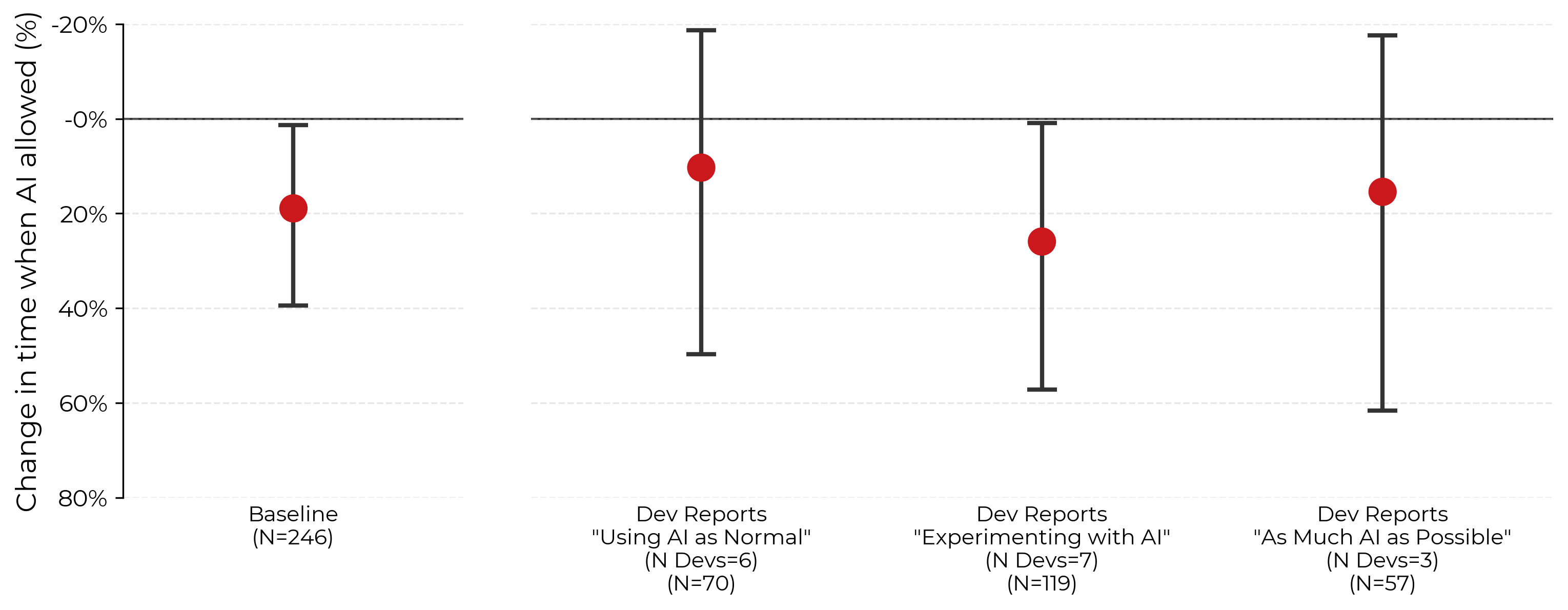}
    \caption{Developers who report that they were experimenting with AI or using AI as much as possible see greater slowdown than developers who report using AI as they normally would. See \autoref{sec:heterogeneous_treatment_effect_estimation} for details on how we estimate heterogeneous treatment effects.}
    \label{fig:subset-ai-usage-patterns}
\end{figure}

Developers were instructed to use AI to whatever degree they thought would make them most productive. After the study's conclusion, we ask developers which best describes their AI usage: ``using AI as normal'', ``experimenting with AI'' or  ``using as much AI as possible.'' \autoref{fig:subset-ai-usage-patterns} doesn't show a clear effect when estimating speedup broken down by these labels. 

However, qualitatively, several developers reported that their participation in the experiment sometimes led to them overusing AI in ways that were unproductive.

\subsubsection{Unrepresentative task distribution \textnormal{\textit{(Experimental artifact)}}}
\label{sec:factor-unrepresentative-task-distribution}

The issues are intended to be as similar as possible to those that would have been worked on if this study never took place. Developers who completed the study noted that issues were  \devquote{completely standard} and that \devquote{there’s nothing I did for this that I wouldn’t have done otherwise.}, and manual review of issues confirm that the issues represent normal work on each repository.

However, the issues completed during this study do not capture all contributions developers make to these repositories. As developers were asked to break larger issues into $\leq 2$ hour issues if possible, developers note that issues were smaller than average than their normal work. One developer comments \devquote{It was the same set of tasks, but sampled from the small end.} Furthermore, developers make other types of contributions to these repositories---the work of resolving issues does not capture PR review, or design discussions, for example.

These selection pressures may have biased issues to be better scoped and more clearly defined than the average work that developers complete on repositories. Existing literature would suggest that this means AIs would perform better on these issues, which we might expect to speed up developers (in contrast to our observed slowdown). However, better scoped issues may also be issues where these expert developers perform better as well, making AIs less relatively useful, so the net effect of this selection is unclear.

\subsubsection{AI increasing issue scope \textnormal{\textit{(Experimental artifact)}}}
\label{sec:factor-scope-creep}

A key design decision for our study is that issues are defined \textit{before} they are randomized to AI-allowed or AI-disallowed groups, which helps avoid confounding effects on the outcome measure (in our case, the time issues take to complete). However, issues vary in how precisely their scope is defined, so developers often have some flexibility with what they implement for each issue. This raises a concern for measuring the impact of AI assistance on developer productivity---if developers expand the scope of their work when using AI tools, even if those issues take longer their productivity might be similar to AI-disallowed issues (because they are getting more done).

\begin{figure}[t]
    \centering
    \includegraphics[width=1\linewidth]{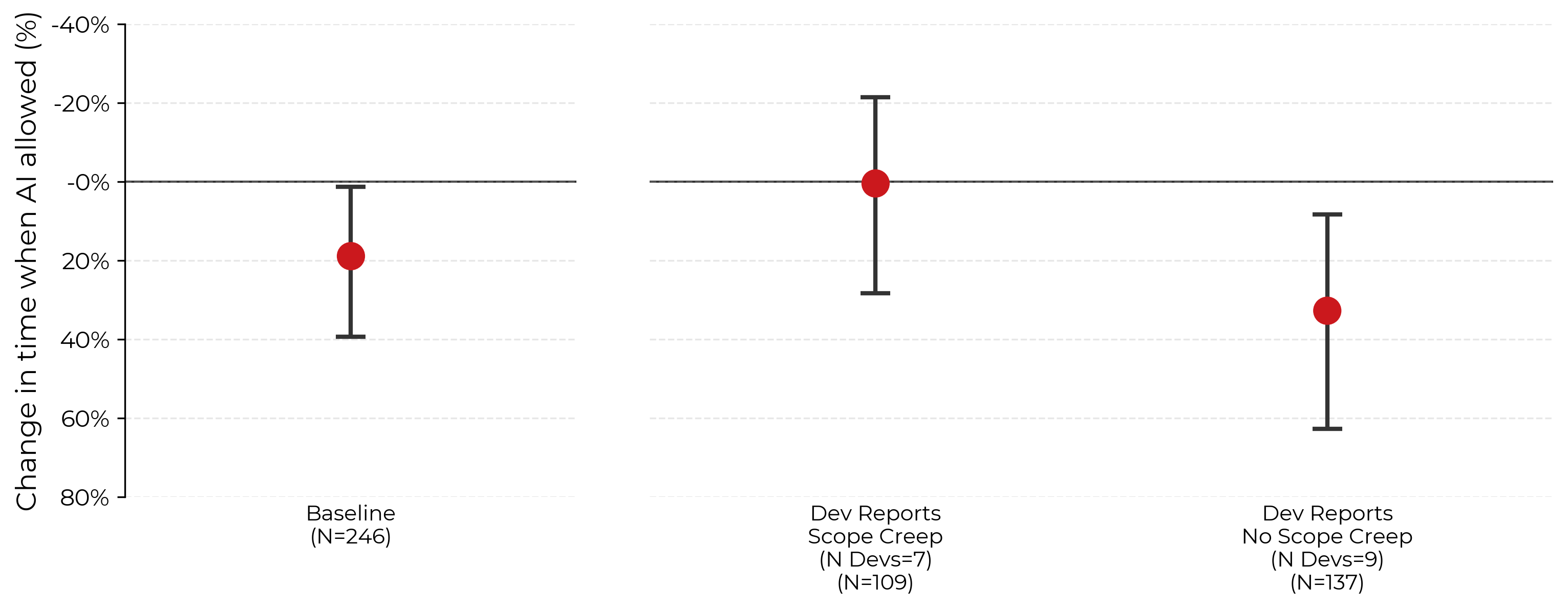}
    \caption{Speedup on issues broken down by whether the developer completing the issue reports scope creep when using AI. See \autoref{sec:heterogeneous_treatment_effect_estimation} for details on how we estimate heterogeneous treatment effects.}
    \label{fig:subset-scope-creep}
\end{figure}

We survey developers after the study period has ended, and ask if on average, they believed that they experienced scope creep when working on AI-allowed issues. \autoref{fig:subset-scope-creep} shows percentage change in issue completion time due to AI broken down by responses to this question---interestingly, we see that developers who report scope creep on AI-allowed issues are actually slowed down \textit{less} than developers who don't report experiencing scope creep. This is evidence against the hypothesis that slowdown is caused by developers increasing the scope of AI-allowed issues, however, it relies on developer self-reports of scope creep, which may be unreliable or subjective.

Qualitatively, some developers self-report that AI does not change their implementations. One developer notes that due to his experience and knowing how changes should be made, \devquote{the scope is not variable.} Another developer notes that AI would not change his approach to an issue, as \devquote{AI was much more suggestable than I was.} 

On the other hand, some developers note that the scope of issues sometimes changes as a result of AI. One developer comments \devquote{If I didn't have AI, I probably [...] not gone so ham on nailing the correct Python tooling.} Another developer notes \devquote{some of [code changes] were a little tedious, and so I am like is this going to [be worth it without AI], but with AI [I make the changes].}

Quantitatively, we observe that developers write \PctMoreLinesOfCodeInAIAllowedIssuesPerAIDisallowedForecastedHour \% (\textit{p = \PValuePctMoreLinesOfCodeInAIAllowedIssuesPerAIDisallowedForecastedHour}) more code per hour of AI-disallowed forecasts, on issues where AI is allowed. This is not statistically significant, and as discussed in \autoref{sec:introduction}, an increase in the number of lines of code does not necessarily correspond to greater productivity or a larger scope, because code can be more verbose but functionally equivalent. Furthermore, this metric contains additional noise because files can be programatically generated by automated systems, and we're unable to collect the exact lines of code written by the human (or human plus AI).

\subsubsection{Bias from issue completion order \textnormal{\textit{(Experimental artifact)}}}
\label{sec:factor-bias-from-issue-completion-order}

Each developer typically has several issues that are randomized in batches, and can then choose the order they complete these issues. This could cause a bias in completion times. For example, if developers prefer to complete AI-disallowed issues when they have more energy, they may choose to do do AI-disallowed issues first more often than AI-allowed issues, which could cause AI-allowed issues to take longer even if AI would have decreased the implementation time without this ordering effect. 

While we receive no qualitative reports from developers that they prioritize issues in this way, we do not have access to developers entire workflows, so we cannot fully rule out this effect.

\subsubsection{Sampling bias in developer recruitment \textnormal{\textit{(Experimental artifact)}}}
\label{sec:factor-developer-sampling-bias}

While we attempt to recruit a representative sample of experienced, open-source developers who contribute to large, mature repositories, it's possible that there is systematic bias in which developers agree to participate. 

For example, it may be the case that developers who rely heavily on AI tools for their work may be less likely to participate, because they would be concerned/unhappy with losing access to AI on 50\% of their tasks. While this is not surfaced in any conversations/discussions with developers during the recruitment process, it's possible that the developers who ended up participating may be less likely to use AI effectively for this reason.\footnote{After the study's conclusion, one developer noted that they were unlikely to participate in future iterations of the study, because they didn't want to be assigned to not use AI tooling on 50\% of their work.} See \autoref{sec:factor-below-average-use-of-ai-tools} for discussion of developers' prior experience with AI tools.

\subsubsection{Trading speed for ease \textnormal{\textit{(Direct productivity loss)}}}
\label{sec:factor-productivity-for-leisure}

One explanation for developers continuing to use AI assistance throughout the study despite it decreasing their productivity is that they are trading productivity for ease---perhaps the development experience when using AI tools like Cursor is so much cognitively easier that developers choose to use it even though they are slower on average. 
We qualitatively receive mixed reports from developers regarding how effortful work with AI tools is relative to work without them. One developer notes that completing issues when using AI \devquote{felt like less effort.} Another developer noted that he 
\devquote{didn’t really think about some of the problems} when using AI. However, another developer noted that there was \devquote{no effort or approach difference [with AI].}

Quantitatively, we track whether developers continue using Cursor after the experiment period ends, and find that \PctDevelopersUsedCursorAfterExperiment\% of developers continue using it after the study period has ended.
This impressive retention rate suggests that developers are getting \textit{some} significant value from using Cursor, and it seems unlikely this is solely a result of miscalibration on their productivity.

\subsubsection{Low quality initial pull requests \textnormal{\textit{(Direct productivity loss)}}}
\label{sec:factor-low-quality-inital-pull-requests}

One possible explanation for AI-allowed issues taking longer than AI-disallowed issues could be that developers submit lower quality pull requests when using AI (e.g. because they do not fully review or fix lower quality AI-generated code), and then have to spend extra time after PR review making improvements/fixes.

We observe a statistically insignificant difference in the mean post-review implementation time (\MeanReviewTimeMinutesNoAI\ and \MeanReviewTimeMinutesAIAllowed\ for AI-disallowed and AI-allowed issues respectively).

However, qualitatively, AI-allowed PRs do not appear to be of lower quality. Because developers have typically spent many years contributing to their respective repositories, they anticipate the high quality PR review standards; \PctDevelopersExitSurveySubmitHighQualityPRs\% of developers report that they ``only submit high quality PRs''.

\subsubsection{Below-average use of AI tools \textnormal{\textit{(Limits AI performance)}}}
\label{sec:factor-below-average-use-of-ai-tools}

Although all developers have used AI tools previously (most have used LLMs for tens to hundreds of hours), only \PctDevelopersPreviouslyUsedCursor\% of developers have prior experience with Cursor. \textit{A priori}, we could imagine significant learning effects for these tools, such that individuals with experience using these tools may be slowed down less than individuals without this experience.

\begin{figure}[t]
    \centering
    \includegraphics[width=1\linewidth]{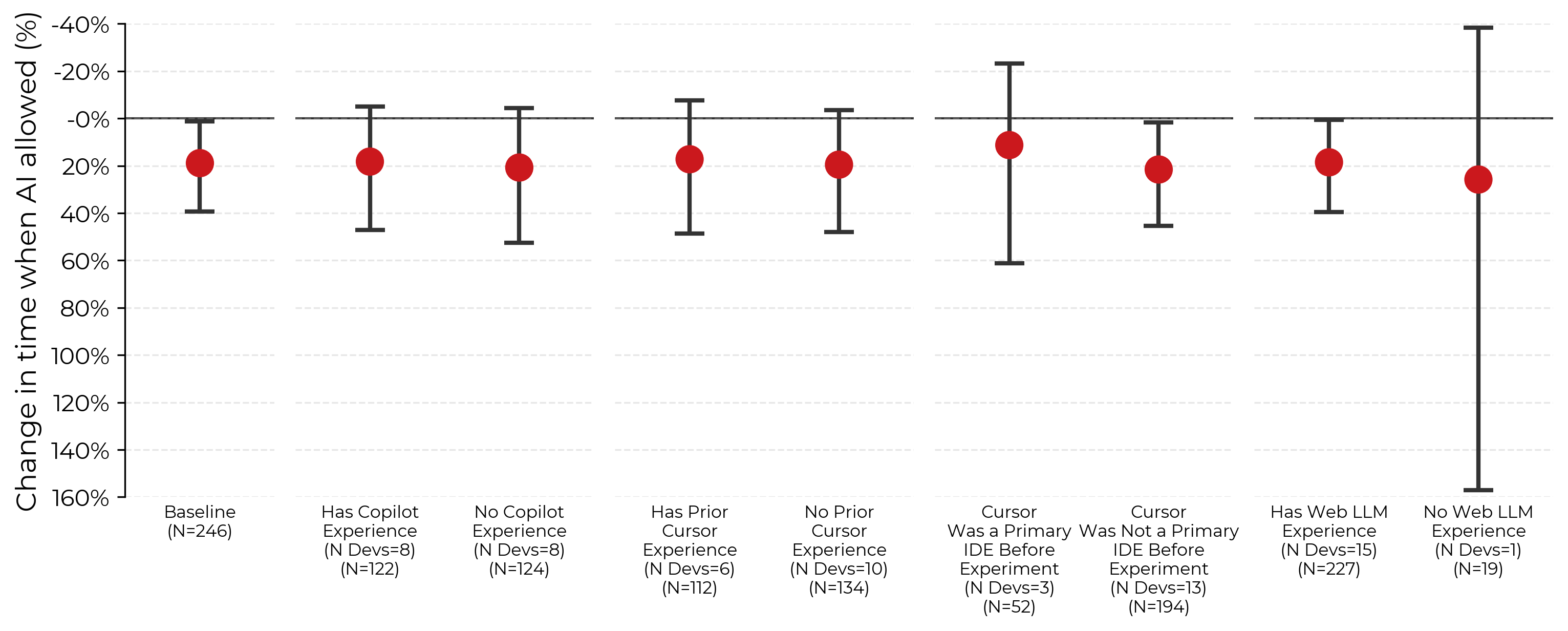}
    \caption{We evaluate speedup on various subsets of developers' prior experience with GitHub Copilot, Cursor, and web LLMs (e.g. ChatGPT). Developers with prior Cursor experience (who use Cursor in the study) are slowed down similarly to developers without prior Cursor experience, and we see no difference between developers with/without Copilot or web LLM experience. See \autoref{sec:heterogeneous_treatment_effect_estimation} for details on how we estimate heterogeneous treatment effects.}
    \label{fig:subset-prior-experience}
\end{figure}

\autoref{fig:subset-prior-experience} breaks down the percentage change in issue completion time due to AI by different levels of developers' prior experience using AI tools. We don't see meaningful differences between developers based on prior experience with AI tooling. 

\begin{figure}
    \centering
    \includegraphics[width=1\linewidth]{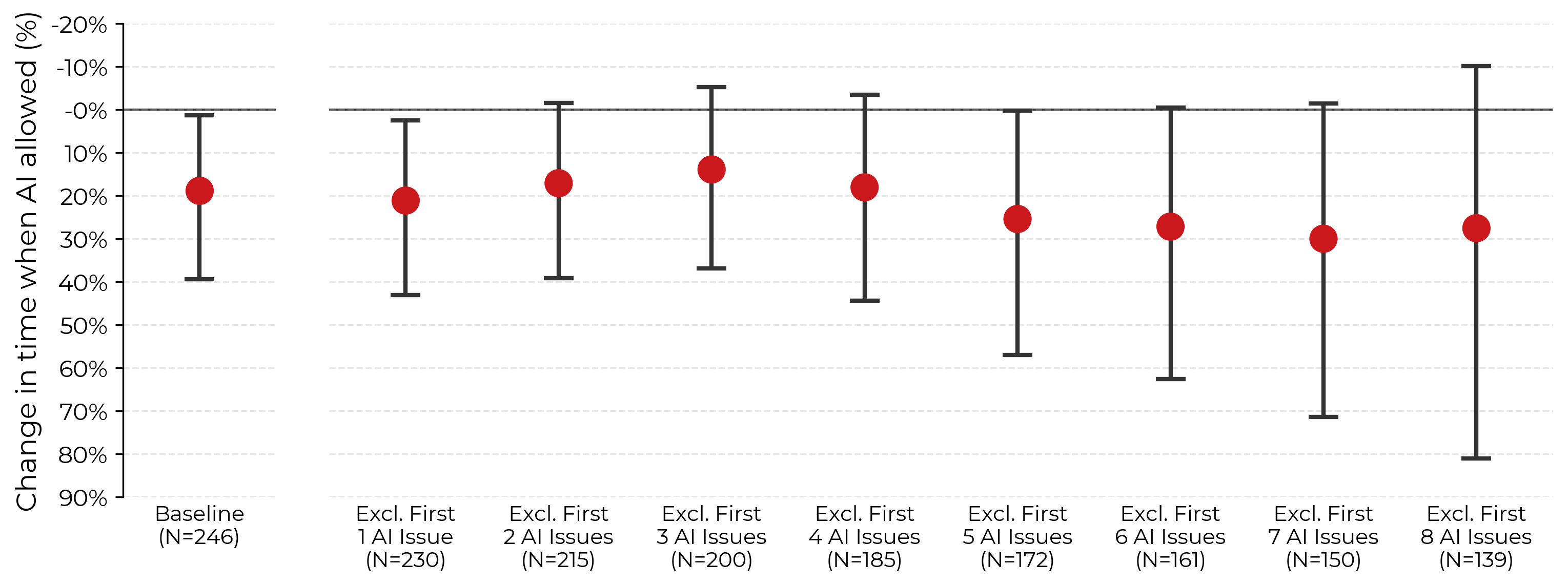}
    \caption{We see similar slowdown percentages when excluding up to the first eight AI-allowed issues developers work on, suggesting that developers lacking basic skills around using AI effectively does \textit{not} contribute substantially to the slowdown result. See \autoref{sec:heterogeneous_treatment_effect_estimation} for details on how we estimate heterogeneous treatment effects.}
    \label{fig:subset-issue-positions}
\end{figure}

We further check if developers appear to get better at using AI over the course of the experiment (\autoref{fig:subset-issue-positions}). There does not appear to be a meaningful difference in slowdown when excluding up to the first eight AI-allowed issues each developer completes. This is evidence against the hypothesis that slowdown is caused by our developers lacking basic skills in AI tool use that can be developed in a short period of time. 

\begin{figure}[t]
    \centering
    \includegraphics[width=1\linewidth]{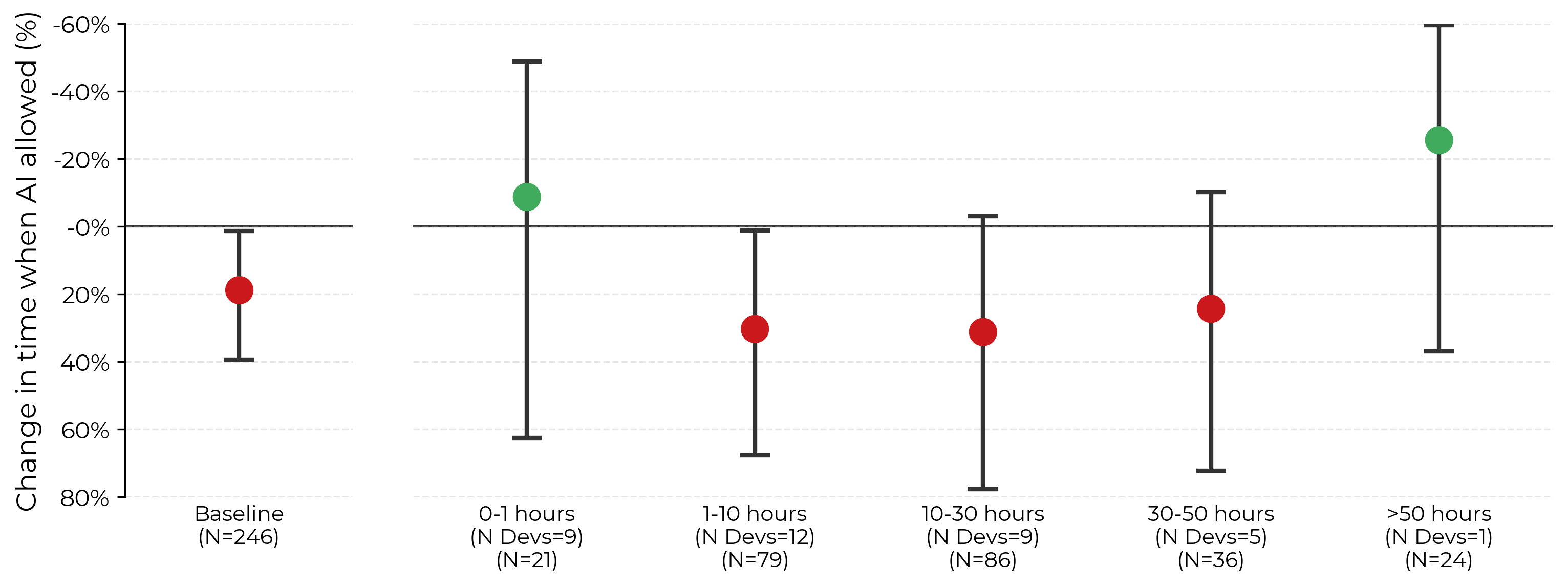}
    \caption[]{Speedup on issues where developers have varying hours of experience using Cursor (including prior Cursor experience, plus their usage during the study period). We don't see large differences across the first 50 hours that developers use Cursor, but past 50 hours we observe positive speedup.\protect\footnotemark However, we are underpowered to draw strong conclusions from this analysis. See \autoref{sec:heterogeneous_treatment_effect_estimation} for details on how we estimate heterogeneous treatment effects.}
    \label{fig:subset-learning-effects}
\end{figure}

\footnotetext{After this paper's online publication, a second developer communicated that they made a mistake when reporting their prior Cursor experience, and that in fact they had $>100$ hours of Cursor experience before the study. We observe slowdown on AI-allowed issues for this developer, and including their issues in the $>50$ hours bucket in this figure moves that bucket's point estimate from roughly 25\% speedup to roughly 0\%. We leave this figure as-is to preserve methodological consistency between developers.}

To more directly assess the impact of learning effects and AI tool use skill on productivity, we estimate speedup on issues bucketed by the number of hours of Cursor experience the developer had when working on the issue (\autoref{fig:subset-learning-effects}). This includes the number of hours of Cursor experience they self-report having before the experiment, as well as the number of hours they spend working on AI-allowed issues. Up to 50 hours of Cursor experience, it broadly does not appear that more experience reduces the slowdown effect. However, we see positive speedup for the one developer who has more than 50 hours of Cursor experience, so it's plausible that there is a high skill ceiling for using Cursor, such that developers with significant experience see positive speedup. 
As developers spend more time using AI assistance, however, their development skills without AI assistance may atrophy. This could cause the observed speedup to mostly result from weaker AI-disallowed performance, instead of stronger AI-allowed performance (which is the question we're interested in). Overall, it's unclear how to interpret these results, and more research is needed to understand the impact of learning effects with AI tools on developer productivity.

Broadly, we qualitatively observe that developers use Cursor at a level comparable to how well the authors use Cursor for software development, which is largely unsurprising, given we provide training at the beginning of the study, and periodic feedback throughout (\autoref{sec:cursor_training}). While we don't expect that developers are using AI assistance optimally, we do not find evidence that they are below-average in AI tool use ability.

\subsubsection{AI generation latency \textnormal{\textit{(Limits AI performance)}}}
\label{sec:factor-ai-generation-latency}

All else equal, faster AI generations would result in developers being slowed down less. Qualitatively, a minority of developers note that they spend significant time waiting on AI to generate code. One developer notes that for \devquote{larger refactorings, [AI generation] takes a couple of minutes}. Another developer notes that when waiting on AI generations, he \devquote{spends time on Twitter}. However, not all developers feel majorly affected by this time, for example, one developer notes that he was \devquote{never waiting for more than like 20 seconds.}

Quantitatively, on the subset of \NumLoomVideosFilteredAIAllowed\ issues with valid labeled screen recordings, we find that when AI is allowed, developers spend approximately \PctWaitingOnAIAIAllowed\% of their time waiting on AI generated outputs when working with AI. This percentage is small, but non-trivial.

Particularly given the recent benefits seen from inference/test-time compute, there are likely fundamental tradeoffs between AI output latency and performance/reliability. In general, we can imagine a pareto frontier between these variables (either for a given model, or between models/architectures)---but the optimal point on this frontier plausibly depends both on the domain, and on how exactly humans use AI tools to substitute for their labor.

\subsubsection{Suboptimal elicitation \textnormal{\textit{(Limits AI performance)}}}
\label{sec:factor-suboptimal-elicitation}

Cursor is one of the most widely used AI-enabled IDEs. Developers use Cursor agent/composer and Cursor chat in \PctLoomWithUICursorAgentOrComposer\% and \PctLoomWithUICursorChat \% of AI-allowed issues (respectively), indicating they are reasonably often using Cursor's best (at the time) scaffolding.

However, normal usage of Cursor's AI tools does not typically involve sampling more than a few thousand tokens from models. Recent literature shows that model performance can improve significantly with respect to the number of tokens sampled at inference time \cite{snell2024scalingllmtesttimecompute}, so it’s natural to wonder if the lack of speedup is driven by limited token spend.

We note that limited token spend would not explain the difference between our results and other work that find positive speedup for programming tasks \cite{peng2023impactaideveloperproductivity, cui2025effectsgenerativeai}.

However, we can imagine alternative elicitation strategies that effectively use much higher token spend, like sampling many trajectories in parallel from agents and using an LLM judge (or e.g. self-consistency \cite{wang2023selfconsistencyimproveschainthought}) to filter to the output most likely to be useful for the human. We do not provide evidence about these elicitation strategies, as developers in our study typically use Cursor and web LLMs like chatGPT, so it remains unclear how much effect these strategies would have on developer productivity in the wild.

\subsection{Factors unlikely to contribute to slowdown}
\label{sec:factorsnotdriving}

We provide suggestive evidence that \NumFactorsThatDoNotContribute\ factors are not contributing to observed slowdown, summarized in \autoref{tab:all_factors_table}.

\subsubsection{Unfamiliar development environment \textnormal{\textit{(Experimental artifact)}}}
\label{sec:factor-unfamiliar-dev-environment}

Developers qualitatively report VSCode and Cursor to be equivalent development experiences, when AI features in both are turned off\footnote{This is unsurprising because a) Cursor is a fork of VSCode, and b) Cursor has features to transfer settings, keyboard shortcuts, and other configuration information directly from VSCode to Cursor.}. Thus, to make sure they don't use AI features by accident when AI is disallowed, some developers choose to use VSCode for AI-disallowed issues, and Cursor for AI-allowed issues.

However, some developers do use different IDEs entirely for issues where AI is disallowed, and use Cursor for AI-allowed issues. If developers are substantially more productive in these other environments, e.g. because they are more familiar with keyboard shortcuts or have special personalized tooling set-up, then even if AI is helpful, they might be slowed down substantially because they aren't used to using Cursor.

We find evidence that this does not contribute to slowdown. When we restrict to developers who are using comparible IDEs between treatment conditions, we observe a slowdown of \PctSpeedupUsedComparableIDEs\%.


Furthermore, if Cursor was difficult to use effectively, we might expect to see learning effects as developers personalize it and learn its features. To measure this, we estimate the percentage speedup as developers use Cursor for different lengths of time. \autoref{fig:subset-learning-effects} does not show meaningful differences in slowdown across the first 50 hours of Cursor usage. Combined with the fact that developers rarely note having any issues or bottlenecks using Cursor, this suggests that Cursor being an unfamiliar environment does not substantially contribute to slowdown.

\subsubsection{Cheating or under-use of AI \textnormal{\textit{(Experimental artifact)}}}
\label{sec:factor-cheating-or-underuse-of-ai}

Observing screen recordings provides evidence that slowdown is not driven by developers not using AI when they are allowed to, or because they cheat and do use AI when they are not allowed to.

On the subset of \NumLoomVideosFilteredAIAllowed\ manually labeled valid screen recordings where AI is allowed, we find that developers do not use AI in only \PctTreatmentNotTakenRate \% of them, indicating that AI assistance was typically used on issues where AI was allowed.

We observe \NumLoomVideosCheating\ specific instances where AI tools were used on AI-disallowed issues, out of \NumLoomVideosInputNoAI\ labeled AI-disallowed screen recordings, suggesting an overall cheating rate of $\sim6\%$. At least one instance of observed cheating appears to be accidental.

\subsubsection{Issue dropout \textnormal{\textit{(Experimental artifact)}}}
\label{sec:factor-issue-dropout}

One potential source of bias are selection effects with respect to which issues are completed by developers. For example, if developers can choose which issues they want to complete after the issues are assigned as AI-allowed or AI-disallowed, they may choose to complete only easy AI-disallowed issues (given their expectation that AI will speed them up). This would make AI-allowed issues harder on average, which could contribute to slowdown.

To avoid this, developers commit to completing \textit{all} issues after they are assigned to AI-allowed and AI-disallowed buckets. However, as developers do not work on these repositories as their full-time job, sometimes they fail to meet these commitments. \NumDevsWithRandomizedIncompleteNoAIIssues\ developers fail to complete \NumNoAIIssuesRandomizedIncompleteIssues\ total issues assigned to AI-disallowed, and \NumDevsWithRandomizedIncompleteAIIssues\ developers fail to complete \NumAIAllowedRandomizedIncompleteIssues\ total issues assigned to AI-allowed.

However, this does not appear to be driving results. When we subset developers to those who have no uncompleted AI-disallowed issues, we still find a slowdown of \PctSpeedupDevelopersDroppedNoAIIssues\%. When we subset to developers who have no uncompleted AI-allowed issues, we find a slowdown of \PctSpeedupDevelopersDidNotDropAIIssues\%. And when we subset to developers with no dropout on any issues, we find a slowdown of \PctSpeedupDevelopersDidNotAnyIssues\%. 

Separately, \NumIssuesInvalidOtherReasons\ issues were intentionally dropped by developers, the leading reasons being: the issue ended up being too hard to implement (\NumIssuesInvalidDitchedTooHard\ issues), the issue was no longer relevant to their work by the time they got to it (\NumIssuesInvalidDitchedNotRelevant\ issues), or the issue was implemented by another developer instead (\NumIssuesInvalidTakenByOtherMaintainer\ issues). If developers made the choice to drop these issues because they were not allowed to use AI, this selection effect could bias results. However, when asked about de-prioritized issues, developers confirm that the issues are not de-prioritized due to AI usage. We manually review these intentionally dropped issues to confirm that they are not dropped for reasons related to their condition assignment.

\subsubsection{Non-robust outcome measure \textnormal{\textit{(Experimental artifact)}}}
\label{sec:factor-specific-outcome-measure}

\paragraph*{Imputation of Post-Review Time on Unreviewed Issues}
\label{sec:imputation_robustness}

In \autoref{sec:effect_estimation}, we describe how we impute a small number ($\sim 12\%$) of post-review implementation times for issues that were not reviewed by the completion of the study. These \NumIssuesInitialImplementationStartedButNotFinished\ issues are balanced between AI-allowed and AI-disallowed conditions, and the forecasted times to complete are similar, which substantially reduces the likelihood of strong selection effects biasing results.

To quantitatively check the robustness of our post-review time imputation, we re-run our speedup estimation in two ways: 1) we impute AI-allowed post-review implementation times as a constant 1 hour (while AI-disallowed post-review time is imputed as 0), and 2) we impute AI-disallowed post-review implementation times as one hour (while AI-allowed is set to 0). This is intended to simulate the case if post-review times for unreviewed issues were systematically biased for one condition or the other, and because the mean post-review implementation times are \MeanReviewTimeMinutesAIAllowed\ and \MeanReviewTimeMinutesNoAI\ minutes for AI-allowed and AI-disallowed issues respectively, this would represent a very extreme bias. When imputing using this extreme bias, we still do not observe substantial impacts on our speedup estimates in either case (\PctSpeedupBaselineImputationAddOneHourAIDisallowedIssuesAbsValue \% and \PctSpeedupBaselineImputationAddOneHourAIAllowedIssuesAbsValue \% slowdowns when using AI, respectively).

\paragraph*{Self-Reported Times}

To validate that there is not significant bias in self-reported times that contributes to slowdown, we also measure the slowdown percentage computed using the total length of screen recordings associated with a given issue, instead of using self-reported times. We observe a \PctSpeedupLoomDuration \% slowdown when using total screen recording time instead of the self-reported time spent, which is similar to the \PctSpeedupSelfReportedTimeForThoseWithLoomDuration\% we observe when estimating speedup from self-reported implementation times on the same subset of issues.

\subsubsection{Non-robust estimator \textnormal{\textit{(Experimental artifact)}}}
\label{sec:factor-specific-estimator}

We evaluate the slowdown implied by alternative estimators, in addition to the regression-based estimator that yields our primary results.

\begin{figure}[t]
    \centering
    \includegraphics[width=1\linewidth]{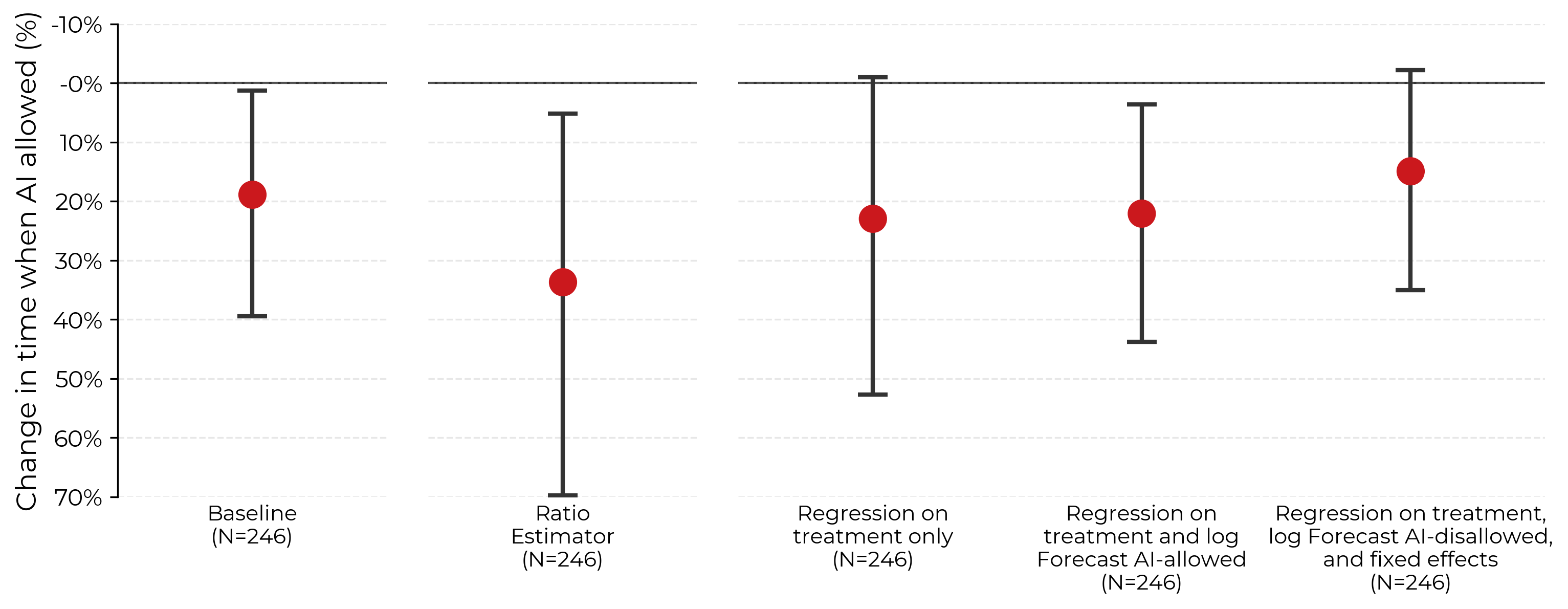}
    \caption{In addition to a basic ratio estimator, we also evaluate speedup when regressing $\log$(implementation time) on whether AI is allowed; whether AI is allowed and the log forecast of implementation time with AI-allowed; and whether AI is allowed, the log forecast of implementation time with AI-disallowed, and developer fixed effects.}
    \label{fig:subset-alternative-estimators}
\end{figure}

In \autoref{fig:subset-alternative-estimators} we report the slowdown observed when using four alternative estimators other than our baseline regression described in \autoref{sec:empirical-strategy}. We compare a simple ratio estimator, which reports $\frac{\text{Mean}(\text{time with AI allowed})}{\text{Mean}(\text{time with AI disallowed})} - 1$ (\autoref{sec:ratio-estimator}, as well as three other regression-based estimators with different covariate specifications (described in \autoref{fig:subset-alternative-estimators}'s caption).

The alternative estimators all report similar results, suggesting that slowdown is robust to our particular estimator specification.

\subsubsection{Non-frontier model usage \textnormal{\textit{(Limits AI performance)}}}
\label{sec:factor-non-frontier-model-usage}

Given the significant rate of progress in AI (particularly in software/coding domains), one explanation for slowdown could be that developers used older, less capable models on AI-allowed issues.

This does not appear to be the case. Across the \NumLoomVideosFilteredAIAllowed \ valid labeled AI-allowed loom videos, we find that developers used Claude 3.7 Sonnet (thinking mode), Claude 3.7 Sonnet, and Claude 3.5 Sonnet in \PctLoomWithModelClaudeSonnetThreeSevenThinking\%, \PctLoomWithModelClaudeSonnetThreeSeven\%, and \PctLoomWithModelClaudeSonnetThreeFive\% of issues respectively. Other models used are GPT-4o (\PctLoomWithModelGPTFourO\%), Gemini 2.5 Pro (\PctDaysWithModelGeminiTwoFivePro\%), and o1 (\PctLoomWithModelOOne\%). Developers prompt no AI model \PctTreatmentNotTakenRate \% of the time. These percentages do not include AI autocomplete, which developers use on the vast majority of AI-allowed issues.

Most issues were completed in February and March 2025, before models like Claude 4 Opus or Gemini 2.5 Pro were released.


\section{Empirical Strategy}
\label{sec:empirical-strategy}

\subsection{Regression}
\label{sec:regression}

For each issue \(i\) we observe the realised completion time \(T_i>0\), a binary treatment flag \(\text{AI}_i\in\{0,1\}\), and the developer’s \emph{ex-ante} forecast of how long the task would take without AI, denoted \(\widehat{T}^{\text{NoAI}}_i>0\).

We estimate the log-linear model

\begin{equation}
\label{eq:main-reg}
\log T_i = \alpha + \beta\,\text{AI}_i + \delta\,\log \widehat{T}^{\text{NoAI}}_i + \varepsilon_i,
\end{equation}

via ordinary least squares.  Random assignment of \(\text{AI}_i\) guarantees consistency of \(\hat\beta\) for \(\beta=\mathbb{E}[\log T\,|\,\text{AI}=1]-\mathbb{E}[\log T\,|\,\text{AI}=0]\).

We include forecasts as a control variable because they serve as a proxy for issue difficulty and are highly predictive of completion times. This substantially increases our statistical power without introducing bias, as forecasts were elicited prior to treatment assignment and thus cannot be affected by treatment status.\footnote{We do not generally include developer fixed effects because they explain minimal variation in the outcome conditional on forecasts. \autoref{sec:factor-specific-estimator} displays estimates from a regression specification including developer fixed effects.}

\begin{figure}[t]
    \centering
    \includegraphics[width=1\linewidth]{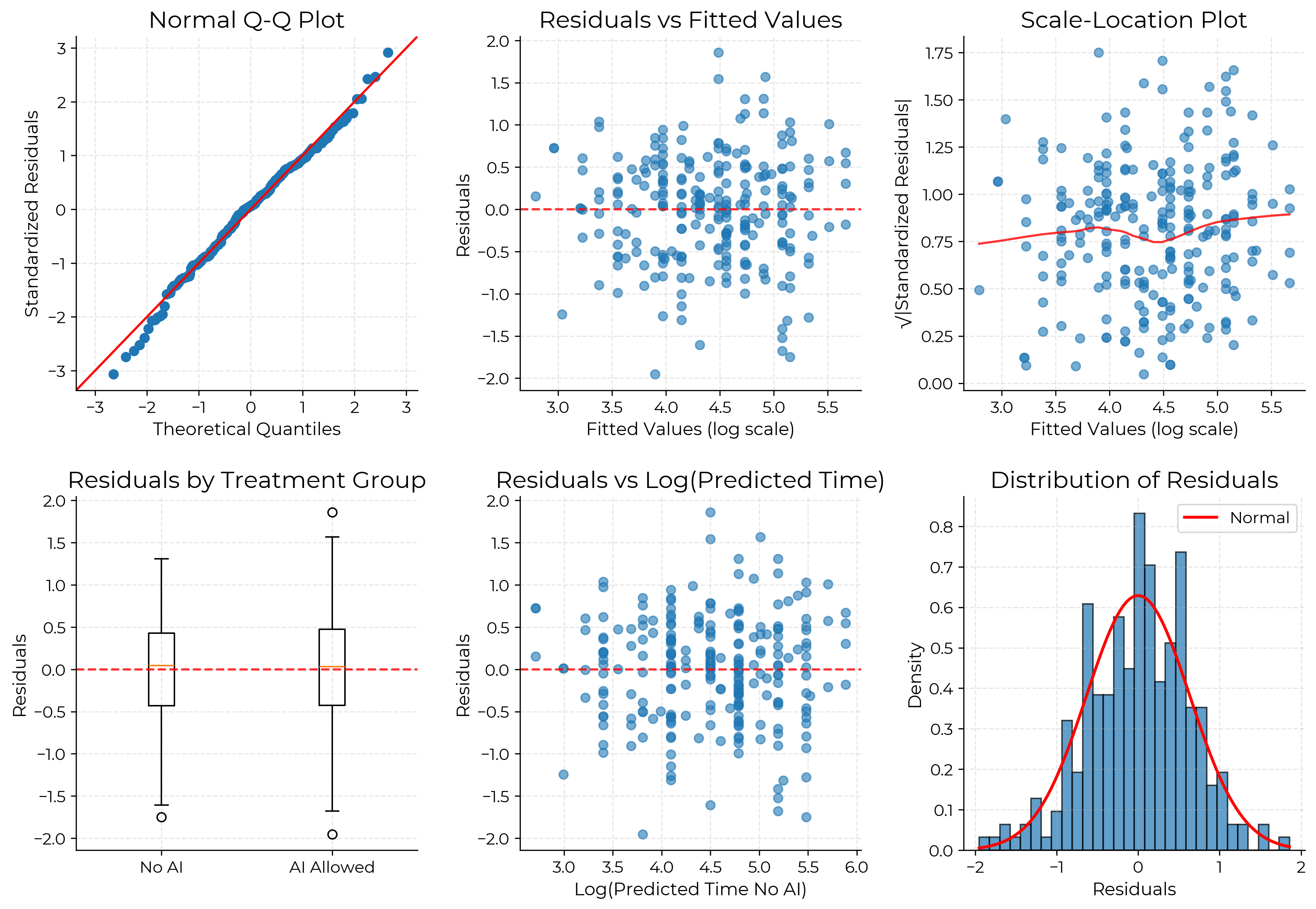}
    \caption{Regression diagnostics associated with the regression specification in equation \eqref{eq:main-reg}.}
    \label{fig:reg-diagnostics}
\end{figure}

\autoref{fig:reg-diagnostics} displays regression diagnostics associated with this specification.

\subsection{Confidence intervals}
\label{sec:confidence-intervals}

\begin{figure}[t]
    \centering
    \includegraphics[width=1\linewidth]{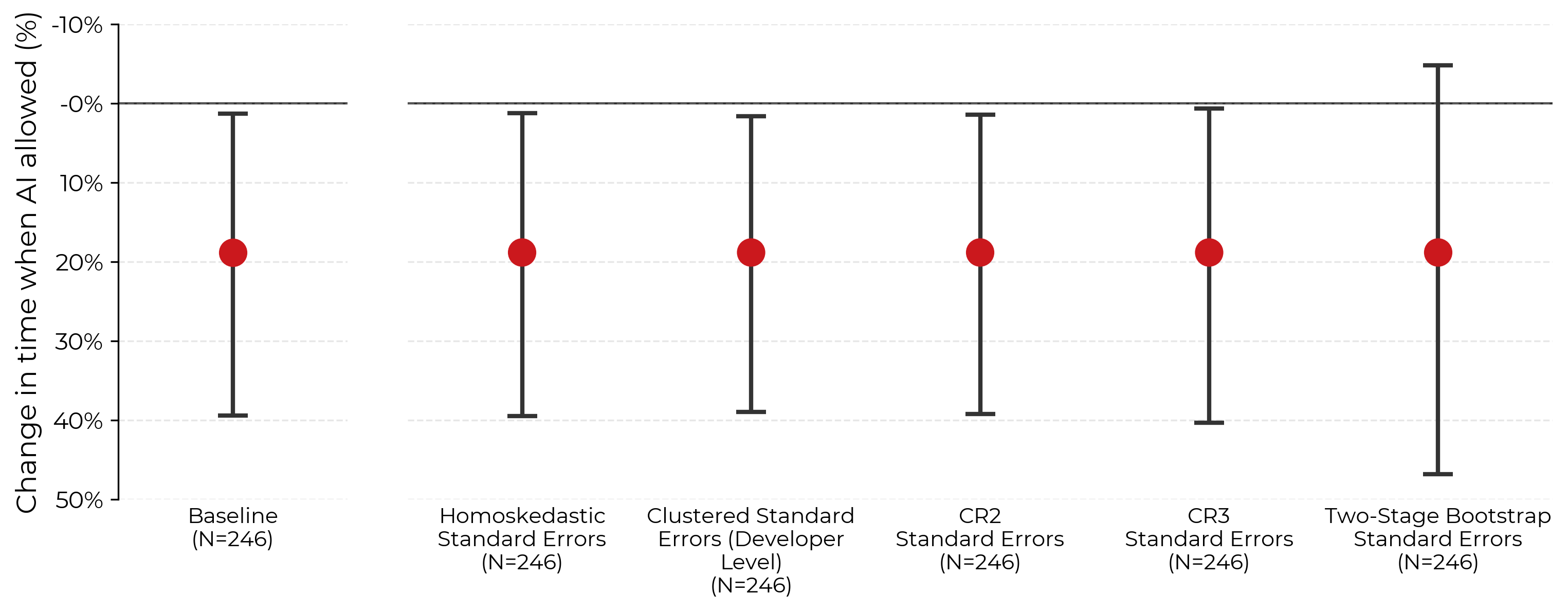}
    \caption{Speedup calculated using the regression specification in equation \eqref{eq:main-reg} and alternative uncertainty estimation procedures. CR2 and CR3 standard errors are bias-corrected cluster-robust standard errors \cite{ColinCameron317, articlebell}. Two-stage bootstrap resamples developers, then issues within developers.}
    \label{fig:subset-ses}
\end{figure}

By default, we report 95\% confidence intervals using HC3 standard errors.\footnote{Given that our sampling of developers is non-random and treatment is assigned at the issue level, \citet{10.1093/qje/qjac038} do not necessarily recommend developer-level clustering even if residuals are correlated within-developer. If we are willing to treat developer sampling as random then clustering is appropriate.} \autoref{fig:subset-ses} displays 95\% confidence intervals from alternative uncertainty estimation procedures. Standard errors clustered at the developer level and bias-corrected cluster-robust standard errors give similar results; a hierarchical bootstrap resampling developers and then issues within developers yields somewhat wider confidence intervals.

\subsection{Converting to speedup}
\label{sec:effect-transformation}

Taking conditional expectations of equation \eqref{eq:main-reg} given treatment status and forecast:

\begin{align}
\mathbb{E}[\log T\,|\,\text{AI}=1, \widehat{T}^{\text{NoAI}}] &= \alpha + \beta + \delta\,\log \widehat{T}^{\text{NoAI}} \\
\mathbb{E}[\log T\,|\,\text{AI}=0, \widehat{T}^{\text{NoAI}}] &= \alpha + \delta\,\log \widehat{T}^{\text{NoAI}}
\end{align}

Therefore, $\beta = \mathbb{E}[\log T\,|\,\text{AI}=1, \widehat{T}^{\text{NoAI}}] - \mathbb{E}[\log T\,|\,\text{AI}=0, \widehat{T}^{\text{NoAI}}]$ represents the average treatment effect on the log scale, conditional on forecast.

To convert to a speedup measure, we note that:

\begin{align}
\label{eq:exponentiate-estimand}
\exp(\beta) &= \frac{\mathbb{E}[T\,|\,\text{AI}=1, \widehat{T}^{\text{NoAI}}]}{\mathbb{E}[T\,|\,\text{AI}=0, \widehat{T}^{\text{NoAI}}]} \quad\\
\text{S} &= \exp(\beta) - 1
\end{align}

(Step \eqref{eq:exponentiate-estimand} holds provided the disturbance $\varepsilon_i$ is independent of both $\text{AI}_i$ and $\widehat{T}^{\text{NoAI}}_i$ and that $\mathbb{E}\!\left[e^{\varepsilon_i}\right]$ exists.\footnote{Because $T_i=e^{\alpha}e^{\beta\,\text{AI}_i}\,
(\widehat{T}^{\text{NoAI}}_i)^{\delta}\,e^{\varepsilon_i}$, taking expectations conditional on $\text{AI}_i$ and $\widehat{T}^{\text{NoAI}}_i$ gives $\mathbb{E}[T\,|\,\text{AI}=j,\widehat{T}^{\text{NoAI}}]= e^{(\alpha + \beta j)}\, (\widehat{T}^{\text{NoAI}})^{\delta}\, \mathbb{E}[e^{\varepsilon_i}]$.  
The common factor $\mathbb{E}[e^{\varepsilon_i}]$ cancels when we form the ratio of these conditional means, yielding $\mathbb{E}[T\,|\,\text{AI}=1,\widehat{T}^{\text{NoAI}}]\big/ \mathbb{E}[T\,|\,\text{AI}=0,\widehat{T}^{\text{NoAI}}]=\exp(\beta)$.})

Then under standard regularity assumptions (which our diagnostics suggest hold) our OLS estimator $\hat\beta$ is normally distributed $\mathcal{N}(\beta, \sigma^2)$ for some $\sigma$ which we can estimate using it's standard error $\text{SE}[\hat\beta]$. This lets us construct a 95\% confidence interval for $\beta$ in the usual way:

\begin{equation}
\text{CI}_{95\%} = \left[\hat\beta - 1.96 \cdot \text{SE}[\hat\beta], \hat\beta + 1.96 \cdot \text{SE}[\hat\beta]\right]
\end{equation}

As $\text{S} = \exp(\beta) - 1$ is a monotonic function of $\beta$ we can construct a confidence interval for $\text{S}$ by simply applying the function to the endpoints of $\beta$'s confidence interval:

\begin{equation}
\text{CI}_{95\%} = \left[e^{\hat\beta - 1.96 \cdot \text{SE}[\hat\beta]} - 1, e^{\hat\beta + 1.96 \cdot \text{SE}[\hat\beta]} - 1\right]
\end{equation}

\subsection{Heterogeneous treatment effects}
\label{sec:heterogeneous_treatment_effect_estimation}

To test for differential treatment effects across subgroups, we estimate models with interaction terms. For a binary characteristic $X_i$ (e.g., prior Cursor experience), we estimate:

\begin{equation}
\log T_i = \alpha + \beta_1\,\text{AI}_i + \beta_2\,X_i + \beta_3\,(\text{AI}_i \times X_i) + \delta\,\log \widehat{T}^{\text{NoAI}}_i + \varepsilon_i
\end{equation}

Taking conditional expectations:

\begin{align}
\mathbb{E}[\log T\,|\,\text{AI}=1, X=0, \widehat{T}^{\text{NoAI}}] - \mathbb{E}[\log T\,|\,\text{AI}=0, X=0, \widehat{T}^{\text{NoAI}}] &= \beta_1\\
\mathbb{E}[\log T\,|\,\text{AI}=1, X=1, \widehat{T}^{\text{NoAI}}] - \mathbb{E}[\log T\,|\,\text{AI}=0, X=1, \widehat{T}^{\text{NoAI}}] &= \beta_1 + \beta_3
\end{align}

Thus, the treatment effect for the $X=0$ group is $\beta_1$, while for the $X=1$ group it is $\beta_1 + \beta_3$. We similarly transform these to speedup measures: $\text{S}_{X=0} = \exp(\beta_1) - 1$ and $\text{S}_{X=1} = \exp(\beta_1 + \beta_3) - 1$.

To construct confidence intervals for these subgroup effects, we test linear hypotheses of the form $L^T\boldsymbol{\theta} = c$. For the $X=1$ group effect, $L^T = [0, 1, 0, 1, 0]$ selects $\beta_1 + \beta_3$. The Wald statistic:

\begin{equation}
W = \frac{L^T\hat{\boldsymbol{\theta}} - c}{\sqrt{L^T\hat{V}L}} \sim \mathcal{N}(0,1)
\end{equation}

where $\hat{V}$ is the variance-covariance matrix. This asymptotic normality allows us to construct confidence intervals using the same critical values. For the linear combination $\hat\beta_1 + \hat\beta_3$:

\begin{equation}
\text{CI}_{95\%} = \left[(\hat\beta_1 + \hat\beta_3) - 1.96 \cdot \text{SE}[\hat\beta_1 + \hat\beta_3], (\hat\beta_1 + \hat\beta_3) + 1.96 \cdot \text{SE}[\hat\beta_1 + \hat\beta_3]\right]
\end{equation}

We then apply the same transformation as above to obtain confidence intervals for $S$:

\begin{align}
\text{CI}_{95\%} = \left[e^{(\hat\beta_1 + \hat\beta_3) - 1.96 \cdot \text{SE}[\hat\beta_1 + \hat\beta_3]} - 1, e^{(\hat\beta_1 + \hat\beta_3) + 1.96 \cdot \text{SE}[\hat\beta_1 + \hat\beta_3]} - 1\right]
\end{align}

\subsection{Ratio estimator}
\label{sec:ratio-estimator}

For robustness, we also compute a simple ratio-of-means estimator: $\hat{S}_{\text{ratio}} = \frac{\bar{T}_{\text{AI-allowed}}}{\bar{T}_{\text{AI-disallowed}}} - 1$, where $\bar{T}$ denotes the sample mean of implementation times. We obtain 95\% confidence intervals using quantiles of $\hat{S}_{\text{ratio}}$ from a pooled bootstrap.

We use this same estimator to calculate aggregate percentage change in implementation times implied by developer forecasts. Note that this procedure is distinct from averaging over percentage changes in implementation time per-task.

\subsection{Participant post-study estimator}
\label{sec:retrodiction-estimator}

After the study period, developers estimate post hoc their speedup from AI across their completed issues. To make these estimates comparable to their forecasts (which they provide before each issue is randomized to allow or disallow AI), we compute a weighted arithmetic mean of the ratios of times they report. Each developer's ratio \( r_i \) is weighted proportionally to the number of issues \( n_i \) they completed:

\begin{equation}
\bar{r}_w = \frac{\sum_{i=1}^{N} n_i r_i}{\sum_{i=1}^{N} n_i}
\end{equation}

We calculate the standard error of this weighted mean using the weighted variance formula:

\begin{equation}
\text{SE}(\bar{r}_w) = \sqrt{\frac{\sum_{i=1}^{N} w_i (r_i - \bar{r}_w)^2}{\left(\sum_{i=1}^{N} w_i\right)^2}}
\end{equation}

where $w_i = n_i$ are the issue-based weights. The 95\% confidence interval around the weighted mean is then derived using the standard normal approximation: $\bar{r}_w \pm 1.96 \times \text{SE}(\bar{r}_w)$.

\section{Other Analysis}

\subsection{Balance and proportions checks}

\begin{table}[htbp]
\centering \small
\begin{tabular}{lrrrr}
\toprule
Characteristic & AI-allowed & AI-disallowed & Difference & p-value \\
 & (n=136) & (n=110) & & \\
\midrule
AI-disallowed forecasted time & 116 (80) & 106 (63) & 9 & 0.28 \\
AI-allowed forecasted time & 87 (62) & 82 (53) & 4 & 0.56 \\
\bottomrule
\end{tabular}
\caption{Balance table comparing forecasted issue completion times between AI-allowed and AI-disallowed groups. Means are shown with standard deviations in parentheses. P-values are from two-sided Welch t-tests. There is no meaningful difference in forecasted difficulty, indicating successful randomization.}
\label{tab:balance_and_proportions}
\end{table}

\begin{table}[htbp]
\centering \small
\begin{tabular}{lrrr}
\toprule
Treatment Group & Intended Proportion & Actual Proportion & Sample Size \\
\midrule
AI-allowed & 50\% & 55.3\% (3.2\%) & 136 \\
AI-disallowed & 50\% & 44.7\% (3.2\%) & 110 \\
\midrule
P-value & & 0.10 & \\
\bottomrule
\end{tabular}
\caption{Treatment assignment proportions. Actual proportions are shown with standard errors in parentheses. P-value is from chi-square test for deviation from intended 50/50 allocation.}
\label{tab:treatment_proportions}
\end{table}

\autoref{tab:balance_and_proportions} confirms that randomization leads to balanced average difficulty between AI-allowed and AI-disallowed groups. \autoref{tab:treatment_proportions} shows that the difference in the proportion of issues randomly assigned to AI-allowed and AI-disallowed groups is not statistically significant at common p-value thresholds.

\begin{figure}[t]
    \centering
    \includegraphics[width=0.8\linewidth]{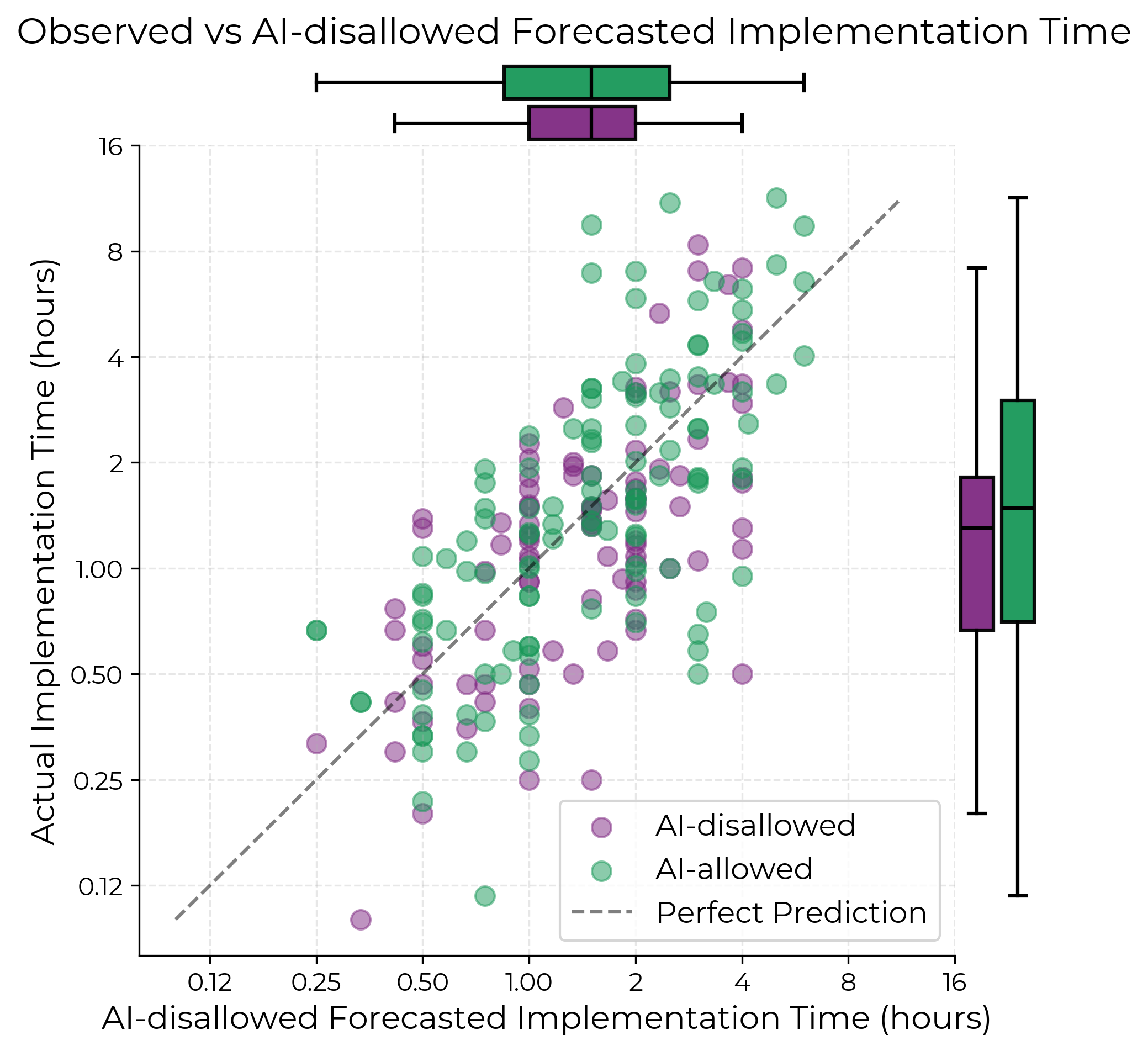}
    \caption{Distributions of issue implementation time with AI allowed and disallowed as a function of the forecasted implementation time without AI.}
    \label{fig:time-vs-forecast-no-ai-scatter}
\end{figure}

\subsection{Per-developer speedup and forecast calibration}

\autoref{fig:time-vs-forecast-no-ai-scatter} shows the relationship between the times developers forecast issues will take without AI (which we interpret as a forecast of issue difficulty), and how long the issues actually end up taking them (colored by whether AI was allowed or disallowed for each issue). We can see that the median forecasted implementation time is almost identical across treatment conditions, while AI-allowed issues take longer on average than AI-disallowed issues (note the log axes).

\begin{figure}[t]
    \centering
    \includegraphics[width=1\linewidth]{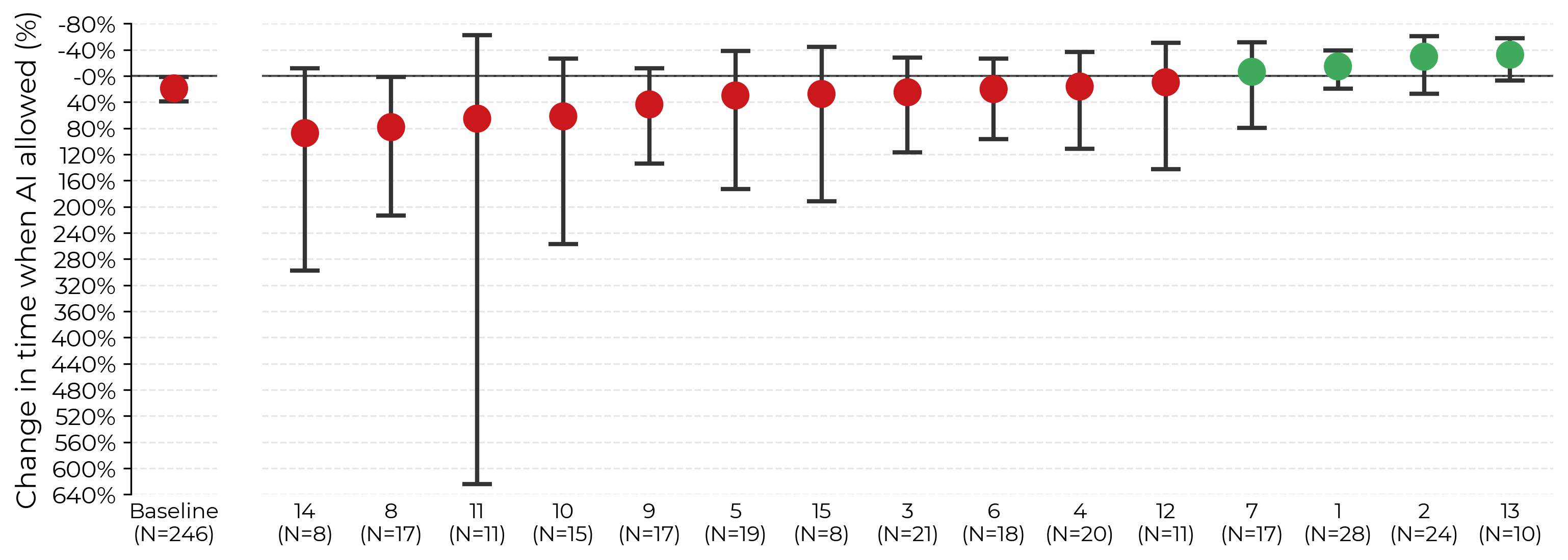}
    \caption{Speedup estimates per developer. Developer IDs correspond to the developer's rank by number of issues in our data.}
    \label{fig:per-developer-speedup}
\end{figure}

We estimate speedup per developer (\autoref{fig:per-developer-speedup}) using our standard methodology for estimating heterogeneous effects (\autoref{sec:heterogeneous_treatment_effect_estimation}). 75\% of developers experience slowdown.

\begin{figure}[t]
    \centering
    \includegraphics[width=1\linewidth]{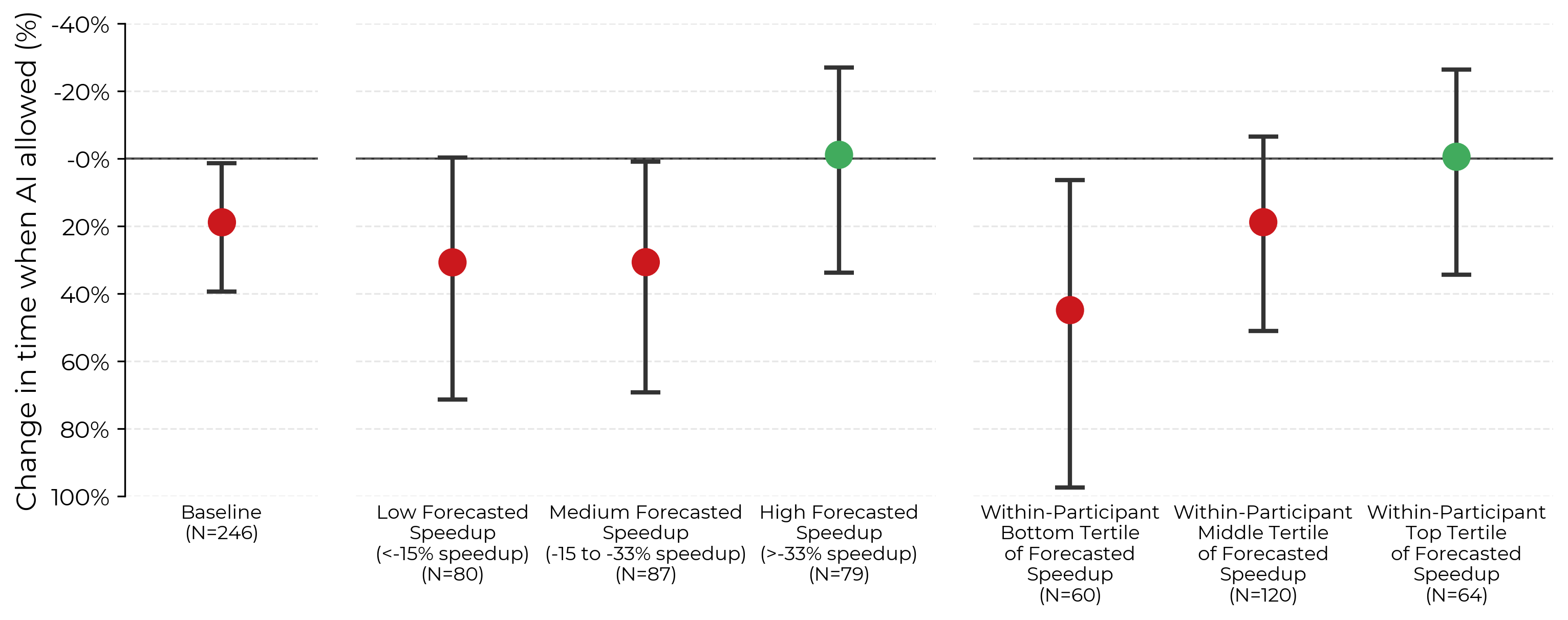}
    \caption{Speedup broken down by forecasted speedup between and within developers (developers forecast how long they expect each issue to take with and without AI). Speedup cutoffs are chosen to make bins approximately similarly sized. Tertiles are imbalanced because forecasted speedup contains duplicates that we assign to a single bin. Developers experience less slowdown on issues that they forecast high speedup. See \autoref{sec:heterogeneous_treatment_effect_estimation} for details on how we estimate heterogeneous treatment effects.}
    \label{fig:subset-stats-forecast-uplift}
\end{figure}

Interestingly, despite developers reliably forecasting incorrectly that AI-allowed issues will take less time, they are still calibrated in a relative way on the speedup from AI (\autoref{fig:subset-stats-forecast-uplift}). Specifically, on issues that developers predict significant speedup (i.e. $\geq33\%$ or the top tertile of forecasted speedup), developers are not slowed down by AI, and slowdown monotonically decreases as the forecasted speedup increases. 

\subsection{Randomization}
\label{sec:randomization}

There were \NumIssuesNonCoinFlipAllInitialImplementationFinishedValid\ issues early in the study that were randomized using a block randomization scheme intended to increase statistical power. Developer issue lists ended up being too small for this strategy to be viable, so we abandoned it early on in favor of simply using a simulated fair coin flip.

Excluding these issues does not affect our result---we still find a slowdown of \PctSpeedupInitialImplementationFinishedIssuesCoinFlipRandomizationInitialImplementationTime \%. Given this, we include these issues in our analysis and results to increase statistical power.

\subsection{Fine-Grained Screen Recording Labels}
\label{sec:other_loom_plots}

\begin{figure}[t]
    \centering
    \includegraphics[width=1\linewidth]{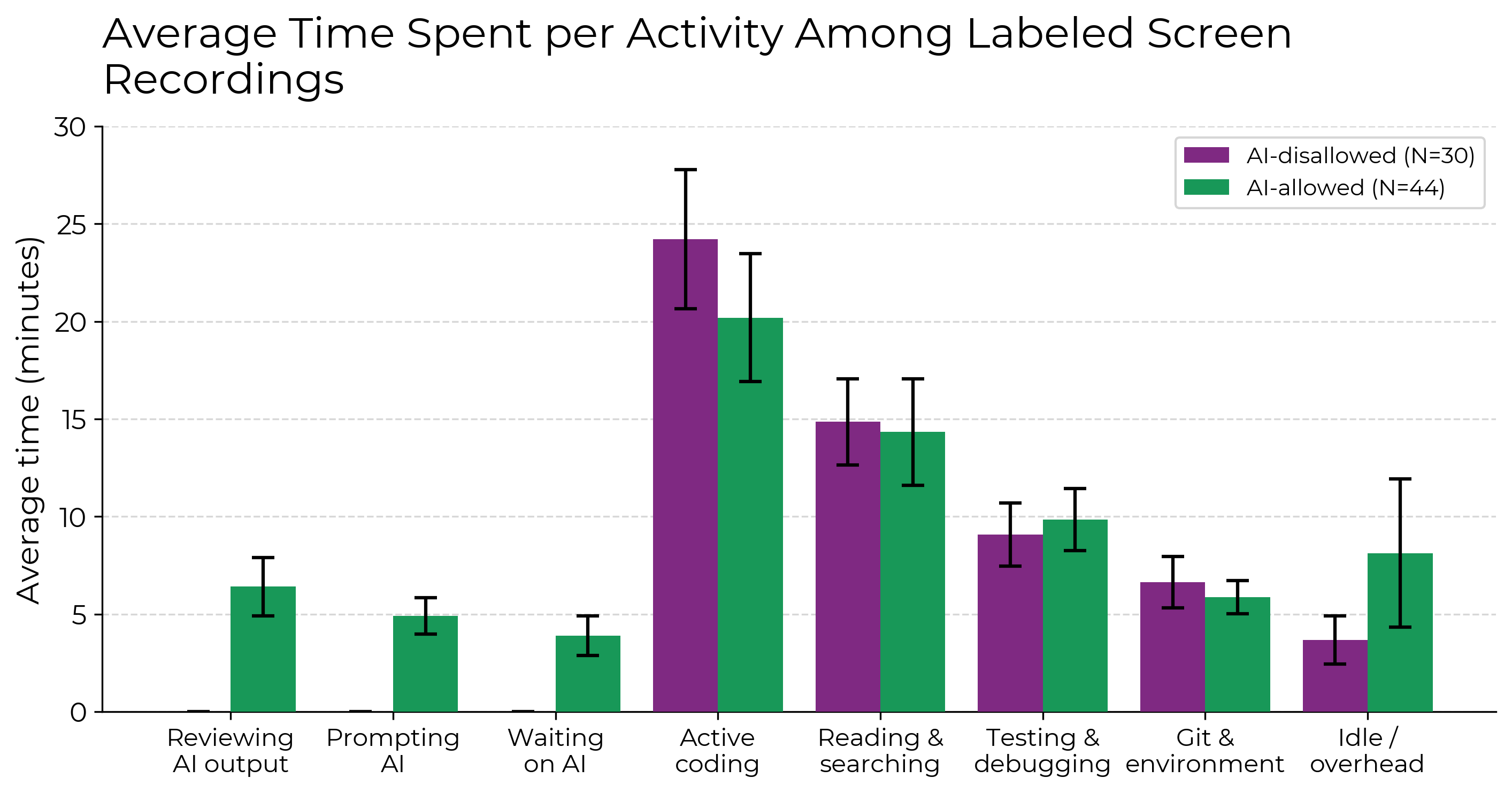}
    \caption{Average minutes spent on different high-level activities when completing issues with AI-allowed vs. AI-disallowed.}
    \label{fig:loom-high-category-minutes}
\end{figure}

\begin{figure}[t]
    \centering
    \includegraphics[width=1\linewidth]{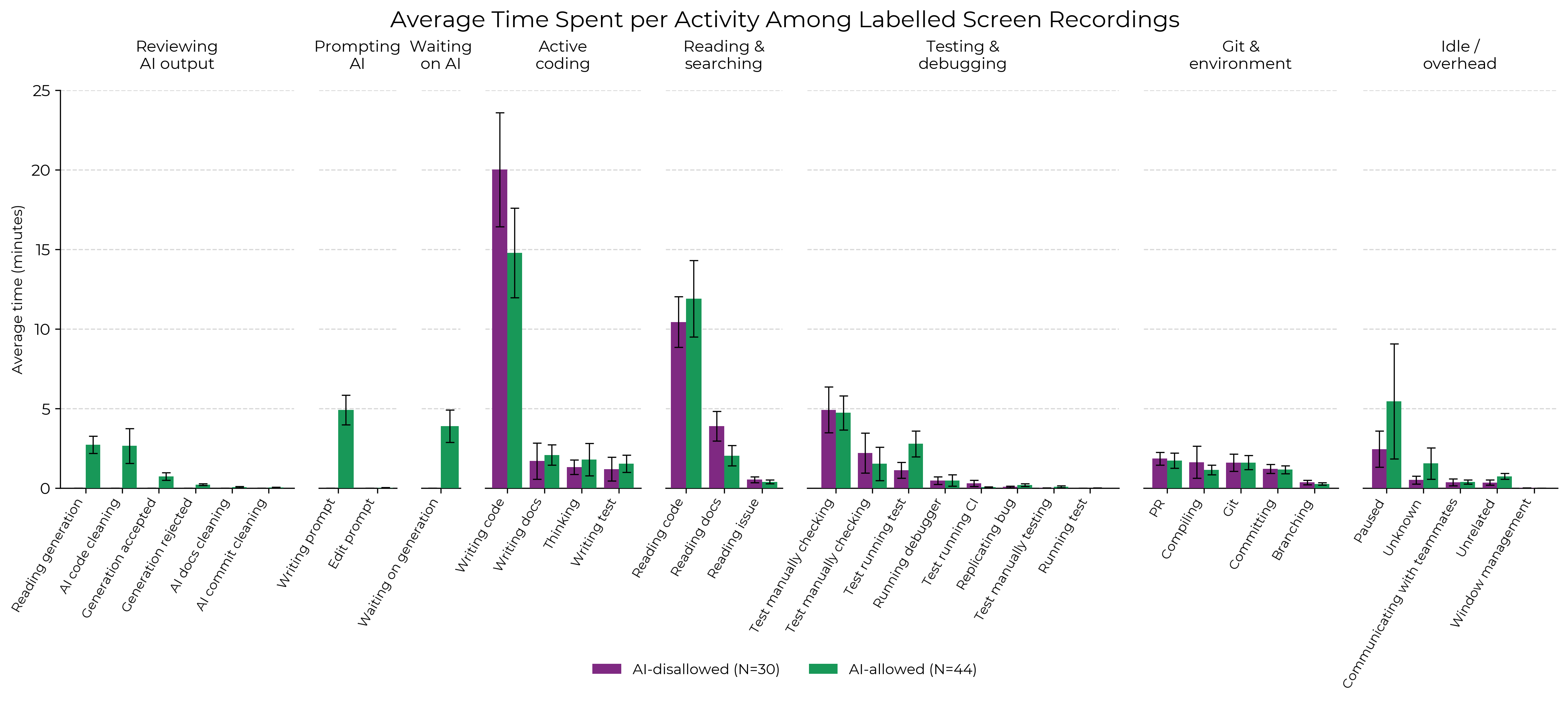}
    \caption{Average minutes spent across 27 fine-grained activity categories.}
    \label{fig:loom-low-category-minutes}
\end{figure}

\begin{figure}[t]
    \centering
    \includegraphics[width=1\linewidth]{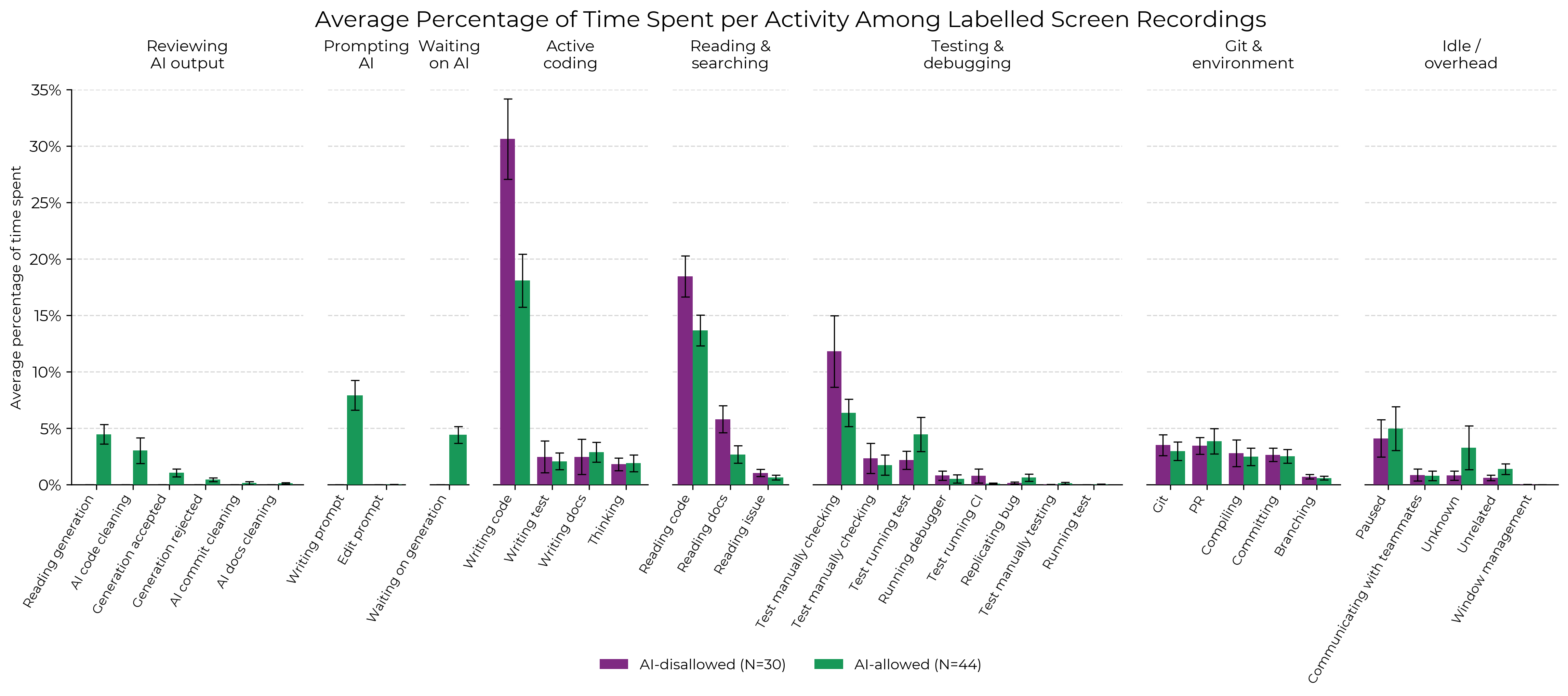}
    \caption{Percentage of time spent on fine-grained activities when AI is allowed vs. disallowed.}
    \label{fig:loom-low-category-percentage}
\end{figure}

Figures \ref{fig:loom-high-category-minutes}, \ref{fig:loom-low-category-minutes}, and \ref{fig:loom-low-category-percentage} present various breakdowns of the time developers spend on different activities as they work. See \autoref{sec:screen_recording_labels} and \autoref{sec:screen_recording_instructions} for more detail on screen recording labels.

\subsection{Expert forecasts}
\label{sec:other_analysis_expert_forecasts}

We display summary statistics regarding expert forecasts of our result. Recall that we originally elicited expert forecasts for $\frac{\mathbb{E}[T\,|\,\text{AI}=0]}{\mathbb{E}[T\,|\,\text{AI}=1]}$, but here report results on the $\frac{\mathbb{E}[T\,|\,\text{AI}=1]}{\mathbb{E}[T\,|\,\text{AI}=0]} - 1$ scale we use for \autoref{fig:horizontal_iceberg}.

\begin{center}
    \begin{table}[tbp]
\centering
\label{tab:expert_forecasts}
\begin{tabular}{lccccccc}
\toprule
Expert Group & N & Mean & Min & P25 & P50 & P75 & Max \\
\midrule
Economics & 34 & -38.7 & -80.0 & -56.0 & -37.5 & -26.3 & 81.8 \\
Machine Learning & 54 & -38.0 & -88.9 & -55.5 & -33.3 & -20.2 & 0.0 \\
\bottomrule
\end{tabular}
\vspace{1em}
\caption{Expert Forecast Statistics}
\end{table}

\end{center}

\subsection{Other treatment effects}
\label{sec:other_treatment_effects}

\begin{figure}[t]
    \centering
    \includegraphics[width=0.8\linewidth]{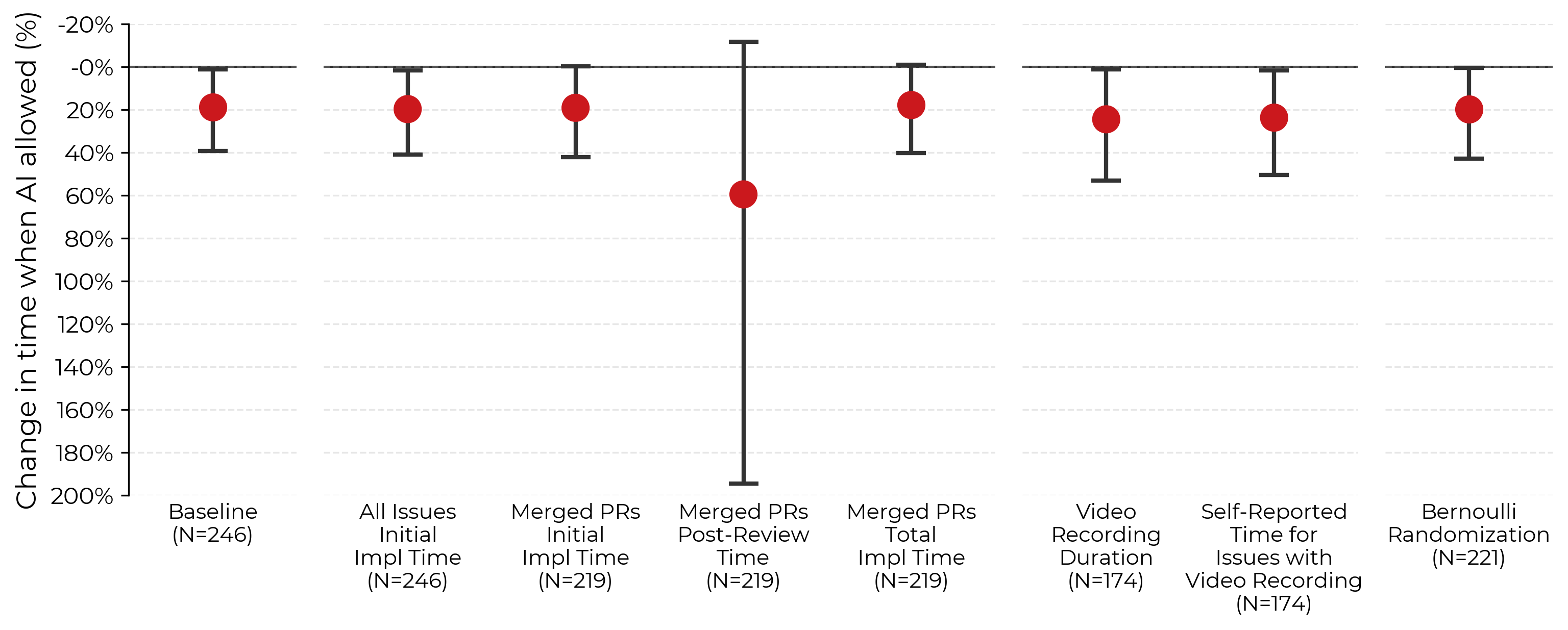}
    \caption{Speedup by alternative outcomes measures or subsets of our data. The Bernoulli randomization subset excludes the \NumIssuesNonCoinFlipAllInitialImplementationFinishedValid\ issues randomized using a block randomization scheme (see \autoref{sec:randomization}).}
    \label{fig:subset-alternative-outcome}
\end{figure}

\begin{figure}[t]
    \centering
    \includegraphics[width=0.8\linewidth]{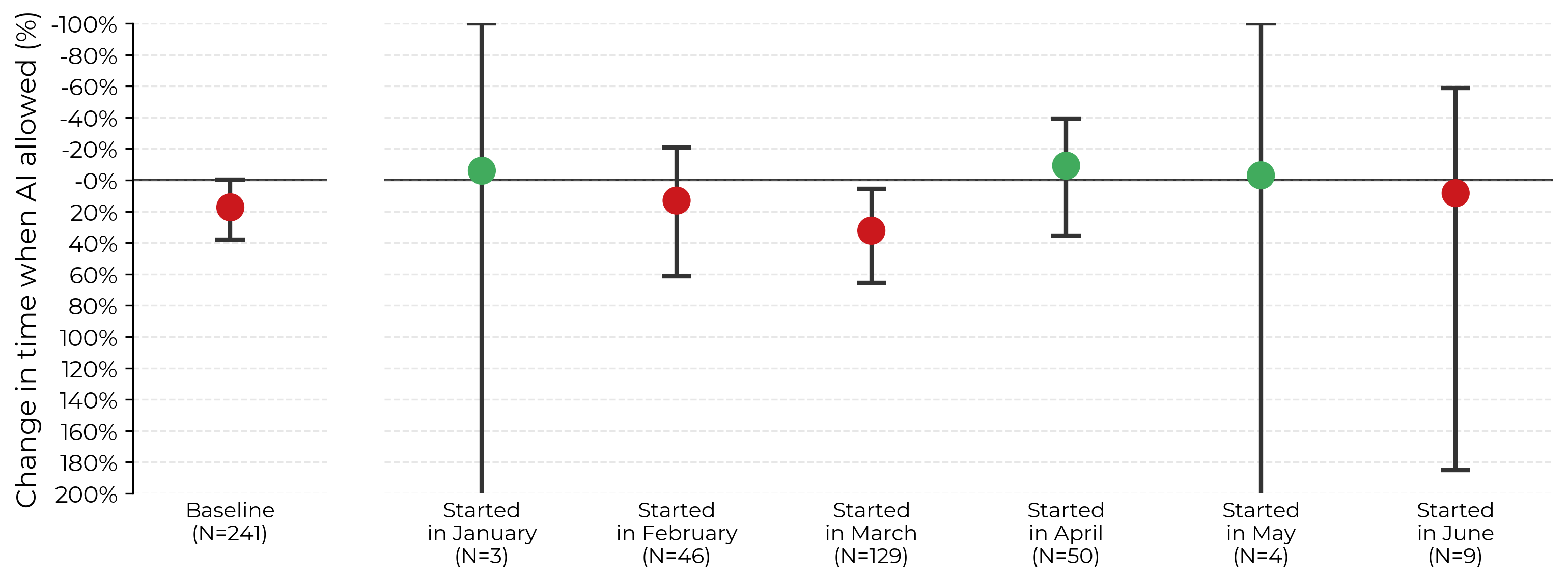}
    \caption{Speedup by month issue implementation started, as measured by first commit. Confidence intervals for January and May effects are cut-off for readability; the lower bounds are at approximately 1800\%. See \autoref{sec:heterogeneous_treatment_effect_estimation} for details on how we estimate heterogeneous treatment effects.}
    \label{fig:subset-started-month}
\end{figure}

\autoref{fig:subset-alternative-outcome} displays estimates using alternative outcome measures or subsets of our data. \autoref{fig:subset-started-month} displays treatment effects by the calendar month in which an issue implementation was started.

\section{Open-Source Development and AI Tooling Primers}
\label{sec:Primer on Open Source Development}

\subsection{Open-Source Development}

An open source software (OSS) project is typically defined by a \textit{repository}, which is a collection of code and assets. The repository for the popular pandas Python package, for example, can be found \href{https://github.com/pandas-dev/pandas}{here}. 

Any developer who contributes to a given repository is known as a \textit{contributor}. Active contributors for the pandas library are listed \href{https://pandas.pydata.org/about/team.html}{here}.

Contributors to an OSS project work off of \textit{issues}, which implicitly or explicitly describe tasks (bugs to fix, features to build, etc.) tracked within the repository. A few example pandas issues include \inlinecode{BUG: to\_dict(orient=`dict') does not convert np.nan to None in Pandas 2.2.3} (\href{https://github.com/pandas-dev/pandas/issues/61323}{link}) and \inlinecode{ENH: Enable nsmallest/nlargest on object dtype.} (\href{https://github.com/pandas-dev/pandas/issues/61166}{link}).

Contributors resolve an issue by submitting a \textit{pull request} (PR), which is a proposal to make changes to the repository. For example, \href{https://github.com/pandas-dev/pandas/pull/61154}{this PR} to the pandas repository fixes a bug by \inlinecode{Clip[ing] corr edge cases between -1.0 and 1.0}, addressing the corresponding issue \inlinecode{BUG: .corr() values significantly higher than 1.} (\href{https://github.com/pandas-dev/pandas/issues/61120}{link}).

After a contributor opens a PR, another contributor (often a maintainer) will review the PR. \textit{PR review} consists of reading and testing the code changes while paying attention to correctness, performance, and repository-specific code style. The reviewer may leave comments requesting changes; in the above PR, the author was asked if other functions needed a similar bug fix (\href{https://github.com/pandas-dev/pandas/pull/61154#pullrequestreview-2706439865}{link}).

After a PR is reviewed, the original contributor may make changes to address review comments. Multiple rounds of review may occur, although this is rare. After all review comments are addressed, the PR is merged into the repository. This results in the issue being marked as completed or closed.

Though this description of open-source software development is a reasonable default, diversity abounds. Some projects have many contributors, others only have a single contribute; some contributors do in-depth reviews, others merge in PRs without review at all.  

\textit{Stars} indicate the number of developers who have expressed interest in the repo. \textit{Forks} indicate the number of copies of the repository that have been made by developers so they can make their own modifications. Stars and forks can be seen as measures of repo popularity. The pandas library has 45,200 stars and 18,400 forks, making it an extremely popular repository.

\subsection{Primer on AI Tooling}
\label{sec:Primer on AI Tooling}

\subsubsection{Web Interfaces}

Many companies training large language models (LLMs) offer web-based user interfaces wherein users can chat with AIs. For example, users can interact with OpenAI models at \href{https://chatgpt.com/}{chatgpt.com}, Google DeepMind models at \href{https://gemini.google.com/}{gemini.google.com}, and Anthropic models at \href{https://claude.ai/}{claude.ai}. During our study period, popular LLMs offered by these developers include OpenAI’s GPT-4o, GPT-4.5, o1, o3-mini, o3, and o4-mini, Google DeepMind’s Gemini 2.5 Flash and Gemini 2.5 Pro, and Anthropic’s Claude 3.5 Sonnet (New) and Claude 3.7 Sonnet (although many/most issues were completed before the more recent models were released).

\subsubsection{Cursor}

Cursor is an integrated development environment (IDE) or `code editor'---a desktop application from which developers write and otherwise interact with code. It is a fork of the most popular code editor, Visual Studio Code (VSCode) \cite{stackoverflow2024}.

Cursor being a fork of VSCode enables developers to transfer their workflows from VSCode to Cursor to retain existing extensions and settings that they are most familiar with. Low switching costs are deepened by strong similarities in user interface and features between the two IDEs.

Relative to VSCode, Cursor is notable for having well-integrated AI tools, in particular ``Cursor Chat'' (previously separated into ``agent mode'' and ``chat mode'') and performant AI-powered autocomplete features.

\subsubsubsection{Chat and Agent Mode}

Cursor Chat allows users to prompt LLMs to make changes from inside the IDE. This LLM has tools enabled allowing it to autonomously explore your codebase, read documentation, run commands, and edit files. In practice, this LLM will generate code attempting to satisfy your prompt, and then show an in-file highlighted view of code changes that users can choose to accept or reject. 

(Previously ``agent mode'' was very similar to Cursor Chat, and ``chat mode'' was more similar to LLM web-based user interfaces, except model-agnostic and existing inside the IDE.)

The AIs that users typically interact with in Cursor are functionally identical to those they might interact with via web-based user interfaces, except for their additional access to relevant information in the repository, and often additional tool use that allows them autonomously run, test, and debug code they (or others) have written.

\subsubsubsection{AI Autocomplete}

Traditional IDEs have autocomplete functionality that suggests code completions as you type, primarily by fuzzy-matching on existing defined names in your codebase. For example, if you have a defined function called \inlinecode{add\_two\_numbers} and then later begin typing \inlinecode{add\_tw}, traditional autocomplete will suggest that you finish the completion with \inlinecode{o\_numbers}.

AI autocomplete is a feature in both VSCode and Cursor that uses an LLM to suggest edits to code as you write, and goes well-beyond just suggesting previously defined names. For example, if you started by defining a function with the signature \inlinecode{add\_two\_numbers(a,\ b):}, AI autocomplete would suggest a completion like \inlinecode{return\ a\ +\ b}.


\section{Recruitment and Onboarding}
\label{sec:RecruitmentAndOnboarding}

Open-source developers are recruited through a multi-stage process to select for active contributors to repositories that had more than 500 stars. Initial outreach was conducted via professional networks, ML-focused communities (Reddit’s r/Python, r/MachineLearning), and through GitHub profiles.
\begin{itemize}
    \item GitHub profiles are found by searching GitHub for the 250 most popular repositories, as well as those tagged with: ai, llm, deep-learning, neural-networks. 
    \item Contributors to these repositories are filtered to exclude those who had committed fewer than than five times in the previous three months.
\end{itemize}

51 developers filled out a preliminary interest survey, and we further filter down to about 20 developers who had significant previous contribution experience to their repository and who are able to participate in the study. Several developers drop out early for reasons unrelated to the study.

These developers are then given access to \href{https://cursor.com}{Cursor Pro}. We conduct a live 30-minute call with each developer where we provide a data collection template, answer basic questions about the experiment and their instructions, and give them training on how to use Cursor. Developers are considered trained once they can use Cursor agent mode to prompt, accept, and revert changes to a file on their own repository.

Additionally, for the duration of the study, we periodically provide feedback to developers on their implementation notes and video recordings. We occasionally email developers with tips on how to use Cursor more effectively if we notice low-hanging fruit (e.g. reminding developers to explicitly tag relevant files when prompting agents) from reviewing their screen recordings.

\subsection{Incentivization Scheme}
\label{sec:incentivization}

We pay developers \$150 per hour to participate in the study. Developers spend the majority of this time implementing issues, with fewer than five hours going to study overhead, including the onboarding call, check-in/feedback calls, the exit interview/survey, and the time they spend collecting their lists of issues.

An alterative incentivization scheme could give developers bonuses for completed issues, to incentivize developers to work as quickly as possible. However, this could cause developers to break issues into smaller chunks (e.g. to increase the total number of issues they complete) or reduce their quality standards to finish work more quickly, which could bias results. We expect that paying developers per hour overall has minimal effect on their behavior, beyond encouraging them to participate in the study.

\subsection{Developer Instructions and Survey Data}
\label{sec:dev_instructions_and_survey_data}

\subsubsection{Developer Instructions}
\label{sec:dev_instructions}

\begin{instructionbox}

\textbf{Overview}

METR is seeking software engineers who regularly work on large open-source projects to test the effectiveness of AI software engineering tools.
    
Apply here (bit.ly/ai-speedup-apply)

\textbf{Eligibility:}

You must:
\begin{enumerate}
    \item Have at least 1 year of professional experience as a software engineer
    \item Have at least 6 months experience as an active maintainer of the repository
    \item The repository you work on must be:
    \begin{enumerate}
        \item Open source
        \item At least 500 stars on GitHub or be manually reviewed by METR staff and deemed a high-quality, mature codebase (we know many good code bases don’t have a lot of stars)
        \item Have at least 3,000 lines of code (written by humans/in a major programming language, data etc doesn’t count)
        \item Have some kind of list of projects to improve it which would take between a few minutes to a few days, and which are relatively independent (i.e. a list of issues to fix, a list of features you intend to add, a general kanban board, etc). It’s ok if you make this list specifically for this experiment.
    \end{enumerate}
    \item Nice-to-haves:
    \begin{enumerate}
        \item The codebase is relevant to AI research and development or AI capabilities
    \end{enumerate}
\end{enumerate}

\textbf{Compensation}: 
\begin{enumerate}
    \item The total time commitment from a participant is a minimum of 20 hours, but we are interested in larger commitments. 
    \item We will pay you \$150 per hour. Note that during this experiment you will be working on tasks you’d already want to work on in your open source repository. We will be slightly randomizing the order of these tasks as well as what AI tooling you can use [note: we didn't do this, and we clarified this with developers before they began their work], but we don’t expect this to be a large impediment to your work. 
    \item This pilot study will last between 1-2 months, and we will limit funded development hours around 40.
\end{enumerate}

\textbf{Wait, how does it work?}

\begin{enumerate}
    \item Engineers will start by selecting a set of issues/to-dos from their open source repositories that they are looking to solve. 
    \item METR will then randomize these tasks into two buckets - on one set of issues, AI is allowed, and on the other set of issues, AI won’t be allowed. 
    \item You’ll work through these issues in whatever order you want - just making sure to only use AI when it’s allowed.
    \item As a participant, you will be doing work of your choosing on a repository of your choosing. This experiment will only change the order of tasks that you do and what LLMs you can use (including potentially restricting you to no LLMs)
    \item We are very flexible on when you complete these tasks. You can choose the date and time that works best for you (including weekends!).
    \item See more details in the Detailed Timeline.
\end{enumerate}

\textbf{Why this work matters:}
\begin{enumerate}
    \item AI models are becoming increasingly capable and automating parts of the workforce. We want to understand if or when it could reshape software engineering so we can predict and prepare for its effects.
    \begin{enumerate}
        \item In particular, we want to know when models might greatly speed up AI R\&D work, creating a feedback loop that would greatly accelerate AI progress
    \end{enumerate}
    \item Models are traditionally evaluated using simple, artificial benchmarks, where they are tested on their ability to answer multiple choice questions or fix some basic test cases in a Python library. These benchmarks:
    \begin{enumerate}
        \item Fail to measure how much models actually speed up engineers in their real workflow, the main real-world use-case for AI right now - and this is exactly what we’re attempting to measure with this experiment.
        \item Are typically artificial or have many tasks with no right answer, and lack the nuances and detail of real-world software engineering work
        \item Often require building “scaffolding” for the agents to autonomously write code etc. This scaffolding can be hard to develop and often means the AIs get stuck in places because of silly scaffolding issues. If a human is using the LLM, the scaffolding matters less (and is already widely commercially available) and the human can help get the LLM unstuck
        \item Get saturated quickly because the space between “a model can make any progress on a task at all” and “the model can do almost perfectly at the task” is small. Having the model speed up humans might address this issue because even very weak models can provide some human speedup and even very powerful ones are currently a ways away from being able to replace humans entirely.
    \end{enumerate}
    \item Although it isn’t a primary motivation of our work, we expect you might personally find it useful to know if AI actually speeds you up!
\end{enumerate}

\textbf{Detailed timeline}

Welcome to the METR Human Uplift Pilot!

This document contains an in-order list of the steps in this Uplift Study. Please feel free to leave any comments on the document.

\textbf{The Steps}

\textbf{Step 1: We have an introduction call}

We’ll have an introduction call, where I can give you an overview of the experiment and answer any questions you may have. You can book an introduction call here.

If you’re interested in moving forward, we’ll schedule a kickoff call (see below) for later that week

\textbf{Step 2: You collect issues}

Once you’re onboard, it’s time to make an issues list. 
The issue list can come in any format: an email, a Google Sheet, a Github project board. Make it whatever format is easiest for you.
Each issue should contain:
\begin{enumerate}
    \item A description. I don’t need to understand this, so feel free to keep it short.
    \item A label: bug fix, new feature, exploration, or refactor.
    \item Two time estimates:
    \begin{enumerate}
        \item No AI Time estimate:if you didn’t use any AI tools, how long would this take you?
        \item AI Time Estimate:  if you did use AI tooling to the best of your ability, how long would this take you?
    \end{enumerate}
    \item Task Expertise: 
    \begin{enumerate}
        \item Prior Task Exposure: Rate your previous experience with this specific type of task.
        \begin{enumerate}
            \item 1: Never done this type of task before
            \item 2: Have seen this type of task done but never done it myself
            \item 3: Have attempted this type of task once before
            \item 4: Have done this type of task multiple times before
            \item 5: I am an expert at this type of task
        \end{enumerate}
        \item External Resource Needs: how much documentation/reference material/research will you need to complete this task?
        \begin{enumerate}
            \item 1: I need extensive documentation / reference material / googling to complete this task.
            \item 2: I would need an occasional documentation / reference check / googling to complete this task.
            \item 3: I could complete this task entirely from memory and experience.
        \end{enumerate}
    \end{enumerate}
    \item Ideally, these issues should be less than 4 hours. If you can break them into $\leq$2 hour tasks, this would be ideal. If larger issues can reasonably be broken down into smaller PRs, feel free to take big issues and break them down into relevant steps.
    \item You should have at least 10 issues, and aim for at least 20 hours of issues, and up to 40 hours. 
\end{enumerate}

\textbf{Step 3: You send me issues, I bucket them}

Once your issue list is done, you can send it to me.
I’ll randomize this issue list into two buckets:
\begin{enumerate}
    \item AI bucket: you can use AI to help you on these issues
    \item No AI Bucket: you cannot use AI on these issues.
\end{enumerate}

\textbf{Step 4: We have a kickoff meeting
}
During the kickoff meeting, I’ll give you:
\begin{enumerate}
    \item The bucketed issue list
    \item Access to Cursor Pro (if you don’t already have it) as well as a basic training
    \item Access to Loom so you can record your screen.
    \item The Code of Conduct and Consent form
    \item Additionally, I can answer any final questions you might have about this experiment.
\end{enumerate}

\textbf{Step 5: You work on issues}

You’re ready to start now.  This should mostly look exactly like your normal work.
\begin{enumerate}
    \item You can work on the issues in any order you like. 
    \item You can work using any tools you like.
    \item However, if an issue is labeled “No AI”, then don’t use any AI tooling.
\end{enumerate}

You’ll record the data described here as you implement these issues.

Note: we will not share Loom videos without any humans outside of METR. We may watch them for quality control or use private LLMs to analyze these videos.

\textbf{Step 6: Checkin Call}
We’ll have one quick check in call to see how you’re doing, resolve any issues, and make sure we’re making progress.

\textbf{Step 7: Get Paid}
At the end of your issues, you’ll get paid. You’ll get \$150/hour for the number of hours you worked on tasks and created your issues - with a limit of 2 hours for issue creation.

\textbf{Data to Collect}
As you implement the issues in this project with and without AI, here is the additional information that you should collect.

\textbf{Implementation Notes}
The most important implementation note: if you’re working on an issue where AI is allowed, please record which models you use, and where you use them.

Other information is really useful to record as well. Please record any useful notes about the implementation that might be interesting for this study. For example:
\begin{enumerate}
    \item “Cursor implemented most of this code, with just a simple prompt from me.”
    \item “Cursor edited my package.json and I didn’t notice, which caused me to lose 30 minutes fixing dependencies.”
    \item “Not being able to use AI was tough, as there was a lot of boilerplate could I could have easily auto-generated”
\end{enumerate}

\textbf{Link to PR}
Link to the final PR that you implemented to solve this issue.

Note that if you implement a fix to multiple issues within one PR, just make sure to tag which commits correspond to which changes in that one PR.

\textbf{Screen Recording Link}
Link to a screen recording of the implementation of this issue. We ask that participants record their screen for all of the issues that they work on.

\textbf{Time Tracking}
We ask that you track two separate time categories: initial implementation time and post-review implementation time. These two time categories should sum to the “total amount of time it took for you to complete this feature to the point that it was mergable into the codebase.”

\textbf{Initial Implementation Time}
How long did it take you to get the PR up for review? 

Note that this should only include active time on your part. So for example, if you spent 2.5 hours over a week working on an issue, and then get a PR up and request a review, your initial implementation time should be 2.5 hours.

This chunk should include the time you spent:
\begin{enumerate}
    \item Understanding the issue.
    \item Implementing new code. 
    \item Writing tests or checking your work
    \item Getting a PR up for review.
    \item Etc.
\end{enumerate}

\textbf{Post-Review Implementation Time}

How long did it take to get the PR ready to merge post-first review?

Note that this also includes active time on your part. So if you get a PR up for review, have to wait three days for a review, and then have to make 20 minutes of changes as a result of the review, the post-review implementation time would be only 20 minutes.

This time bucket might include include:
\begin{enumerate}
    \item Time spent fixing code because of requested review changes.
    \item Time resolving merge conflicts.
\end{enumerate}
NOTE: If you did not get a review on your PR, or if the PR just approved your changes, then this time bucket would be zero minutes!

\textbf{Perceived Effort}
We ask you to rate the effort required to solve this issue on a scale of 1-5:
\begin{enumerate}
    \item Minimal effort: this issue was extremely easy to implement, and required very little effort or concentration. For example: making a simple text change to a webpage, refactoring code following a well-established pattern, copying an existing solution.
    \item Below-average effort: this issue was easy to solve, and required less effort than the average issue. For example: creating a new feature with a well-established design, writing unit tests for well-encapsulated functionality.
    \item Average effort: this issue required an average amount of effort to implement, and was not notably different from other issues. For example: creating a new feature with some novel components, tracking down a reproducible logic bug. 
    \item Above-average effort: this effort was hard to solve, and required more effort and concentration than the average issue. For example: Refactoring legacy code with limited tests, implementing complex algorithms or data structures.
    \item Maximum effort: this issue was extremely difficult to solve, and required very heavy effort and concentration. For example: re-architecting a major system redesign, debugging critical and complex production bugs with limited information.
\end{enumerate}

\end{instructionbox}

\subsection{Onboarding call and Cursor Training}
\label{sec:cursor_training}

All participating developers started the study with a 30 minute introduction and onboarding call. Before the call, developers were asked to set up an account on screen recording software (Loom), install and setup Cursor for their codebase, and read through the data they would be asked to collect over the course of the study.

As all developers had some previous experience with VSCode, developers were all able to setup and use Cursor on their codebases with little overhead. On the onboarding call, developers were given a basic training on Cursor agent mode to ensure they could:
\begin{itemize}
    \item Create a new agent mode instance on their own codebase.
    \item Add a relevant file to the context window of the agent.
    \item Prompt the agent to do make a change to this file. 
    \item Accept changes that the agent suggested to this file.
    \item Revert changes that they had previous accepted, undoing the agents changes.
\end{itemize}

Developers were also given a verbal overview of the data they were asked to collect, described in \autoref{sec:dev_instructions}, and given a chance to ask any questions they had about this data.

\subsection{Mid-experiment check-in calls}

All developers were offered periodic 15 minute check-in calls to assess their progress, answer any questions they had about the study, and ensure they were on track to complete issues in a timely manner. Most developers had between 1-4 check-in calls over the course of the experiment.  These calls also provided an opportunity to ask developers about their experience with using AI at that point in the study.

\subsection{Exit Interview}

All participating developers were interviewed at the termination of the study, during a 30 minute - 1 hour exit interview. Interview time ranged from 1 day to $\sim6$ weeks after developers finished their last issue, depending on available scheduling. 

The exit interview was unstructured, and designed to encourage developers to share their qualitative experience during the study. The following outline was followed during the exit interview, but not all questions were asked to all developers, depending on relevance.

\begin{instructionbox}

\textbf{Prior Usage:}

Collect the prior [AI, Cursor, etc.] usage information we have to confirm it in detail.

\textbf{Data Audit}

Look through their tracked issue data and confirm any data cleanup with them.

\textbf{Exit interview:}
\begin{enumerate}
    \item On task selection: how did the tasks you worked on compare to the average tasks you do on this open source repository? How were they different?
    \item During the study:
    \begin{enumerate}
        \item Did you use the same IDE for AI and non-AI tasks? Why?
        \item Was your experience in this study majorly different from your standard development on this repo? Why?
    \end{enumerate}
    \item On amount of effort:
    \begin{enumerate}
        \item Do you feel using AI or not affected how much effort you used on a given issue?
        \item How did your level of focus on these issues compare to normal work on this repo?
        \item How did time tracking or screen recording affect your working?
    \end{enumerate}
    \item On AI code-cleaning:
    \begin{enumerate}
        \item How good did you find the AIs outputs?
        \item How much cleaning did you do on the AI outputs?
        \item What is the code quality bar in your repo?
    \end{enumerate}
    \item On scope-creep:
    \begin{enumerate}
        \item Are there any issues that you gave up on because they were harder than you expected and so not worth it?
        \item Do you feel like the “size” of issues changed as a result of using AI? Specifically, do you feel like the issues were variable sized, and AI pushed you to go bigger?
    \end{enumerate}
    \item Going forward:
    \begin{enumerate}
        \item Do you plan to use AI tools going forward?
        \item Is this more than you planned to use them before the study?
        \item Did you increase the amount of AI that you used outside of the study as a result of the study?
        \item How did the study affect your belief in AI tools?
    \end{enumerate}
    \item On your AI skill level:
    \begin{enumerate}
        \item How confident are you that you use AI effectively now vs. at the start of the study? Do you feel you have improved at using AI?
        \item Did you notice an improvement in your ability to get useful work from the AI?
        \begin{enumerate}
            \item What specific strategies worked here?
        \end{enumerate}
    \end{enumerate}
    \item On AI effecting your work:
    \begin{enumerate}
        \item Did you find yourself sitting around and waiting on AI to generate code?
        \item Did you notice a change in idle or distracted time as a result of using AI or not AI?
    \end{enumerate}
    \item On the effectiveness of AI
    \begin{enumerate}
        \item Before the study:
        \begin{enumerate}
            \item What effect did you think AI tools would have on your time to complete issues? 
            \item What were the primary reasons you thought this?
        \end{enumerate}
        \item During the study?
        \begin{enumerate}
            \item How much do you believe AI changed your time to complete your issues?
            \item Specifically: Where did AI seem to speed you up? Where did AI seem to slow you down?
        \end{enumerate}
    \end{enumerate}
    \item Most effective AI tools:
    \begin{enumerate}
        \item What AI usage pattern feels the most effective for you?
        \begin{enumerate}
            \item Cursor vs. a web-browser? Why?
            \item What model do you prefer, why?
        \end{enumerate}
    \end{enumerate}
    \item Study Experience
    \begin{enumerate}
        \item Would you participate in this study again? Why or why not?
        \item What is one thing you liked about this study, and one thing we could improve?
        \item Anything else you wished I asked about?
    \end{enumerate}
\end{enumerate}

\end{instructionbox}

\subsubsection{Exit Survey}
\label{sec:exit_interviews}

\begin{instructionbox}

\textbf{METR Experiment Exit Interview}

\begin{enumerate}
    \item This form will take you about 15 minutes to complete.
    \item Please follow the instructions closely for each question.
    \item Do your best to answer accurately.
\end{enumerate}
Thank you for your participation - this is the last step in study participation!

\textbf{Questions:}

\begin{enumerate}
    \item What is your name?
    \item How many hours had you spent using LLMs before the start of this experiment?
    \begin{enumerate}
        \item 0 hours
        \item 1 - 10 hours
        \item 10 - 100 hours
        \item 100 - 1000 hours
        \item $>$ 1000 hours
    \end{enumerate}
    \item How many hours had you used Cursor before the start of this experiment?
    \begin{enumerate}
        \item 0 hours
        \item 1 - 10 hours
        \item 10 - 100 hours
        \item 100 - 1000 hours
        \item $>$ 1000 hours
    \end{enumerate}
    \item By the end of this study, how would you rate your skill level at Cursor?
    \begin{enumerate}
        \item Very Bad
        \item Below Average
        \item Average
        \item Above average
        \item Very Good
    \end{enumerate}
    \item On this repository, I typically make code changes through pull requests. \textit{True/False}.
    \item On this repository, I typically check my own code to make sure it's high quality. \textit{True/False}.
    \item On this repository, another developer typically reviews my code to ensure high code quality. \textit{True/False}.
    \item On this repository, I typically attempt to match repository style guidelines with my contributions. \textit{True/False}.
    \item This repository has a high quality bar for code contributions. \textit{True/False}.
    \item I typically only submit high quality PRs to this repository. \textit{True/False}.
    \item How much did AI decrease or increase the time it took you to complete the issues as part of this experiment?
    \begin{enumerate}
        \item If using AI resulted in you completing issues 2x faster, put 2.
        \item If using AI resulted in you completing issues 2x slower, put .5 (because 1/2 = .5) 
        \item If using AI did not change how long it took you to complete issues, put 1.
    \end{enumerate}
    \item During this study, what best describes how you read AI generated code that you included as part of your implementation?
    \begin{enumerate}
        \item I don't read AI generated code I use. I just check if it's outputs are correct.
        \item I typically skim AI generated code I use to see if it's correct.
        \item I typically read every line of AI generated code I use to check it's correct.
    \end{enumerate}
    \item During this study, what best describes how you edit AI generated code that you used as part of your implementation?
    \begin{enumerate}
        \item I usually take AI code as-is, without making edits.
        \item I usually make minor changes to AI generated code (like deleting comments or changing formatting).
        \item I usually make major changes to AI generated code (like deleting pieces of code, adding new features, or refactoring code)
    \end{enumerate}
\end{enumerate}
    
\end{instructionbox}

\subsection{Participant Dropout}

Over the course of the study, we stopped collecting work from three developers. Two of them were because the repository they contributed to paused development indefinitely, and the third developer was due to widespread cheating in the first set of issues they contributed. These developers were compensated fully for their work as part of the study, and we exclude their issues from all results.

\subsection{Developer and Repository Statistics}
\label{sec:devandrepostatistics}

Repository names and descriptions for repositories for which developers did not give consent to share their names are redacted. In \autoref{tab:repo_stats}, the repository age is calculated using the date of the first Git commit, which may be different from the actual start of the project (e.g. GHC was created 16 years before Git was invented).

\begin{center}
\begin{table}[htbp]
\centering \small
\begin{tabular}{rrrrrrrr}
\toprule
Dev & Repository & Months Since & Commit & Commit & AI-allowed & AI-disallowed \\
 & & First Commit & Count & Rank & Issues & Issues \\
\midrule
2 & \href{https://github.com/mito-ds/mito}{mito-ds/mito} & 30 & 3000 & 3/30 & 13 & 11 \\
3 & \href{https://github.com/stdlib-js/stdlib}{stdlib-js/stdlib} & 300 & 30000 & 3/300 & 9 & 12 \\
4 & \href{https://github.com/ghc/ghc}{ghc/ghc} & 30 & 300 & 30/3000 & 8 & 12 \\
5 & \href{https://github.com/haskell/cabal}{haskell/cabal} & 30 & 30 & 30/300 & 11 & 8 \\
6 & \href{https://github.com/stdlib-js/stdlib}{stdlib-js/stdlib} & 30 & 300 & 3/300 & 11 & 7 \\
7 & \href{https://github.com/flairNLP/flair}{flairNLP/flair} & 30 & 3000 & 3/300 & 12 & 5 \\
8 & \href{https://github.com/jsdom/jsdom}{jsdom/jsdom} & 300 & 300 & 3/300 & 8 & 9 \\
9 & \href{https://github.com/HypothesisWorks/hypothesis}{HypothesisWorks/hypothesis} & 30 & 300 & 3/300 & 11 & 6 \\
10 & \href{https://github.com/devflowinc/trieve}{devflowinc/trieve} & 30 & 300 & 3/30 & 10 & 5 \\
11 & \href{https://github.com/scikit-learn/scikit-learn}{scikit-learn/scikit-learn} & 30 & 300 & 30/3000 & 4 & 7 \\
13 & \href{https://github.com/EleutherAI/gpt-neox}{EleutherAI/gpt-neox} & 30 & 30 & 3/300 & 5 & 5 \\
16 & \href{https://github.com/huggingface/transformers}{huggingface/transformers} & 30 & 300 & 3/3000 & 1 & 1 \\
1 & Anonymized & 300 & 3000 & 3/300 & 15 & 13 \\
12 & Anonymized & 30 & 3000 & 30/300 & 9 & 2 \\
14 & Anonymized & 30 & 300 & 3/30 & 4 & 4 \\
15 & Anonymized & 30 & 300 & 3/30 & 5 & 3 \\
\bottomrule
\end{tabular}
\caption{Maintainer statistics for the study participants, sorted by total number of issues. The table shows representative values (nearest to $3 \times 10^x$) and percentages rounded to nearest bucket (10\%, 30\%, 50\%, 70\%, 90\%) to preserve anonymity while maintaining relative scale.}
\label{tab:maintainer_stats}
\end{table}

\end{center}

\begin{center}
\begin{table}[htbp]
\centering \small
\begin{tabular}{p{3cm}rrrrrrr}
\toprule
Repository & Stars & Forks & Committers & LoC & Age & AI-allowed & AI-disallowed \\
 &  &  &  &  & (years) & Issues & Issues \\
\midrule
\href{https://github.com/stdlib-js/stdlib}{stdlib-js/stdlib} & 5288 & 854 & 128 & 8M & 9 & 20 & 19 \\
\href{https://github.com/mito-ds/mito}{mito-ds/mito} & 2479 & 182 & 10 & 700k & 3 & 13 & 11 \\
\href{https://github.com/ghc/ghc}{ghc/ghc} & 3144 & 725 & 1008 & 1M & 19 & 8 & 12 \\
\href{https://github.com/haskell/cabal}{haskell/cabal} & 1679 & 717 & 532 & 300k & 21 & 11 & 8 \\
\href{https://github.com/flairNLP/flair}{flairNLP/flair} & 14234 & 2119 & 278 & 60k & 7 & 12 & 5 \\
\href{https://github.com/jsdom/jsdom}{jsdom/jsdom} & 21128 & 1745 & 350 & 1M & 15 & 8 & 9 \\
\href{https://github.com/HypothesisWorks/hypothesis}{HypothesisWorks/hypothesis} & 7939 & 617 & 355 & 100k & 12 & 11 & 6 \\
\href{https://github.com/devflowinc/trieve}{devflowinc/trieve} & 2380 & 205 & 68 & 800k & 2 & 10 & 5 \\
\href{https://github.com/scikit-learn/scikit-learn}{scikit-learn/scikit-learn} & 62760 & 26078 & 3164 & 400k & 15 & 4 & 7 \\
\href{https://github.com/EleutherAI/gpt-neox}{EleutherAI/gpt-neox} & 7266 & 1072 & 132 & 100k & 4 & 5 & 5 \\
\href{https://github.com/huggingface/transformers}{huggingface/transformers} & 147403 & 29761 & 2956 & 2M & 6 & 1 & 1 \\
Anonymized & 30000 & 3000 & 300 & 300k & 30 & 15 & 13 \\
Anonymized & 300 & 30 & 30 & 30k & 3 & 9 & 7 \\
Anonymized & 300 & 300 & 300 & 300k & 30 & 9 & 2 \\
\bottomrule
\end{tabular}
\caption{Repository statistics for the study, sorted by total number of issues. The table shows representative values (nearest to $3 \times 10^x$) for anonymized repositories.}
\label{tab:repo_stats}
\end{table}

\end{center}

\subsection{Screen Recordings}
\label{sec:screen_recording_instructions}

The screen-recording labeling process is time-intensive. As a result, screen recording labeling was started early in the data collection process, and to maximize the number of fully-labeled recordings, shorter recordings were prioritized first. Additionally, many developers choose to not record their screen if they were making a small set of changes due to a review. These factors may bias estimates of time allocation.

The following instructions were given to coordinate labeling screen recordings with fine-grained activity labels.

\begin{instructionbox}

\textbf{Overview}
\begin{enumerate}
    \item As part of an experiment we’re running, we’re labeling the loom videos that developers recorded of them implementing PRs on large open source repositories.
    \item The goal of this labeling is to understand how these developers actually spend their time when they are programming – so the labels include things like “writing\_code” or “reading\_code” or “reading\_docs.”
    \item A very important piece for us to understand is how they use and interact with AI. This practically means special labels around their use of Cursor Composer / Agent Mode.
\end{enumerate}

\textbf{Requirements:}
You’re a good fit for labeling this data if:
\begin{enumerate}
    \item You know how to program well, and when watching someone program over their shoulder can figure out what they are working on. 
    \item You have used Cursor Composer before, and know how that works!
\end{enumerate}

\textbf{Compensation}
\begin{enumerate}
    \item We’ll pay standard per hour rates for image labeling. 
    \item We’ll be checking 1/10 of the submissions. If your timing labels are sufficiently accurate (close to our hand-checked solutions), we’ll give you a \$250 bonus.
\end{enumerate}

\textbf{How to Label:}
\begin{enumerate}
    \item First, scroll down and read the labels below. Feel free to leave comments if you have any questions about these.
    \item Then, open the tracking sheet:
    \begin{enumerate}
        \item Claim one of the unclaimed videos in the “To Label” sheet by putting your name in one of the columns. Go in-order, so we get shorter videos first.
        \item Then, make a new tab, and copy over the `Template` tab. Name the new tab as the initials of the person who made the recording, followed by a dash, and then the issue\_id number.
    \end{enumerate}
    \item Open the loom video link:
    \begin{enumerate}
        \item You probably need access to METR’s loom account for this; if you do not have access, please ask us and we’ll add you!
        \item If you have access to METR’s loom account but do not have access to the particular loom video you opened from the sheet, please do not request access. Just mark this in the sheet, and move on to the next video.
    \end{enumerate}
    \item Take notes using this tool. It makes Loom note taking much easier, and means you don’t have to leave the loom page!
    \begin{enumerate}
        \item The default rate is 5, but you can adjust this with `rate 2` to make it slower. 
        \item You can drag this note taking app around the screen, and it works in fullscreen mode. Click instructions at the bottom to see more commands!
        \item Note: please try and make the start and end time of your notes correspond to the actual start and end times of the things users are doing. This might require rewinding the video! 
    \end{enumerate}
    \item After you’re done watching the video and taking notes, type `done` and copy the results into the sheet.
    \item Note: if the video is $>$20 minutes long, copy your notes out in 20 minute chunks, to make sure that your labels are as accurate as possible.
    \item Then, go through the `Label` column of the sheet, and label each chunk of time with the columns below.
    \item Then, go through the `AI Use Label` and `AI Type Label` column and label any of the AI usage with what the developer is using the AI for as well as the model/UI being used.
\end{enumerate}

\textbf{The Labels}

Label accuracy is very important for this data work. As such, it requires a fair bit of critical thought about what the user is really engaged in.

If you’re not sure what the user is doing, please put “unknown” as the label. We can always go back and fill things in, but only if you note this. You can also leave a comment to the side of the row describing what’s confusing to you!

If you think that there’s a better label for things than one provided, feel free to add it + tag me in a comment on top of it. I can then add it to the list here :) 

\textbf{The Labels}
Most Common
\begin{enumerate}
    \item reading\_issue: the dev is reading the issue that they are planning to implement a fix for as a part of this loom video. 
    \item writing\_code: the dev is actively writing code. They might also be reading a bit or navigating around, but mostly they are editing/writing code on the page. (Note that writing testing code is counted differently).
    \item reading\_code: the dev is primarily reading existing code. They might be navigating through the codebase to find specific things, but in practice they are reading.
    \item reading\_docs: the dev is reading documentation. Potentially of their own codebase, potentially of some other codebase/tool. It’s not code they are reading.
    \item writing\_docs: the users are writing documentation. This could include release notes or a documentation page.
    \item writing\_tests: the user is writing testing code, rather than writing some other type of code.
        \item test\_running\_tests: is actually running the testing code, and looking at the results.
    \item test\_running\_ci: is running or waiting on CI checks that are running on Github/Gitlab.
    \item test\_manually\_checking: testing some solution, but doing this by hand (either by writing some code, or looking at some artifact/output).
    \item replicating\_bug: replicating a bug, normally the bug described in the initial issue.
    \item running\_debugger: if the user running the debugger, then note this here.
    \item compiling: waiting on some code to compile.
    \item setup: running some setup process (e.g. opening their IDE, installing extensions, etc).
\end{enumerate}

\textbf{Git Related Things:}
\begin{enumerate}
    \item branching: creating a new branch, and adding on to it.
    \item committing: adding files or writing a commit message. Some folks try hard on these!
    \item pr: getting up a PR and potentially writing a PR message.
    \item git: some other misc. git operation (e.g. if they are viewing Git diffs, or something).
\end{enumerate}

\textbf{Misc:}
\begin{enumerate}
    \item thinking: the user is not AFK, but appears to be thinking through what they are going to do next.
    \item unrelated: the user is doing something unrelated, like watching a youtube video or changing their music.
    \item paused: the user appears to have stepped away from their computer.
    \item communicating\_with\_teammates: for example, the user switches to slack or discord and asks a question.
    \item broken: something is wrong with the loom video.
\end{enumerate}

\textbf{On using AI tooling}

Note: it can be a bit hard to label AI generated usage. The key details here are to describe the full flow of how the user writes a prompt, waits, and then either accepts or rejects it – and then what they do after.

For a good example of what this should look like, see this labeling.

\begin{enumerate}
    \item writing\_prompt: the user is writing a composer prompt.
    \item waiting\_on\_generation: the user is waiting on the AI to generate code or a response.
    \item reading\_generation: the user is spending time actually reviewing the suggestions the AI has made.
    \item generation\_taken: the user takes the suggestion from the AI, and it turns out to be useful / they don’t ditch it in the future.
    \item generation\_rejected: the user does not take the suggestion from the AI, or takes the suggestion and then reverts back to before they took the suggestion. In other words, they don’t use the AI generated code because it’s broken.
    \item ai\_code\_cleaning: if the user takes a suggestion from the AI, and then spends time cleaning that code, then we label this not as writing\_code but instead as ai\_code\_cleaning. This includes changing spacing, minor refactors, etc. As long as the users keeps the bulk of the code, this is considered a generation\_taken.
\end{enumerate}

Feel free to also use `ai\_docs\_cleaning` or `ai\_commit\_cleaning` if this is what the user is cleaning up.

\textbf{AI Use Label:}
We also ask that you fill out the AI use label column to describe 
\begin{enumerate}
    \item new\_feature: the user is using AI to extend functionality of the codebase.
    \item bug\_fix: the user is prompting the AI to fix a bug in the existing codebase.
    \item code\_search: the user is using the AI to search their codebase for some code / implementation detail. 
    \item tests: the user is having the AI generate code for testing reasons.
    \item docs: the user is using the AI to write docs. Could include readme, or git commit messages.
    \item question: the user has a question (e.g. one they could ask google) that they are adding here.
    \item integration: the user is integrating some code with an external system, and so the code-gen is primarily for understanding or integrating with that system.
    \item refactoring: improving code, without extending the code’s functionality or fixing bugs
\end{enumerate}

NOTE: if you think there are other composer usages that this better fits into: please feel free to just write what you think best describes what the user is doing here!

\textbf{AI Labels:}
We also ask that you mark three columns that describe where / how the user is using AI:
\begin{enumerate}
    \item AI Model Label: The model being used E.g. “3.7 Sonnet” “3.5 sonnet” “o1” “o1-preview” “gpt4.5”
    \item AI UI Label: The UI being used E.g. “cursor composer” “cursor chat” “web UI” (the respective LLM providers’ chat website)
\end{enumerate}

This should co-exist with the AI Use Label above. So, for example, a segment of video in which someone is writing/reading cursor compose might have AI Use Label “new\_feature” and AI Type Label “Sonnet 3.7, cursor composer”

\textbf{FAQ:}
\begin{enumerate}
    \item \emph{How accurate do the timestamps need to be?} Roughly correct. It’s ok if the timestamps are off by a few seconds on each end, but in general you should try and avoid large (e.g. 10+ second) errors.
    \item \emph{When is a user writing code vs. reading code?} These are often interleaved. In practice, if the user is writing code for $>$50\% of the chunk of time, this is writing code – only if they are reading code for like $>$30 seconds is it really like a concrete “reading\_code” time. 
    \item \emph{I have never used Cursor composer.} You’re probably not a great fit for this labeling, in this case. 
    \item \emph{I am not sure how to label things.} If you are confused about how things should be labeled (e.g. there’s some weird cursor flow you don’t understand where a user accepts code changes, and then later reverts), just label things as “unknown” and we can come back to it.
\end{enumerate}

\end{instructionbox}

\subsection{Instructions Given to Expert Forecasters}
\label{sec:instructions-given-to-expert-forecasters}

\begin{instructionbox}

\textbf{Supplementary Information for METR AI speedup study survey}

METR is currently running a field experiment measuring how AI tools impact open source developer productivity.

The TL;DR is that we recruit experienced developers who contribute to popular open source projects, randomize their tasks to having no AI or AI allowed, and measure the ratio between the time it takes a human to complete tasks with AI vs. without AI. The study aims to measure speedup in conditions that closely mirror normal software development.

In this supplementary information document, we first describe two pieces of relevant background: the structure of open source software development, and AI tooling. (If you are highly familiar with open source software development or cursor agent mode you should probably skip the respective sections.) We then describe the experiment in more detail: how we sampled developers and repositories, the tasks developers work on, how developers participate in the study, and finally how we intend to estimate speedup due to AI.

(We are only part-way through running the study, so do not yet know the final result ourselves.)

\textbf{Background}

[\autoref{sec:Primer on AI Tooling} was then included.]

\textbf{Experiment}

\textbf{Contributor recruitment}

Open-source contributors were recruited through a multi-stage process to select for active contributors to repositories that had more than 500 stars.

\begin{enumerate}
    \item Initial outreach was conducted via professional networks, ML-focused communities (Reddit’s r/Python, r/MachineLearning), and through GitHub profiles.
    \begin{enumerate}
        \item GitHub profiles were found by searching GitHub for the 250 most popular repositories, as well as those tagged with: ai, llm, deep-learning, neural-networks. 
        \item Contributors to these repositories were filtered to exclude those who had committed less than five times in the previous three months, and then emailed.
    \end{enumerate}
    \item Interested contributors (n=50) filled out a preliminary survey to assess:
    \begin{enumerate}
        \item Years of software development experience
        \item The repositories they contribute to
    \end{enumerate}
\end{enumerate}

All contributors who planned to contribute to repositories with more than 500 stars were offered an introductory call to provide an overview of the study timeline and parameters. 31 calls were conducted, with half of developers being filtered out for a lack of previous contribution experience or because the timeline didn’t work.

The remaining 16 developers were then given access to Cursor Pro. We had a 30-minute call with each developer where we set them up with a data collection template, answered questions, and trained them on Cursor. Developers were considered trained once they could use Cursor agent mode to prompt, accept, and revert changes to a file on their own repository. 94\% of developers noted that they had used web-based LLMs as part of their development workflow before participating in our experiment. Rates of past usage of Visual Studio Code, Github Copilot, and Cursor are 63\%, 56\%, and 25\% respectively [note: these were preliminary numbers, and are lower than the true values reported in the paper].

All participating repositories are listed below.

[\autoref{tab:repo_stats} was then included.]

\textbf{Issues}

Each contributor maintained a list of issues to work on as part of this study. Contributors were asked to select issues as they would during normal development on this study, with the caveat that they should break issues that were likely to take $>4$ hours into sub-issues that take $\geq2$ hours if possible.

The issues are intended to be as similar as possible to those that would have been worked on if this study never took place.

After collecting this issue list, each issue was randomized to either AI-allowed or AI-disallowed conditions. If AI is allowed, developers can use any AI tools they so choose, including no AI tooling if they deem it not helpful to the problem. If AI is disallowed, no generative AI tooling can be used.

\textbf{Study Participation}

Contributors completed issues much as they would outside of our experiment, with two important differences: they record their screen as they work, and they take implementation notes post issue completion. (We use human-labelled video recordings covering the majority of issues to confirm compliance.)

For the duration of the study, we periodically check in with developers and provide feedback on their implementation notes and loom videos. We occasionally emailed developers with tips on how to use Cursor more effectively if we notice some easy wins in their Loom videos.

\textbf{Measuring speedup}

We aim to measure the speedup factor due to AI, defined as:

S = mean(completion time with no AI) / mean(completion time with AI allowed).

S = 2 would indicate issues assigned to AI allowed taking half the time of issues assigned to no AI (100\% speedup); S = 1 would indicate that issues take the same time to complete with and without AI being allowed (0\% speedup); S = 0.5 would indicate that issues assigned to AI allowed take twice the time of issues assigned to no AI (-50\% speedup).

(We are asking you to predict S, i.e. the quantity taking value 2/1/0.5 rather than 100\%/0\%/-50\% in the examples.)

\end{instructionbox}

\end{document}